\pdfoutput=1

\documentclass{article} %
\PassOptionsToPackage{table,dvipsnames}{xcolor}
\usepackage{iclr2027_conference,times}

\usepackage{amsmath,amsfonts,bm}

\def\eqref#1{equation~\ref{#1}}

\def\1{\bm{1}}

\DeclareMathAlphabet{\mathsfit}{\encodingdefault}{\sfdefault}{m}{sl}
\SetMathAlphabet{\mathsfit}{bold}{\encodingdefault}{\sfdefault}{bx}{n}

\usepackage{hyperref}
\usepackage{url}
\usepackage{xspace}
\usepackage{booktabs}
\usepackage{amssymb}
\usepackage{placeins}
\usepackage{wrapfig}

\newcommand{\method}{LAW\xspace}
\newcommand{\statemap}{\Phi}

\newcommand{\eot}{{\color{gray}\texttt{<|endoftext|>}}}
\newcommand{\nl}{{\color{gray}\P}}
\newcommand{\qdbl}{{\fontencoding{T1}\selectfont\symbol{34}}}
\newcommand{\lawsample}[1]{%
  \begin{tcolorbox}[colback=ourblue!6, colframe=ourblue!45, boxrule=0.5pt, arc=3pt,
                    left=8pt, right=8pt, top=6pt, bottom=6pt, fontupper=\small]
  #1
  \end{tcolorbox}}
\usepackage{amsmath}
\usepackage{amssymb}
\usepackage{mathtools}
\usepackage{amsthm}
\theoremstyle{plain}

\theoremstyle{definition}

\theoremstyle{remark}

\usepackage{bm}
\usepackage{graphicx}
\usepackage{subcaption}
\usepackage{float}
\usepackage{wrapfig}
\usepackage{array}
\usepackage{multirow}
\usepackage{dsfont}
\usepackage{cancel}
\usepackage{soul}
\usepackage[normalem]{ulem}
\usepackage{siunitx}
\usepackage{algorithm}
\usepackage{algorithmic}
\usepackage[page,toc,titletoc,title]{appendix}
\usepackage[capitalize,noabbrev]{cleveref}
\usepackage{colortbl}
\usepackage{tabularx}
\usepackage{tcolorbox}
\usepackage{mdframed}
\usepackage{pifont}
\usepackage{xargs}
\usepackage{cases}

\hypersetup{
	final,
	colorlinks,
	linkcolor=ourred,
	citecolor=ourblue,
	urlcolor=url,
}

\definecolor{ourblue}{rgb}{0.368,0.507,0.71}
\definecolor{ourorange}{rgb}{0.881,0.611,0.142}
\definecolor{ourgreen}{rgb}{0.56,0.692,0.195}
\definecolor{ourred}{rgb}{0.923,0.386,0.209}
\definecolor{ourviolet}{rgb}{0.528,0.471,0.701}
\definecolor{ourbrown}{rgb}{0.772,0.432,0.102}
\definecolor{ourlightblue}{rgb}{0.364,0.619,0.782}
\definecolor{ourdarkgreen}{rgb}{0.572,0.586,0.}

\definecolor{ourcyan2}{rgb}{0.125,0.722,0.804}
\definecolor{ourred2}{rgb}{0.863,0.184,0.047}
\definecolor{ouryellow2}{cmyk}{0,0.16,1.0,0.07}
\definecolor{ourviolet2}{cmyk}{0.55,0.56,0,0.47}
\definecolor{ourorange2}{cmyk}{0,0.46,0.89,0.11}

\definecolor{discretecolor}{RGB}{11,83,150}
\definecolor{gaussiancolor}{RGB}{230,145,56}
\definecolor{argmaxcolor}{RGB}{154,0,0}
\definecolor{grayseq}{RGB}{120,120,120}
\definecolor{url}{HTML}{c55a3a}
\definecolor{ourslightgray}{RGB}{235, 235, 235}

\newcommand{\Fig}[1]{Figure~\ref{#1}}
\newcommand{\fig}[1]{Fig.~\ref{#1}}

\newcommand{\tab}[1]{Table~\ref{#1}}
\newcommand{\Eqn}[1]{(\ref{#1})}

\renewcommand{\sec}[1]{Sec.~\ref{#1}}
\newcommand{\supp}[1]{Suppl.~\ref{#1}}

\newcommand{\algo}[1]{Algo.~\ref{#1}}

\newcommand{\mask}{[\textsc{mask}]~}

\newcommand{\barx}{\bar{\mathbf{x}}}

\usepackage{xspace}
\makeatletter
\DeclareRobustCommand\onedot{\futurelet\@let@token\@onedot}
\def\@onedot{\ifx\@let@token.\else.\null\fi\xspace}
\def\eg{e.g\onedot}

\makeatother

\newcommand{\Comment}[1]{\hfill{\color{gray}\footnotesize$\triangleright$~#1}}

\def\x{{\mathbf x}}

\def\supl{{\color{grayseq}\ell}}

\def\z{{\mathbf z}}
\def\at{\alpha_t}
\def\as{\alpha_s}

\def\cat{\mathrm{Cat}}

\def\felbo{\mathcal{L}_{\mathrm{tok}}}
\def\denoise{\mathbf{x}_\theta}
\def\qst{q_{s|t}}

\def\xl{\x^{\supl}}
\def\ztl{\z_t^{\supl}}

\title{Loopholing All the Way: Language Modeling with pure Self-Conditioning}

\author{\begin{tabular*}{\dimexpr\textwidth-2\tabcolsep\relax}[t]{@{}l@{\extracolsep{\fill}}l@{\extracolsep{\fill}}l@{}}
Adrien Deschenaux$^{*\spadesuit}$ & Nadezhda Ilieva$^{*\heartsuit}$ & Justin Deschenaux$^{*\spadesuit}$ \\[3pt]
Manuel Madeira$^{\spadesuit}$ & Dorina Thanou$^{\spadesuit}$ & Aurelien Lucchi$^{\heartsuit}$ \\[3pt]
Antonio Orvieto$^{\clubsuit}$ & Caglar Gulcehre$^{\spadesuit,\diamondsuit}$ & Pascal Frossard$^{\spadesuit}$
\end{tabular*} \\ \noalign{\vskip 2pt}
\normalfont $^{\spadesuit}$EPFL, Lausanne, Switzerland \quad $^{\heartsuit}$University of Basel, Switzerland \quad $^{\diamondsuit}$Microsoft AI \\
\normalfont $^{\clubsuit}$ELLIS Institute T\"ubingen, MPI for Intelligent Systems, T\"ubingen AI Center, Germany
\vspace{-8pt}%
}

\iclrpreprint %
\begin{document}

\maketitle
{\renewcommand{\thefootnote}{*}\footnotetext{Equal contribution.}}

\begin{abstract}
Recent Diffusion Language Models (DLMs) use Self-Conditioning (SC), or Loopholing, to improve their performance.
While SC is traditionally seen as an auxiliary heuristic, its empirical success leads us to ask: \textbf{can a continuous generative model be based purely on Self-Conditioning?}
We answer affirmatively with \method (``Loopholing All the Way''), which propagates the \emph{Expected Embedding} under the denoiser's prediction across steps, decoding the tokens once, at the end.
On Sudoku, \method outperforms previous Discrete and Continuous Diffusion Models, including Loopholing Discrete Diffusion. On Maze, \method is the most accurate from 2 to 64 NFE.  After training on TinyGSM and sampling at temperature 1 with 1024 NFE, \method solves 42.2\% of GSM8K problems against 34.5\% for uniform Loopholing Discrete Diffusion. Additionally, fine-tuning a pre-trained Uniform Diffusion Model (Duo) with \method increases its accuracy from 17.2\% to 48.1\%. Finally, on OpenWebText, \method pushes the Pareto frontier and approaches Autoregressive Models (ARMs) with 64 NFE. Find our code at \href{https://github.com/ilievanadezhda/law}{\texttt{https://github.com/ilievanadezhda/law}}.
\end{abstract}
\begin{figure}[!h]
    \centering
    \includegraphics[width=\linewidth]{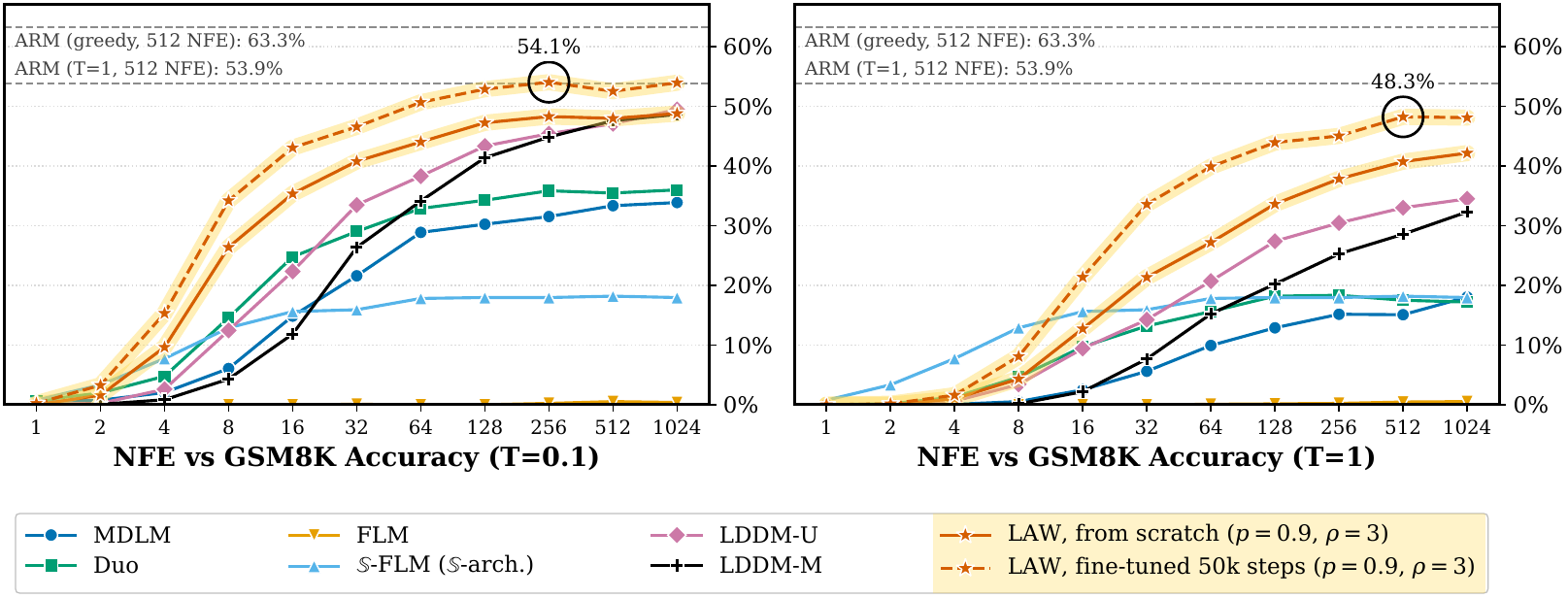}
    \caption{\textbf{GSM8K accuracy against the Number of Function Evaluations (NFE) at sampling temperature $T=0.1$ (left) and $T=1$ (right).} \method (\sec{sec:experiments-gsm8k}) trained from scratch with the training FLOPs of the baselines (solid), and fine-tuned from the Duo baseline for 50k steps (dashed). The dashed gray lines are an Autoregressive Model (ARM), sampled at $T=1$ and with greedy decoding. Trained from scratch, \method is the most accurate model from 32 NFE at $T=1$, and from 4 to 512 NFE at $T=0.1$. Fine-tuned from Duo, it comes closer to the ARM.}
    \label{fig:gsm8k-nfe}
\end{figure}
\section{Introduction}
Although Autoregressive Models (ARMs) dominate language modeling~\citep{vaswani2017attentionneed, kaplan2020scalinglawsneurallanguage, geminiteam2025gemini, grattafiori2024llama3herdmodels}, their generation process is sequential and their left-to-right factorization can be limiting for tasks that need bidirectional context~\citep{kitouni2024factorizationcursetokenspredict, papadopoulos2024arrowstimelargelanguage, nagarajan2025roll}. Discrete Diffusion Models (DDMs)~\citep{austin2023structureddenoisingdiffusionmodels, campbell2022continuoustimeframeworkdiscrete, lou2024discretediffusionmodelingestimating, sahoo2024simpleeffectivemaskeddiffusion, shi2025simplifiedgeneralizedmaskeddiffusion, ou2024radd, sahoo2025diffusionduality, schiff2025simpleguidancemechanismsdiscrete, vonrütte2025generalizedinterpolatingdiscretediffusion}, the most common Diffusion Language Models (DLMs), instead use a bidirectional architecture and can refine tokens in parallel. For tractability, DDMs sample each position independently and discard the denoiser's prediction after each step. This causes two problems. First, factorized, parallel sampling hurts the sample quality in few-step generation~\citep{deschenaux2024promisesoutlookschallengesdiffusion, feng2025theoreticalbenefitlimitationdiffusion}. Second, discarding previous predictions means that the associated uncertainty is lost (\emph{information collapse}; \citealp{jo2025loopholingdiscretediffusion}).

Continuous DLMs \citep{dieleman2022continuousdiffusioncategoricaldata, li2022diffusionlmimprovescontrollabletext, gulrajani2023likelihoodbaseddiffusionlanguagemodels, jo2025continuousdiffusionmodellanguage, batzolis2026cobit, chemseddine2026sphericalflowscategorical, chen2026langflowcontinuousdiffusion, deschenaux2026hypersphericalflows, hu2026elfembeddedlanguageflows, lee2026onesteplanguagemodelingcontinuous, potaptchik2026discreteflowmaps, roos2026categoricalflowmaps, shen2026codarcontinuousdiffusionlanguage, yang2026continuousdiffusionscales, yoo2026selfconditionedflowmaplanguage} instead avoid the factorized sampling by integrating an ODE or SDE from noise to data~\citep{albergo2023buildingnormalizingflowsstochastic, lipman2023flowmatchinggenerativemodeling, liu2022flowstraightfastlearning}. However, on reasoning, they at most match DDMs. For example, FLM~\citep{lee2026onesteplanguagemodelingcontinuous} and $\mathbb{S}$-FLM~\citep{deschenaux2026hypersphericalflows} remain close to MDLM and Duo on Sudoku and Maze, and $\mathbb{S}$-FLM matches MDLM on GSM8K only with greedy generation.

In contrast, Self-Conditioning (SC; \citealp{strudel2022selfconditionedembeddingdiffusion, chen2022analogbits}) feeds the previous predictions or hidden representations of the denoiser back as input, and improves both discrete and continuous DLMs. For example, Loopholing Discrete Diffusion Models (LDDMs; \citealp{jo2025loopholingdiscretediffusion}) and DiffusionGemma~\citep{diffusiongemma2026} propagate such a state across denoising steps. On hard Sudoku puzzles, LDDM-U solves 91.0\%, against 58.4\% for Duo and at most 45.0\% for FLMs (\sec{sec:experiments-sudoku}). So far, despite its empirical success, SC always plays an auxiliary role. We therefore ask:
\begin{tcolorbox}[colback=ourblue!10, colframe=ourblue!10, boxrule=0pt, sharp corners, left=4pt, right=4pt, top=3pt, bottom=3pt, halign=center]
\emph{``Can we find a generative model based \underline{purely} on Self-Conditioning?''}
\end{tcolorbox}
\begin{figure}[t]
    \centering
    \includegraphics[width=\linewidth]{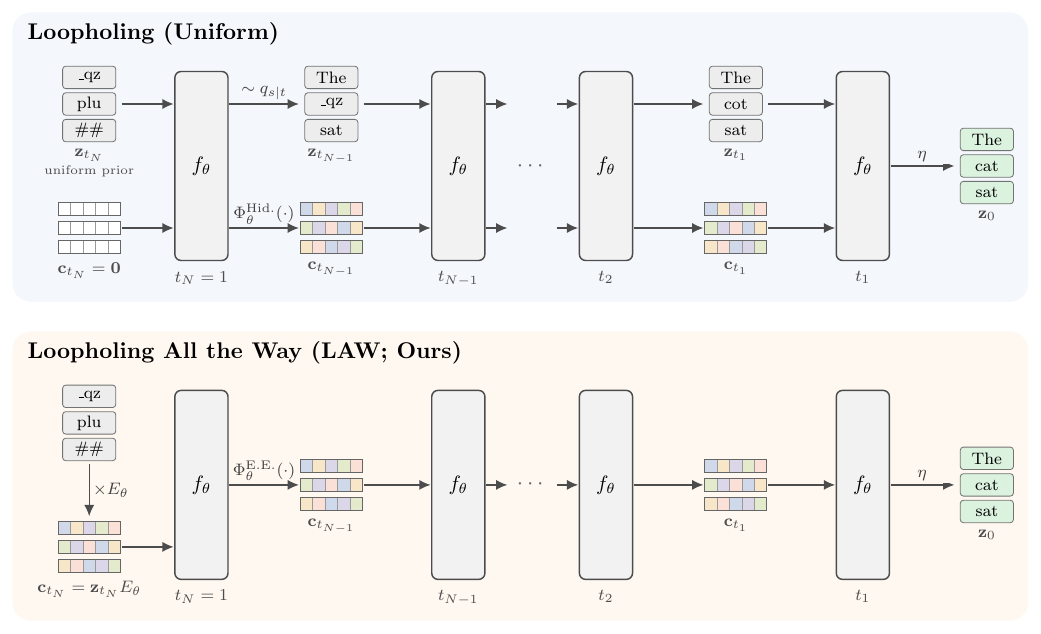}
    \caption{\textbf{Overview.} Sampling goes from left to right, applying the backbone $f_\theta$ repeatedly. Boxed words are discrete tokens, colored cells SC states and white cells zero-valued states. \textbf{Top:} Loopholing (LDDM; \citealp{jo2025loopholingdiscretediffusion}). At each step, discrete tokens are updated and forwarded alongside the SC state~\Eqn{eq:loopholing}. \textbf{Bottom:} \method. At each step, we only propagate the SC state, computed by the state map $\statemap_\theta^{\mathrm{E.E.}}$~\Eqn{eq:law-transition}. The initial tokens are drawn from the uniform prior, and we decode back to discrete tokens only at the end~\Eqn{eq:law-decoding}.}
    \label{fig:law-overview}
\end{figure}
We answer this question positively with \textbf{``Loopholing All the Way''} (\method, \Fig{fig:law-overview}):
\begin{itemize}
\item We propose \method, \textbf{a continuous generative model based purely on Self-Conditioning}. As in Uniform Diffusion Models (UDMs), sampling starts from a sequence of uniformly random tokens. Afterwards, \method refines the \emph{Expected Embedding} of the clean token, under the denoiser's prediction. Thus, we decode back to tokens only once, at the last sampling step.
\item We identify the two choices that make pure SC work: the \emph{Expected Embedding} as SC state (35.8\% on GSM8K, against below 1.5\% with the final-layer hidden representation of \citealp{jo2025loopholingdiscretediffusion}), and \emph{temperature regularization}~\Eqn{eq:law-temperature}, which randomly scales the denoiser's logits during training, doubles the GSM8K accuracy at $T=0.1$ (24.8\% to 48.8\%).
\item \method surpasses LDDM-U on hard Sudoku (96.5\% vs.\ 91.0\%, \sec{sec:experiments-sudoku}) and is the most accurate model on Maze with up to 64 NFE (\sec{sec:experiments-maze}). On GSM8K, it outperforms the diffusion and flow baselines trained with the same FLOPs, and fine-tuning a pre-trained Duo checkpoint with \method increases the accuracy from 17.2\% to 48.1\% (\sec{sec:experiments-gsm8k}). On OpenWebText, at the same entropy, its Generative Perplexity (Gen.\ PPL) comes close to that of ARMs with 64 NFE. With 1024 NFE, \method's samples have lower Gen.\ PPL than those of the AR baseline, though the entropy reaches at most 5.1--5.2 over the swept temperatures (\sec{sec:experiments-owt}).
\end{itemize}
\section{Background}
\label{sec:background}
\paragraph{Notation.}
The vocabulary $\mathcal{V}=\{\mathbf{e}_1,\dots,\mathbf{e}_{|\mathcal{V}|}\}$ is the set of one-hot vectors $\mathbf{e}_i\in\{0,1\}^{|\mathcal{V}|}$, each associated to a distinct token. Sequences of $L$ tokens, clean ($\mathbf{x}$) or noisy ($\mathbf{z}_t$, $\mathbf{z}_s$), are one-hot matrices in $\mathcal{V}^L\subset\{0,1\}^{L\times|\mathcal{V}|}$, and $\mathbf{x}^{\supl}$ is the $\ell$-th row. For $\boldsymbol{\mu}$ in the probability simplex $\Delta^{|\mathcal{V}|}\subset\mathbb{R}^{|\mathcal{V}|}$, $\mathrm{Cat}(\cdot\,;\boldsymbol{\mu})$ is the categorical distribution and $\mu_i=\langle\boldsymbol{\mu},\mathbf{e}_i\rangle$ is the $i$-th entry, where $\langle\cdot,\cdot\rangle$ is the inner product. We write $\odot$ for the elementwise product, $\mathbf{1}$ for the all-ones vector, $\mathds{1}_{\{\cdot\}}$ for the indicator function, $[n]=\{1,\dots,n\}$, $\mathrm{sg}[\cdot]$ for the stop-gradient, and $\mathrm{LN}(\cdot)$ for LayerNorm~\citep{ba2016layer}. The learned embedding matrix $E_\theta\in\mathbb{R}^{|\mathcal{V}|\times d}$, with embedding dimension $d$, maps a sequence $\mathbf{z}$ to $\mathbf{z}E_\theta\in\mathbb{R}^{L\times d}$.
\subsection{Uniform Discrete Diffusion Models (UDMs)}
\label{sec:bg-udm}
DDMs map the data distribution to a simple prior with a fixed \emph{forward} process that gradually corrupts clean tokens, and learn to reverse it~\citep{austin2023structureddenoisingdiffusionmodels, campbell2022continuoustimeframeworkdiscrete}. Most recent works choose as prior either a distribution collapsed onto a special \mask category in Masked Diffusion~\citep{sahoo2024simpleeffectivemaskeddiffusion, shi2025simplifiedgeneralizedmaskeddiffusion, ou2024radd}, or the uniform distribution over the vocabulary, as in UDMs~\citep{hoogeboom2021argmaxflowsmultinomialdiffusion, lou2024discretediffusionmodelingestimating, sahoo2025diffusionduality, schiff2025simpleguidancemechanismsdiscrete}. During sampling, MDMs fixe each token once it is unmasked, whereas UDMs can revise tokens at every step. Given a monotonically decreasing noise schedule $\alpha_t$ from $\alpha_0=1$ to $\alpha_1=0$, the forward process corrupts each token with
\begin{equation}
\label{eq:udm-forward}
q_t(\mathbf{z}_t^{\supl} \mid \mathbf{x}^{\supl}) = \mathrm{Cat}\!\left(\mathbf{z}_t^{\supl};\ \alpha_t\,\mathbf{x}^{\supl} + (1-\alpha_t)\tfrac{1}{|\mathcal{V}|}\mathbf{1}\right).
\end{equation}
Thus, $\mathbf{z}_t^{\supl}$ remains clean with probability $\alpha_t$ and is otherwise drawn uniformly from the vocabulary. The sequence-level forward process applies \Eqn{eq:udm-forward} independently at each position, \mbox{$q_t(\mathbf{z}_t\mid\mathbf{x})=\prod_{\ell=1}^{L}q_t(\mathbf{z}_t^{\supl}\mid\mathbf{x}^{\supl})$}. To reverse the forward process, the denoiser $\mathbf{x}_\theta(\mathbf{z}_t,t)$ learns to predict the clean tokens. It composes the embedding matrix $E_\theta$, a stack of Transformer blocks $f_\theta$ (the \emph{backbone}), the Language Modeling (LM) head $g_\theta$ (an affine projection), and a softmax:
\begin{equation}
\label{eq:denoiser-chain}
\underbrace{\mathbf{z}_t}_{\mathcal{V}^L}
\xmapsto{\;\cdot\,E_\theta\;}
\underbrace{\mathbf{z}_t E_\theta}_{\mathbb{R}^{L\times d}}
\xmapsto{\;f_\theta(\cdot,\,t)\;}
\underbrace{\mathbf{h}_t}_{\mathbb{R}^{L\times d}}
\xmapsto{\;g_\theta\;}
\underbrace{g_\theta(\mathbf{h}_t)}_{\mathbb{R}^{L\times|\mathcal{V}|}}
\xmapsto{\;\mathrm{softmax}\;}
\underbrace{\mathbf{x}_\theta(\mathbf{z}_t,t)}_{(\Delta^{|\mathcal{V}|})^L},
\end{equation}
where $\mathbf{h}_t$ is the final-layer hidden representation. More generally, the backbone acts on any input $\mathbf{u}\in\mathbb{R}^{L\times d}$ in the embedding space, and we write $\mathbf{x}_\theta(\mathbf{u},t;T)=\mathrm{softmax}(g_\theta(f_\theta(\mathbf{u},t))/T)$ for the prediction at temperature $T>0$, omitting $T$ when $T=1$. For token inputs, we use the shorthand \mbox{$\mathbf{x}_\theta(\mathbf{z}_t,t)=\mathbf{x}_\theta(\mathbf{z}_tE_\theta,t)$}. Generation refines the noisy state $\mathbf{z}_t$ from $t=1$ to $t=0$. We write $q$ for the joint law of the data and the forward process, and $q_t$, $q_{s|t}$ for its conditionals. For $s<t$, $q$ induces the categorical \emph{denoising posterior} $q_{s|t}(\mathbf{z}_s^{\supl}\mid\mathbf{z}_t^{\supl},\mathbf{x}^{\supl})$, which depends on the unknown clean token $\mathbf{x}^{\supl}$. A common UDM parameterization, the \emph{bridge plug-in}, draws $\mathbf{z}_s$ from this posterior evaluated at the prediction $\mathbf{x}_\theta^{\supl}(\mathbf{z}_t,t)\in\Delta^{|\mathcal{V}|}$ instead of the one-hot $\mathbf{x}^{\supl}$~\citep{schiff2025simpleguidancemechanismsdiscrete,sahoo2025diffusionduality,gourevitch2026uniformdiffusionmodels} (\supp{app:parameterization_reverse}). The denoiser is trained to minimize the \emph{Negative Evidence Lower Bound} (NELBO), an upper bound on the negative log-likelihood that decomposes into per-token losses $\mathcal{L}_{\mathrm{tok}}$,
\begin{equation}
\label{eq:udm-loss}
\mathbb{E}\!\left[-\log p^\theta(\mathbf{x})\right] \le \mathcal{L}(\theta) = \mathbb{E}_{\mathbf{x}\sim q_{\text{data}},\,t\sim\mathcal{U}[0,1],\,\mathbf{z}_t\sim q_t(\cdot\mid\mathbf{x})}\!\left[\,\sum_{\ell=1}^{L} \mathcal{L}_{\mathrm{tok}}\!\left(\mathbf{x}_\theta^{\supl}(\mathbf{z}_t,t),\, \mathbf{x}^{\supl},\, \mathbf{z}_t^{\supl},\, \alpha_t\right)\right],
\end{equation}
where $p^\theta$ is the model likelihood. \supp{app:nelbo} gives the closed forms of $q_{s|t}$ and $\mathcal{L}_{\mathrm{tok}}$~\citep{schiff2025simpleguidancemechanismsdiscrete, sahoo2025diffusionduality}. To sample, let $t_n=n/N$, $n=0,\dots,N$, be a uniform time grid, where $N$ is the NFE. The initial tokens $\mathbf{z}_{t_N}$ are drawn from the uniform prior $\mathrm{Cat}(\cdot\,;\tfrac{1}{|\mathcal{V}|}\mathbf{1})$, and the backward transition is applied $N$ times (ancestral sampling):
\begin{equation}
\label{eq:udm-sampling}
\mathbf{z}_{t_{n-1}}^{\supl} \sim q_{t_{n-1}|t_n}\!\left(\mathbf{z}_{t_{n-1}}^{\supl} \;\middle|\; \mathbf{z}_{t_n}^{\supl},\ \mathbf{x}_\theta^{\supl}(\mathbf{z}_{t_n},t_n)\right), \qquad n=N,\dots,1.
\end{equation}
\subsection{Loopholing Discrete Diffusion Models (LDDMs)}
\label{sec:bg-loopholing}
The sampling step~\Eqn{eq:udm-sampling} only propagates the discrete tokens $\mathbf{z}_s$ to the next step. Thus, the predictions $\mathbf{x}_\theta(\mathbf{z}_t,t)$ and their uncertainty are discarded after sampling (\emph{information collapse}). LDDMs~\citep{jo2025loopholingdiscretediffusion} address this with SC~\citep{chen2022analogbits}, which feeds the previous predictions or hidden representations of the denoiser back as input. Alongside the sampled tokens, they propagate the \emph{SC state} $\mathbf{c}$, the layer-normalized final-layer hidden representation~\Eqn{eq:denoiser-chain}. At the next step, $\mathbf{c}$ is added to the token embeddings. Starting from $\mathbf{c}=\mathbf{0}$ at $t=1$, one LDDM sampling step from time $t$ to $s$ maps \mbox{$(\mathbf{z}_t,\mathbf{c}_t)$} to $(\mathbf{z}_s,\mathbf{c}_s)$:
\begin{equation}
\label{eq:loopholing}
\underbrace{(\mathbf{z}_t,\,\mathbf{c}_t)}_{\mathcal{V}^L\times\,\mathbb{R}^{L\times d}}
\xmapsto{\;f_\theta(\mathbf{z}_t E_\theta\,+\,\mathbf{c}_t,\ t)\;}
\underbrace{\mathbf{h}_t}_{\mathbb{R}^{L\times d}}
\xmapsto{\;\left(q_{s|t}\,\circ\,\mathrm{softmax}\,\circ\,g_\theta,\ \,\statemap_\theta^{\mathrm{Hid.}}\right)\;}
\underbrace{(\mathbf{z}_s,\,\mathbf{c}_s)}_{\mathcal{V}^L\times\,\mathbb{R}^{L\times d}},
\end{equation}
where $\mathbf{z}_s^{\supl}$ is sampled from the bridge plug-in $q_{s|t}(\cdot\mid\mathbf{z}_t^{\supl},\mathbf{x}_\theta^{\supl}(\mathbf{z}_tE_\theta+\mathbf{c}_t,t))$, as in \Eqn{eq:udm-sampling}, and the \emph{state map} \mbox{$\statemap_\theta^{\mathrm{Hid.}}(\mathbf{h})=\mathrm{LN}(\mathbf{h})$} computes $\mathbf{c}_s=\statemap_\theta^{\mathrm{Hid.}}(\mathbf{h}_t)$. The masked (LDDM-M) and uniform (LDDM-U) variants build on MDLM~\citep{sahoo2024simpleeffectivemaskeddiffusion} and Duo~\citep{sahoo2025diffusionduality}, respectively. During training, with probability $p$, the \emph{multi-pass probability} ($p=0.9$ by default), a first forward pass on $(\mathbf{z}_t,\mathbf{0})$ computes $\mathbf{c}_t=\mathrm{sg}\!\left[\statemap_\theta^{\mathrm{Hid.}}(\mathbf{h}_t)\right]$, and the NELBO~\Eqn{eq:udm-loss} is computed after a second pass on $(\mathbf{z}_t,\mathbf{c}_t)$ only. Otherwise, LDDMs execute a standard UDM training step with $\mathbf{c}_t=\mathbf{0}$. Thus, $\mathbf{c}_t$ is computed at the same time $t$ during training, but comes from the previous, noisier step during sampling. Despite the induced train--test mismatch, the additional contextual information from previous predictions enables LDDMs to outperform the pure discrete baselines~\citep{jo2025loopholingdiscretediffusion}, e.g., on hard Sudoku, LDDM-U solves 91.0\% of the puzzles, against 58.4\% for Duo (\sec{sec:experiments-sudoku}).
\section{Loopholing All the Way (LAW)}
\label{sec:method}
Although LDDMs use SC, they operate on a sequence of discrete tokens, which the SC state only complements (\sec{sec:bg-loopholing}). Since SC can preserve continuous information across steps, we propose \method, a generative model based \emph{purely} on SC. \method propagates \emph{only} the SC state and decodes discrete tokens at the end. As the SC state becomes the only input of the denoiser, it should lie in the same space as the token embeddings that the denoiser is trained on. We therefore use the \emph{Expected Embedding} under the denoiser's prediction, a convex combination of the token embeddings. We present the generative process first (\sec{sec:method-sampling}), since it motivates the training procedure (\sec{sec:method-training}).
\subsection{Generative Process}
\label{sec:method-sampling}
\paragraph{Initialization.}
As for a UDM at $t=1$, we obtain the initial state $\mathbf{c}_{t_N}$ by embedding tokens drawn from the uniform prior:
\begin{equation}
    \label{eq:law-init}
    \mathbf{c}_{t_N}=\mathbf{z}_{t_N}E_\theta,
    \qquad
    \mathbf{z}_{t_N}^{\supl}
    \sim
    \mathrm{Cat}\!\left(\cdot\,;\frac{1}{|\mathcal{V}|}\mathbf{1}\right).
\end{equation}
\paragraph{Continuous State Updates.}
For $n=N,\ldots,2$, we pass the SC state $\mathbf{c}_{t_n}$ through the denoising backbone and map the resulting hidden representation into the next state:
\begin{equation}
    \label{eq:law-transition}
    \underbrace{\mathbf{c}_{t_n}}_{\mathbb{R}^{L\times d}}
    \xmapsto{\;f_\theta(\cdot,\,t_n)\;}
    \underbrace{\mathbf{h}_{t_n}}_{\mathbb{R}^{L\times d}}
    \xmapsto{\;\statemap_\theta^{\mathrm{E.E.}}(\cdot\,;\,T)\;}
    \underbrace{\mathbf{c}_{t_{n-1}}}_{\mathbb{R}^{L\times d}}.
\end{equation}
Thus, the backbone receives the SC state alone, and no tokens are sampled between steps. The state map computes the \emph{Expected Embedding} (E.E.) under the denoiser's prediction:
\begin{equation}
    \label{eq:law-state-map}
    \underbrace{\statemap_\theta^{\mathrm{E.E.}}(\mathbf{h};T)}
                _{\mathbb{R}^{L \times d}}
    =
    \underbrace{
        \operatorname{softmax}\!\left(
            \frac{g_\theta(\mathbf{h})}{T}
        \right)
    }_{\mathbb{R}^{L \times |\mathcal{V}|}}
    \underbrace{E_\theta}
                _{\mathbb{R}^{|\mathcal{V}| \times d}},
\end{equation}
where the softmax is applied over the vocabulary dimension and $T>0$ is the temperature. LDDMs add their SC state $\statemap_\theta^{\mathrm{Hid.}}(\mathbf{h})$ to the token embeddings~\Eqn{eq:loopholing}, whereas the SC state of \method \emph{replaces} them. We compare both state maps in \sec{sec:experiments-gsm8k}.
During sampling, $T$ is the fixed sampling temperature ($T=1$ unless stated otherwise). Since $T$ is randomized during training (\sec{sec:method-training}), the denoiser learns to process tempered SC states.
For conditional generation, we always fix the prompt positions to the clean token embeddings.
\paragraph{Decoding.}
At the last step, a decoder $\eta$ maps the denoiser prediction $\hat{\mathbf{x}}=\mathbf{x}_\theta(\mathbf{c}_{t_1},t_1;T)$ to one-hot tokens $\eta(\hat{\mathbf{x}})\in\mathcal{V}^L$. We consider the \emph{raw} and \emph{argmax} decoders:
\begin{equation}
    \label{eq:law-decoding}
    \eta_{\mathrm{raw}}(\hat{\mathbf{x}})^{\supl}
    \sim
    \mathrm{Cat}\!\left(\cdot\,;\hat{\mathbf{x}}^{\supl}\right),
    \qquad
    \eta_{\mathrm{max}}(\hat{\mathbf{x}})^{\supl}
    =
    \arg\max_{\mathbf{v}\in\mathcal{V}}\,\langle\mathbf{v},\hat{\mathbf{x}}^{\supl}\rangle.
\end{equation}
With the argmax decoder, all the randomness comes from the prior, so sampling is deterministic given $\mathbf{z}_{t_N}$. See \algo{alg:law-sampling} for the pseudocode.
\subsection{Training the Continuous Trajectory}
\label{sec:method-training}
\paragraph{Multi-Pass Training.}
During sampling, the denoiser consumes the SC states that it produced itself~\Eqn{eq:law-transition}. Since backpropagation through the whole trajectory is costly, we adopt the multi-pass training of LDDM-U (\sec{sec:bg-loopholing}). We first sample a corrupted sequence $\mathbf{z}_t$~\Eqn{eq:udm-forward}. With probability $p$, we unroll $M$ forward passes, and otherwise, we execute a single pass on $\mathbf{z}_t$, as in standard UDM training. In the multi-pass case, the first pass receives the token embeddings of $\mathbf{z}_t$, while later ones process the SC state, and we only backpropagate through the last pass. We use $p=0.9$ and $M=2$ by default, following \citet{jo2025loopholingdiscretediffusion}. Formally,
\begin{equation}
    \label{eq:law-unroll}
    \mathbf{c}^{(1)} = \mathbf{z}_t E_\theta,
    \qquad
    \mathbf{c}^{(m+1)} = \mathrm{sg}\!\left[\statemap_\theta^{\mathrm{E.E.}}\!\big(f_\theta(\mathbf{c}^{(m)}, t^{(m)});\, T\big)\right],
    \qquad
    t^{(m)} = t - (m-1)\,\delta,
\end{equation}
where the superscript $(m)$ indexes the pass and $\delta=1/L$ is the step size when sampling with $N=L$ NFE. The loss is applied only to the prediction of the last pass, \mbox{$\hat{\mathbf{x}}=\mathbf{x}_\theta(\mathbf{c}^{(M)}, t^{(M)})$}. We minimize either the per-token NELBO $\mathcal{L}_{\mathrm{tok}}$~\Eqn{eq:udm-loss} or the Cross Entropy (CE):
\begin{equation}
    \label{eq:law-loss}
    \mathcal{L}_{\mathrm{LAW}}(\theta)
    =
    \mathbb{E}_{\mathbf{x},\,t,\,\mathbf{z}_t,\,T}
    \left[
    \sum_{\ell=1}^{L}
    \mathcal{L}_{\mathrm{LAW}}^{\supl}
    \right],
    \qquad
    \mathcal{L}_{\mathrm{LAW}}^{\supl}
    =
    \begin{cases}
    \mathcal{L}_{\mathrm{tok}}\!\left(\hat{\mathbf{x}}^{\supl},\, \mathbf{x}^{\supl},\, \mathbf{z}_t^{\supl},\, \alpha_t\right) & \text{(NELBO)},\\[2pt]
    -\log\langle \hat{\mathbf{x}}^{\supl},\, \mathbf{x}^{\supl}\rangle & \text{(CE)}.
    \end{cases}
\end{equation}
During the multi-pass updates, we sample $t\sim\mathcal{U}[(M-1)\delta, 1]$ to ensure that $t^{(M)} \geq 0$, while single-pass updates draw $t\sim\mathcal{U}[0,1]$. See \algo{alg:law-training} for the pseudocode.
\paragraph{Temperature Regularization.}
Sampling with $T\neq1$ makes the denoiser's logits sharper ($T<1$) or flatter ($T>1$). To expose the denoiser to SC states computed with tempered logits, we draw the temperature $T$ at random during training, to perturb the SC states~\Eqn{eq:law-unroll}. We compute the loss with the untempered prediction $\hat{\mathbf{x}}$~\Eqn{eq:law-loss}, so $T$ only affects the SC states. We draw $T$ as
\begin{equation}
    \label{eq:law-temperature}
    T \sim \tfrac{1}{2}\,\mathcal{U}\!\left[\rho^{-1},1\right]
            + \tfrac{1}{2}\,\mathcal{U}\!\left[1,\rho\right],
\end{equation}
and share it across passes and across tokens in the sequence.
The factor $\rho\geq1$ makes it equally likely to sharpen ($T<1$) or flatten ($T>1$) the logits. We sample both sides because, with a one-sided range, the LM head $g_\theta$ could absorb the temperature by rescaling its weights.
\paragraph{Architecture.} 
\method uses the same denoiser architecture as standard UDMs, a Diffusion Transformer (DiT) with pre-normalization~\citep{peebles2023scalablediffusionmodelstransformers, lou2024discretediffusionmodelingestimating}, so it can be trained from scratch or fine-tuned from a pre-trained UDM checkpoint. We optionally add a learned \emph{mode embedding} to the time conditioning, which indicates whether the denoiser receives the embeddings of noisy tokens or an SC state. We only use mode embedding on OpenWebText, where it prevents the representations from collapsing when $\rho=3$ (\supp{app:owt-ablations}). See \supp{app:mode-emb} for details.
\section{Experiments}
\label{sec:experiments}
We evaluate \method on four datasets: Sudoku (\sec{sec:experiments-sudoku}) to test constraint satisfaction with a unique solution, Maze (\sec{sec:experiments-maze}) for long-range planning with possibly several optimal paths, GSM8K \citep{cobbe2021trainingverifierssolvemath} (\sec{sec:experiments-gsm8k}) for reasoning on grade-school math word problems, and OpenWebText \citep{Gokaslan2019OpenWeb} (\sec{sec:experiments-owt}) for unconditional language modeling.
\paragraph{Baselines and Training.}
\begin{wraptable}{r}{0.47\linewidth}
    \centering
    \footnotesize
    \setlength{\tabcolsep}{3pt}
    \vspace{-12pt}
    \renewcommand{\arraystretch}{0.95}
    \caption{\textbf{\emph{Exact-match} accuracy (\%) on Sudoku with 180 NFE.} Baseline numbers are taken from prior work \citep{deschenaux2026hypersphericalflows}, except LDDMs, which we train with matched training FLOPs. $p$ is the multi-pass probability, and reg.\ is temperature regularization with range $\rho$~\Eqn{eq:law-temperature}. Best per column in \textbf{bold}, second best \underline{underlined}.}
    \label{tab:sudoku}
    \newcommand{\tabrow}{\hspace*{0.8em}}
    \begin{tabular}{lccc}
        \toprule
        Model & Easy & Med. & Hard \\
        \midrule
        \multicolumn{4}{@{}l@{}}{\textit{Autoregressive}} \\
        \tabrow Sample & 13.9 & 5.1 & 0.6 \\
        \tabrow Greedy & 14.6 & 5.1 & 1.0 \\
        \midrule
        \multicolumn{4}{@{}l@{}}{\textit{Discrete Diffusion}} \\
        \tabrow MDLM {\tiny\citep{sahoo2024simpleeffectivemaskeddiffusion}} & 92.0 & 77.1 & 30.2 \\
        \tabrow Duo {\tiny\citep{sahoo2025diffusionduality}} & 96.3 & 84.7 & 58.4 \\
        \tabrow LDDM-M {\tiny\citep{jo2025loopholingdiscretediffusion}} & 98.3 & 95.9 & 87.5 \\
        \tabrow LDDM-U {\tiny\citep{jo2025loopholingdiscretediffusion}} & \underline{99.8} & 97.9 & 91.0 \\
        \midrule
        \multicolumn{4}{@{}l@{}}{\textit{Continuous Models}} \\
        \tabrow CANDI {\tiny\citep{pynadath2025candihybriddiscretecontinuousdiffusion}} & 79.3 & 45.9 & 16.7 \\
        \tabrow CANDI {\tiny pure Gaussian} & 63.9 & 41.9 & 12.5 \\
        \tabrow FLM {\tiny\citep{lee2026onesteplanguagemodelingcontinuous}} & 94.2 & 82.7 & 44.5 \\
        \tabrow $\mathbb{S}$-FLM {\tiny Deschenaux et al.\ \citeyearpar{deschenaux2026hypersphericalflows}} & 94.8 & 85.2 & 45.0 \\
        \midrule
        \rowcolor{gray!15}
            \tabrow \method{} {\tiny NELBO no reg. ($p = 0.9$)} & 98.7 & 94.6 & 85.3 \\
        \rowcolor{gray!15}
        \tabrow \method{} {\tiny CE no reg. ($p = 0.9$)} & \underline{99.8} & \underline{99.0} & 92.2 \\
        \rowcolor{gray!15}
        \tabrow \method{} {\tiny CE no reg. ($p = 0.75$)} & \textbf{99.9} & 98.9 & \underline{94.5} \\
        \rowcolor{gray!15}
        \tabrow \method{} {\tiny CE reg. ($\rho = 2$, $p = 0.9$)} & \underline{99.8} & \textbf{99.4} & \textbf{96.5} \\
        \bottomrule
    \end{tabular}
    \vspace{-40pt}
\end{wraptable}
We compare to an ARM, to the DDMs MDLM~\citep{sahoo2024simpleeffectivemaskeddiffusion}, Duo~\citep{sahoo2025diffusionduality}, LDDM-M and LDDM-U~\citep{jo2025loopholingdiscretediffusion}, and to the continuous models CANDI~\citep{pynadath2025candihybriddiscretecontinuousdiffusion}, FLM~\citep{lee2026onesteplanguagemodelingcontinuous} and $\mathbb{S}$-FLM~\citep{deschenaux2026hypersphericalflows}. Unless stated otherwise, all models are DiTs trained from scratch. Since LDDMs and \method unroll several forward passes, we train those for fewer steps to use similar training FLOPs as the baselines (\supp{app:compute}). Additionally, \method always uses the raw decoder $\eta_{\mathrm{raw}}$~\Eqn{eq:law-decoding}. Full details are in \supp{app:exp-details}.
\subsection{Reasoning on Sudoku}
\label{sec:experiments-sudoku}
\paragraph{Experimental Setup.}
We follow prior work \citep{kim2025finetuningmaskeddiffusionprovable} and generate 48k training and 2k validation puzzles with unique solutions \citep{alp2024sudoku}. We consider three difficulty levels, determined by the number of visible digits (easy: $40/81$, medium: $35/81$, hard: $30/81$), and we report the \emph{exact-match} accuracy. Each sequence concatenates the partial grid with the solution grid. The denoiser is an 8-layer DiT with Rotary Positional Encoding (RoPE; \citealp{su2023roformerenhancedtransformerrotary}). The baselines train for 20k steps. See \supp{app:sudoku-details} for details and \supp{app:sudoku-ablations} for ablations.
\paragraph{Results.} 
With 180 NFE, \method is the most accurate model on all three difficulties (\tab{tab:sudoku}). In particular, \method outperforms LDDMs. \method solves 96.5\% of the puzzles with temperature regularization ($\rho=2$, $p=0.9$) and training with CE performs better than with the NELBO (\supp{app:sudoku-ablations}).
\subsection{Path Finding on Maze}
\label{sec:experiments-maze}
\begin{figure}[t]
    \centering
    \includegraphics[width=\linewidth]{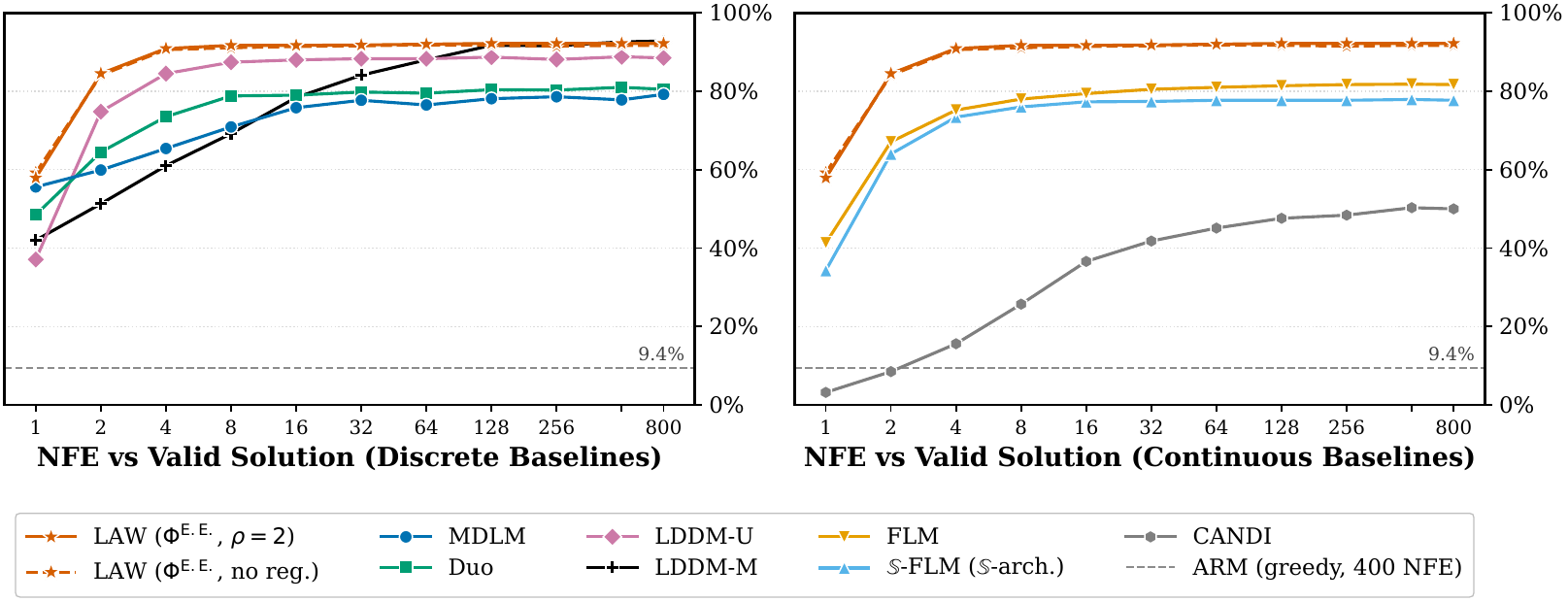}
    \caption{\textbf{Maze \emph{valid-solution} accuracy against NFE.}
    Fraction (\%) of the 1k evaluation mazes for which a valid solution is generated (\supp{app:maze-metrics}). All methods are trained with matched training FLOPs (\supp{app:compute}). \method uses $M=2$, $p=0.75$, the NELBO, batch size 256 and 80k steps, with ($\rho=2$, solid) and without (dashed) temperature regularization. The dashed gray line is the ARM with greedy decoding (9.4\%). \textbf{Left:} \method against the DDM baselines. \textbf{Right:} \method against the continuous baselines.
    }
    \label{fig:maze-nfe}
\end{figure}

\paragraph{Experimental Setup.} 
We evaluate shortest-path finding on mazes inspired by \citet{wang2025hierarchicalreasoning}. We generate 1k training and 1k evaluation mazes on $20\times20$ grids~\citep{lehnert2024abetterplanningtransformers} and keep mazes with long-enough shortest path (71 steps). Each cell is encoded as a \texttt{wall}, \texttt{empty}, \texttt{start}, \texttt{goal}, or \texttt{path} token. As for Sudoku, we concatenate the unsolved grid with the solution (800 tokens).
We report the accuracy of finding \emph{any of} the shortest paths from \texttt{start} to \texttt{goal} (unlike certain work that measure exact match with a particular shortest path). The denoiser is an 8-layer DiT ($d=256$, 8 heads, RoPE), and we train the baselines for 100k steps.
\paragraph{Results.}
\method is the most accurate model with few steps (\Fig{fig:maze-nfe}). With only 4 NFE, it solves 90.9\% of the mazes, more than any baseline with any number of steps besides LDDM-M, and it is ahead of every baseline up to 64 NFE. From 512 NFE, LDDM-M is slightly more accurate, as both models solve about 92\% of the mazes. LDDM-U plateaus below 89\%, and the continuous baselines below 82\%. Temperature regularization barely changes the accuracy and training with the NELBO is generally better than with CE. Details on the ablations on the loss, $M$ and $p$ are in \supp{app:maze-ablations}.
\subsection{Reasoning on GSM8K}
\label{sec:experiments-gsm8k}
\paragraph{Experimental Setup.}
We train on TinyGSM \citep{liu2023tinygsm}, a dataset of 11.8M synthetic math problems, and evaluate on the GSM8K test set \citep{cobbe2021trainingverifierssolvemath}. For each test problem, we generate and execute one Python program. We use a 12-layer DiT denoiser and the \texttt{smollm} tokenizer \citep{allal2025smollm2smolgoesbig}, with a context length of 512 and padding for shorter programs. The baselines train for 250k steps with a batch size of 512, while LDDMs and \method train for 192k steps (\supp{app:flop-matching}). Unless stated otherwise, \method uses the \emph{Expected Embedding}, the NELBO, $M=2$ passes with $p=0.9$, and temperature regularization with $\rho=3$. We report the GSM8K accuracy from 1 to 1024 NFE at the sampling temperatures $T=1$ and $T=0.1$. See~\supp{app:gsm8k-details} for details and \supp{app:tinygsm-ablations} for all ablations.
\paragraph{Results.}
Trained from scratch, \method solves 42.2\% of the problems at $T=1$ with 1024 NFE, against 34.5\% for the strongest diffusion or flow baseline, LDDM-U, and it outperforms all of them from 32 NFE onward, with their standard samplers (\Fig{fig:gsm8k-nfe}, \supp{app:baselines}). At $T=0.1$, it is the most accurate model from 4 to 512 NFE (e.g., 26.4\% against 12.4\% for LDDM-U with 8 NFE), and LDDMs match it only with 1024 NFE. \tab{tab:app-tinygsm-few-step} shows that with $T=0.1$ and 2 to 64 NFE, \method is more accurate than the recent few-step approaches FMLM+~\citep{agarwal2026posterior}, DBTM~\citep{tang2026dbtm}, IDLM~\citep{li2026idlm} and the undistilled SDM~\citep{deschenaux2026simplex}.
\begin{figure}[t]
    \centering
    \includegraphics[width=0.48\linewidth]{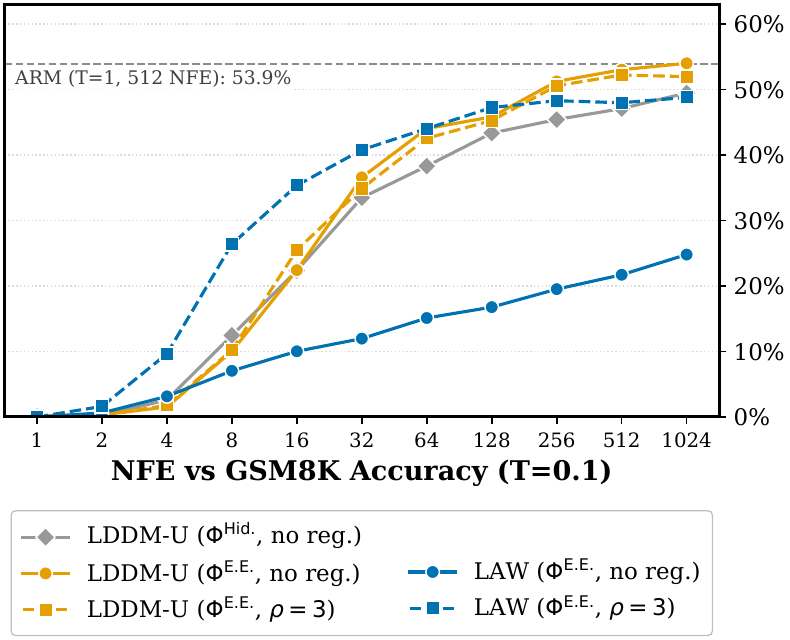}
    \hfill
    \includegraphics[width=0.48\linewidth,trim=0 -8.08 0 0]{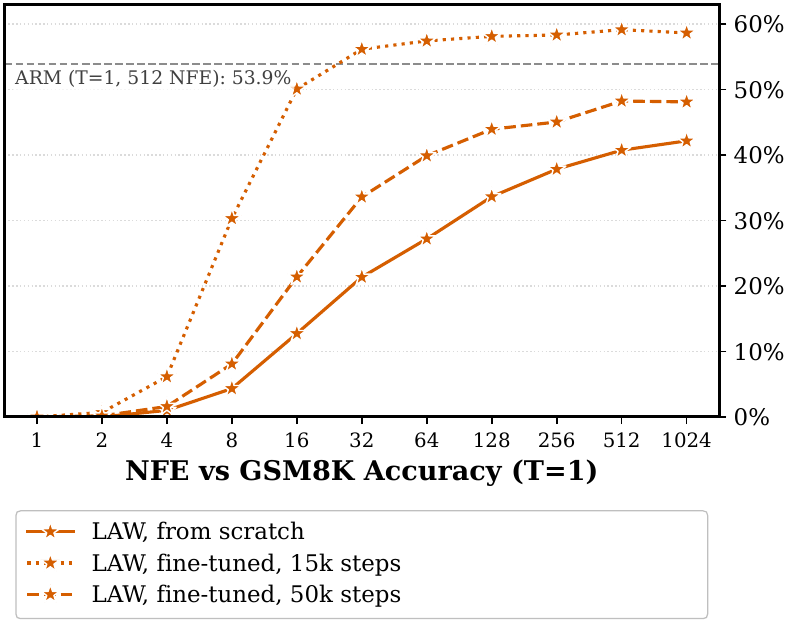}
    \caption{\textbf{Ablations on GSM8K: how \method uses the SC state (left, $T=0.1$) and fine-tuning from Duo (right, $T=1$).} \textbf{Left:} the legend gives the SC state and the temperature regularization ($\rho=3$ or none). \textbf{Right:} \method fine-tuned from Duo for 15k and 50k steps, with $M=2$, $p=0.9$, $\rho=3$ and batch size 512, against \method trained from scratch.}
    \label{fig:gsm8k-ablations}
\end{figure}
\paragraph{Fine-Tuning from Duo.}
We show that \method can be adapted from a pre-trained Duo checkpoint. We fine-tune one for up to 50k steps, with $M=2$ and $p=0.9$ or $M=4$ and $p=0.75$, and batch sizes from 64 to 512 (\Fig{fig:app-gsm-finetune}). With batch size 512 and after 50k steps, \method solves 48.1\% of the problems at $T=1$ with 1024 NFE, against 17.2\% for Duo and 42.2\% for \method trained from scratch (right panel of \Fig{fig:gsm8k-ablations}). This fine-tuning adds 26\% to the training compute of Duo. With $M=2$, the accuracy peaks after 15k to 20k steps and then decreases, whereas with $M=4$ it does not decrease. After 15k steps, which add 8\% to the compute of Duo, \method is more accurate than the ARM at $T=1$ in 32 sampling steps (56.2\% vs.\ 53.9\% for the ARM with 512 NFE). When trained from scratch, the accuracy of \method plateaus from about 70k of the 192k steps (\Fig{fig:app-gsm-backprop-steps}), so additional updates alone do not explain the higher accuracy of the fine-tuned \method. We hypothesize that the representations learned during Duo pre-training ease the adaptation to \method, and leave testing this to future work. Fine-tuned in the same way, LDDM-U adapts more slowly, and hard-token SC states are less accurate (\supp{app:tinygsm-hard-tokens}).
\paragraph{Comparison with AlphaDLM.}
The concurrent work AlphaDLM~\citep{gushchin2026alphadlm} is the most accurate method with up to 8 NFE, and is more accurate than \method trained from scratch. Fine-tuned from Duo for 15k steps, however, \method is more accurate than AlphaDLM from 16 NFE onward at $T=1$, for about 8\% more training compute (250k Duo steps plus 15k fine-tuning steps, against 250k for AlphaDLM). Two differences might explain the higher accuracy of AlphaDLM. First, it uses a confidence-based sampler, which alone increases the accuracy of MDLM from 22.1\% to 43.9\% at 32 NFE. Second, it trades the many-steps performance for higher few-steps accuracy, and its hyperparameter $k$, which sets this trade-off, must be chosen for a target inference budget (\tab{tab:app-tinygsm-few-step}). Distillation shows a similar trade-off, as the distilled SDM plateaus below the original checkpoint with many steps~\citep{deschenaux2026simplex}. We therefore see AlphaDLM as complementary to \method, and leave studying their combination for future work.
\paragraph{Ablations.}
Three design choices are essential (\supp{app:tinygsm-ablations}). First, the state map $\statemap_\theta^{\mathrm{E.E.}}$ performs better than $\statemap_\theta^{\mathrm{Hid.}}$. With $\statemap_\theta^{\mathrm{Hid.}}$, \method solves fewer than 1.5\% of the problems, against 35.8\% with $\statemap_\theta^{\mathrm{E.E.}}$ (\Fig{fig:app-gsm-state-loss}; no regularization). Second, temperature regularization doubles the accuracy at $T=0.1$, from 24.8\% to 48.8\% with 1024 NFE (\Fig{fig:app-gsm-temp-mode}). Third, backpropagating through the $M$ passes decreases the accuracy from 42.2\% to 20.2\% at both matched training FLOPs, \emph{and at equal training steps} (Figures~\ref{fig:app-gsm-backprop} and~\ref{fig:app-gsm-backprop-steps}). Find ablations on the number of passes and $p$ in \supp{app:tinygsm-ablations}.
\paragraph{Pure vs.\ Auxiliary SC.}
To isolate the choice of state map against the effect of pure Self-Conditioning, we train a variant of LDDM-U with $\statemap_\theta^{\mathrm{E.E.}}$ instead of $\statemap_\theta^{\mathrm{Hid.}}$. Using $\statemap_\theta^{\mathrm{E.E.}}$ improves the accuracy (38.7\% instead of 34.5\% at $T=1$ with 1024 NFE). Nevertheless, even with temperature regularization, LDDM-U with $\statemap_\theta^{\mathrm{E.E.}}$ remains less accurate than \method, except at $T=0.1$ with 256 NFE or more (\Fig{fig:gsm8k-ablations}; left), where LDDM slightly ouperforms \method. Moreover, LDDMs solve almost no problems when their token embeddings are set to zero, so they cannot generate from the SC state alone (\supp{app:tinygsm-lddm-tokens} and~\ref{app:loophole-comparison}).
\subsection{Language Modeling on OpenWebText}
\label{sec:experiments-owt}
\paragraph{Experimental Setup.}
We train on OpenWebText~\citep{Gokaslan2019OpenWeb} with the GPT-2 tokenizer~\citep{Radford2019LanguageMA} and a context length of 1024, global batch size 512 with the optimizer configuration in \supp{app:training-hparams}. In addition to previous baselines, we add $\mathbb{S}$-FLM with the Sphere-DiT architecture ($\mathbb{S}$-arch.;~\citealp{deschenaux2026hypersphericalflows}). The baselines train for 400k steps, and LDDMs and \method, with $M=2$ passes and $p=0.9$~\Eqn{eq:law-unroll}, for 308k steps (\supp{app:flop-matching}). We use 12-layer DiTs with embedding dimension $d=768$ and 12 attention heads. We show \method trained with the CE~\Eqn{eq:law-loss} (which gives more diverse samples than the NELBO), the mode embedding (\supp{app:mode-emb}), and temperature regularization with $\rho=3$~\Eqn{eq:law-temperature}, and we ablate these choices in \supp{app:owt-ablations}.
\paragraph{Gen.\ PPL / Entropy Frontier.}
Repetitive samples can lower the Gen.\ PPL without improving sample quality \citep{dieleman2022continuousdiffusioncategoricaldata, velickovic2026perplexitycannotalwaystellright}, and the best sampling temperature differs across models. We therefore sweep the sampling temperature $T$ and plot the Gen.\ PPL against the unigram entropy of the samples for each model and NFE \citep{pynadath2025candihybriddiscretecontinuousdiffusion}. A model is considered better if it reaches a lower Gen.\ PPL at matched entropy (\supp{app:genppl-entropy}).
\begin{figure}[t]
    \centering
    \includegraphics[width=\linewidth]{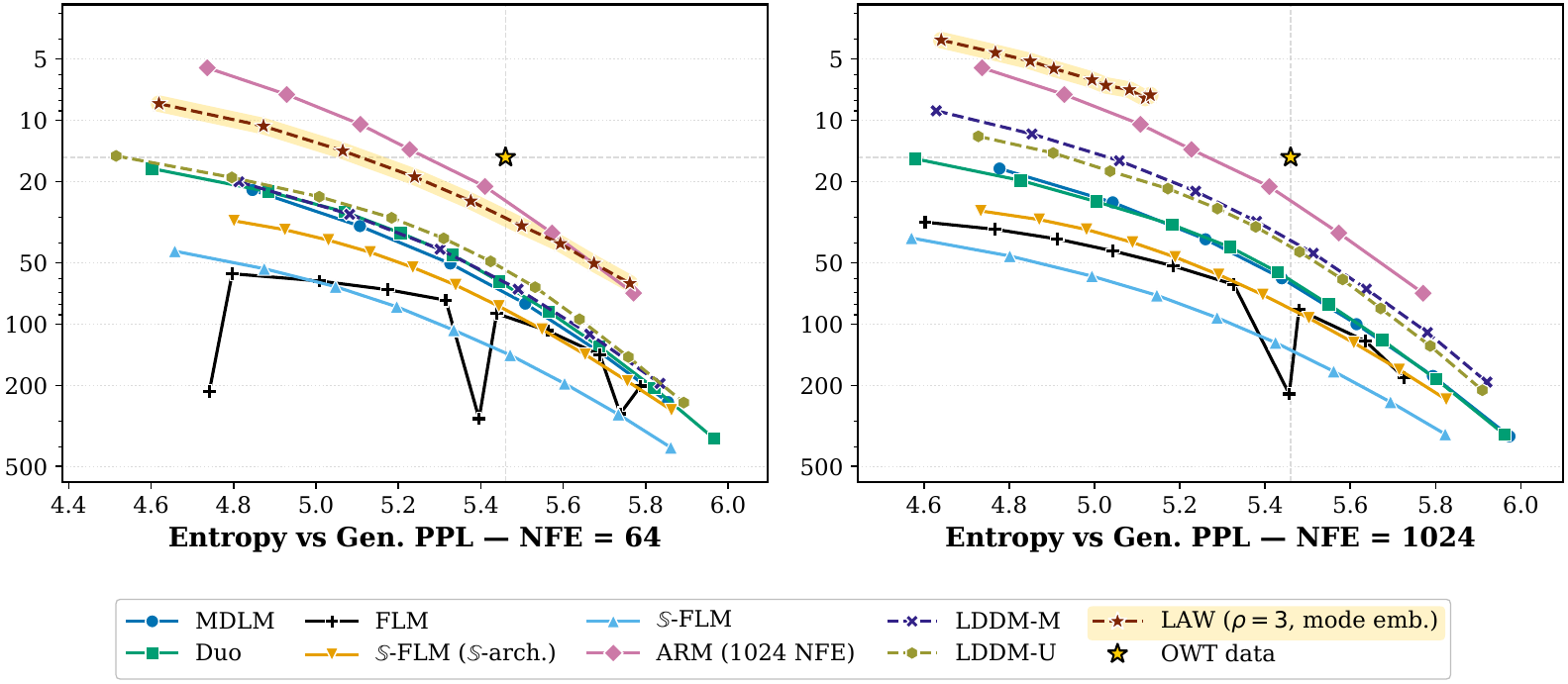}
    \caption{\textbf{Gen.\ PPL against unigram entropy on OpenWebText with 64 (left) and 1024 NFE (right).} Each point is a sampling temperature $T\in\{0.70, 0.75, \dots, 1.20\}$. We sample the ARM with 1024 NFE, but show it in both panels for easier comparison with \method. With 64 NFE, \method dominates all diffusion and flow baselines and comes close to the ARM. With 1024 NFE, it dominates the ARM up to an entropy of 5.1--5.2. The star marks real text, i.e., the Gen.\ PPL and unigram entropy of the 100k documents of the OpenWebText validation split, computed with the same pipeline as the models.}
    \label{fig:owt_entropy_x_gen_ppl}
\end{figure}
\paragraph{Results.}
With 64 NFE, \method outperforms the diffusion and flow baselines and comes close to the ARM that uses 1024 NFEs (\Fig{fig:owt_entropy_x_gen_ppl}). With 1024 NFE, \method also outperforms the ARM at the same entropy, but its samples are less diverse. Over the same sampling temperatures, its entropy reaches at most 5.1--5.2 with 1024 NFE, against 5.8 with 64 NFE. The OpenWebText validation set (star) has a Gen.\ PPL of 15.2 and a unigram entropy of 5.46, which \method reaches with 64 NFE but not with 1024 NFE. At $T=1$ with 1024 NFE, \method has the lowest Gradient Moment (GM;~\citealp{hoogeboom2026discretemmd}) among the diffusion and flow models. With 64 NFE, LDDM-U has the lowest GM (0.721, against 1.210 for \method, \supp{app:owt-gm}).
\paragraph{Ablations.}
As on GSM8K, temperature regularization is essential. Without it, \method only reaches low entropies, at a higher Gen.\ PPL than LDDM-U (\supp{app:owt-frontier-lddm}). With $\rho=3$, the mode embedding also prevents the representations from collapsing midway through training (\supp{app:owt-ablations}). Unlike on GSM8K, the Expected Embedding increases the Gen.\ PPL of LDDM-U.
\section{Related Work}
\label{sec:related-work}
\paragraph{Self-Conditioning.}
SC feeds the previous predictions or hidden representations of the denoiser back as input~\citep{strudel2022selfconditionedembeddingdiffusion, chen2022analogbits}. In DDMs, several methods pass such a continuous state between denoising steps. LDDMs~\citep{jo2025loopholingdiscretediffusion} and DiffusionGemma~\citep{diffusiongemma2026} propagate a continuous state, Soft-Masked Diffusion~\citep{hersche2026softmasked} feeds masked positions a convex combination of the mask embedding and the top-$k$ most confident token embeddings, and Relay~\citep{rozonoyer2026relayrepresentations} and PUMBA~\citep{madeira2026trajectoryaware} pass the last hidden state between the steps of teacher-forced unmasking trajectories. Other methods unroll the denoiser during training. SUNDAE~\citep{savinov2022stepunrolleddenoisingautoencoderstext} unrolls it over discrete tokens, and FReDA~\citep{sun2026freda} refines the discrete drafts of the model without a forward corruption process and, like \method, backpropagates only through the last pass. Relay and PUMBA instead backpropagate through the propagated state and find that it outperforms a stop-gradient, whereas in \method, the stop-gradient roughly doubles the GSM8K accuracy (\supp{app:tinygsm-ablations}). All of these methods still sample discrete tokens between steps and use any continuous state as an auxiliary input. \method propagates only the SC state, at all positions, and decodes once, after the final step (\supp{app:loophole-comparison}).
\paragraph{Continuous Language Models.}
Continuous DLMs and Flow Language Models generate by integrating an ODE or SDE. They corrupt token embeddings or one-hot vectors with Gaussian noise~\citep{li2022diffusionlmimprovescontrollabletext, gulrajani2023likelihoodbaseddiffusionlanguagemodels, chen2026langflowcontinuousdiffusion, hu2026elfembeddedlanguageflows, lee2026onesteplanguagemodelingcontinuous, potaptchik2026discreteflowmaps, roos2026categoricalflowmaps, yoo2026selfconditionedflowmaplanguage} or operate on the hypersphere~\citep{chemseddine2026sphericalflowscategorical, deschenaux2026hypersphericalflows}. Like these models, \method has a continuous sampling trajectory. However, it is trained with the uniform discrete corruption of UDMs and does not use Gaussian noise nor an ODE solver.
\paragraph{Looped Models.}
Looped DLMs reuse layers of the denoiser to vary its depth at inference~\citep{carballocastro2026recursivescaling, lee2026loopeddiffusion, miele2026fixedpointmasked}. These works change the computation within a denoising step, whereas \method changes what is passed between steps, so the two are complementary. Concurrently, Looped Flows~\citep{suleymanzade2026thinkingloopedflows} and Flow Reasoning Models~\citep{helbling2026flowreasoningmodelsturning} pass a recurrent state between the sampling steps of a continuous flow, and evaluate on puzzles such as Sudoku and Maze. We instead focus on language modeling, with math problems solved as code (TinyGSM) and OpenWebText.
\section{Conclusion and Limitations}
\label{sec:limitations}
We ask whether a generative model can be based purely on SC, and we propose \method. Starting from random tokens, \method propagates only the SC state, the \emph{Expected Embedding} under the denoiser's prediction, and decodes once, after the final step. Our results suggest that pure SC can match or outperform SC as an auxiliary input. \method is the most accurate model on Sudoku and on Maze with few steps, and on GSM8K, it outperforms the diffusion and flow baselines trained with the same FLOPs. Fine-tuning a pre-trained Duo checkpoint with \method for 50k steps increases its GSM8K accuracy with 1024 NFE from 17.2\% to 48.1\% at $T=1$, and from 36.0\% to 54.0\% at $T=0.1$. After 15k fine-tuning steps, \method reaches 58.7\% and 53.1\%, respectively, and is more accurate than the ARM at $T=1$ (53.9\%). On OpenWebText, it also outperforms the diffusion and flow baselines at matched entropy and comes close to the ARM with 64 NFE. The ablations identify three necessary choices: the \emph{Expected Embedding} as state map, temperature regularization, and a stop-gradient between passes, as in LDDMs.
\paragraph{Limitations.}
The NELBO is more accurate than the CE on GSM8K and Maze, whereas the CE is more accurate on Sudoku and gives more diverse samples on OpenWebText. Since the NELBO is derived for discrete diffusion processes, \method does not completely depart from the discrete diffusion formalism. With 1024 NFE on OpenWebText, \method's samples cover a narrower entropy range than those of the baselines. Finally, we experiment in the standard academic setting. Whether our results hold at larger scale is left for future work.
\newpage
\subsection*{AI Use Statement}
We acknowledge the use of Large Language Models (LLMs) during the preparation of the manuscript. LLMs assisted in the writing of the paper and code, and certain experiments were submitted using agentic tools. The main ideas were proposed without the use of LLMs. We take full responsibility for the content, claims, citations and conclusions. The TinyGSM dataset is fully generated by GPT-3.5 (as explained by \citealp{liu2023tinygsm}).
\bibliography{iclr2027_conference}

@inproceedings{savinov2022stepunrolleddenoisingautoencoderstext,
      title={Step-unrolled Denoising Autoencoders for Text Generation},
      author={Nikolay Savinov and Junyoung Chung and Mikolaj Binkowski and Erich Elsen and Aaron van den Oord},
      booktitle={International Conference on Learning Representations},
      year={2022},
}

@inproceedings{hoogeboom2021argmaxflowsmultinomialdiffusion,
      title={Argmax Flows and Multinomial Diffusion: Learning Categorical Distributions},
      author={Emiel Hoogeboom and Didrik Nielsen and Priyank Jaini and Patrick Forré and Max Welling},
      booktitle={Advances in Neural Information Processing Systems},
      year={2021},
}

@misc{yoo2026selfconditionedflowmaplanguage,
      title={Self-conditioned Flow Map Language Models via Fixed-point Flows}, 
      author={Jaehoon Yoo and Wonjung Kim and Floor Eijkelboom and Chanhyuk Lee and Nicholas M. Boffi and Seunghoon Hong and Jinwoo Kim},
      year={2026},
      eprint={2607.00714},
      archivePrefix={arXiv},
      primaryClass={cs.CL},
      url={https://arxiv.org/abs/2607.00714}, 
}

@misc{suleymanzade2026thinkingloopedflows,
      title={Thinking with Looped Flows}, 
      author={Ayhan Suleymanzade and Chanhyuk Lee and Floor Eijkelboom and Nicholas M. Boffi and İsmail İlkan Ceylan and Jinwoo Kim},
      year={2026},
      eprint={2609.11801},
      archivePrefix={arXiv},
      primaryClass={cs.LG},
      url={https://arxiv.org/abs/2609.11801}, 
}

@misc{geminiteam2025gemini,
      title={Gemini: A Family of Highly Capable Multimodal Models},
      author={{Gemini Team} and Rohan Anil and Sebastian Borgeaud and others},
      year={2023},
      eprint={2312.11805},
      archivePrefix={arXiv},
      primaryClass={cs.CL},
      url={https://arxiv.org/abs/2312.11805},
}

@inproceedings{lipman2023flowmatchinggenerativemodeling,
      title={Flow Matching for Generative Modeling},
      author={Yaron Lipman and Ricky T. Q. Chen and Heli Ben-Hamu and Maximilian Nickel and Matt Le},
      booktitle={International Conference on Learning Representations},
      year={2023},
}

@inproceedings{albergo2023buildingnormalizingflowsstochastic,
      title={Building Normalizing Flows with Stochastic Interpolants},
      author={Michael S. Albergo and Eric Vanden-Eijnden},
      booktitle={International Conference on Learning Representations},
      year={2023},
}

@inproceedings{liu2022flowstraightfastlearning,
      title={Flow Straight and Fast: Learning to Generate and Transfer Data with Rectified Flow},
      author={Xingchao Liu and Chengyue Gong and Qiang Liu},
      booktitle={International Conference on Learning Representations},
      year={2023},
}

@misc{grattafiori2024llama3herdmodels,
      title={The {Llama} 3 Herd of Models}, 
      author={Aaron Grattafiori and Abhimanyu Dubey and Abhinav Jauhri and Abhinav Pandey and Abhishek Kadian and Ahmad Al-Dahle and Aiesha Letman and Akhil Mathur and Alan Schelten and Alex Vaughan and Amy Yang and Angela Fan and Anirudh Goyal and Anthony Hartshorn and Aobo Yang and Archi Mitra and Archie Sravankumar and Artem Korenev and Arthur Hinsvark and Arun Rao and Aston Zhang and Aurelien Rodriguez and Austen Gregerson and Ava Spataru and Baptiste Roziere and Bethany Biron and Binh Tang and Bobbie Chern and Charlotte Caucheteux and Chaya Nayak and Chloe Bi and Chris Marra and Chris McConnell and Christian Keller and Christophe Touret and Chunyang Wu and Corinne Wong and Cristian Canton Ferrer and Cyrus Nikolaidis and Damien Allonsius and Daniel Song and Danielle Pintz and Danny Livshits and Danny Wyatt and David Esiobu and Dhruv Choudhary and Dhruv Mahajan and Diego Garcia-Olano and Diego Perino and Dieuwke Hupkes and Egor Lakomkin and Ehab AlBadawy and Elina Lobanova and Emily Dinan and Eric Michael Smith and Filip Radenovic and Francisco Guzmán and Frank Zhang and Gabriel Synnaeve and Gabrielle Lee and Georgia Lewis Anderson and Govind Thattai and Graeme Nail and Gregoire Mialon and Guan Pang and Guillem Cucurell and Hailey Nguyen and Hannah Korevaar and Hu Xu and Hugo Touvron and Iliyan Zarov and Imanol Arrieta Ibarra and Isabel Kloumann and Ishan Misra and Ivan Evtimov and Jack Zhang and Jade Copet and Jaewon Lee and Jan Geffert and Jana Vranes and Jason Park and Jay Mahadeokar and Jeet Shah and Jelmer van der Linde and Jennifer Billock and Jenny Hong and Jenya Lee and Jeremy Fu and Jianfeng Chi and Jianyu Huang and Jiawen Liu and Jie Wang and Jiecao Yu and Joanna Bitton and Joe Spisak and Jongsoo Park and Joseph Rocca and Joshua Johnstun and Joshua Saxe and Junteng Jia and Kalyan Vasuden Alwala and Karthik Prasad and Kartikeya Upasani and Kate Plawiak and Ke Li and Kenneth Heafield and Kevin Stone and Khalid El-Arini and Krithika Iyer and Kshitiz Malik and Kuenley Chiu and Kunal Bhalla and Kushal Lakhotia and Lauren Rantala-Yeary and Laurens van der Maaten and Lawrence Chen and Liang Tan and Liz Jenkins and Louis Martin and Lovish Madaan and Lubo Malo and Lukas Blecher and Lukas Landzaat and Luke de Oliveira and Madeline Muzzi and Mahesh Pasupuleti and Mannat Singh and Manohar Paluri and Marcin Kardas and Maria Tsimpoukelli and Mathew Oldham and Mathieu Rita and Maya Pavlova and Melanie Kambadur and Mike Lewis and Min Si and Mitesh Kumar Singh and Mona Hassan and Naman Goyal and Narjes Torabi and Nikolay Bashlykov and Nikolay Bogoychev and Niladri Chatterji and Ning Zhang and Olivier Duchenne and Onur Çelebi and Patrick Alrassy and Pengchuan Zhang and Pengwei Li and Petar Vasic and Peter Weng and Prajjwal Bhargava and Pratik Dubal and Praveen Krishnan and Punit Singh Koura and Puxin Xu and Qing He and Qingxiao Dong and Ragavan Srinivasan and Raj Ganapathy and Ramon Calderer and Ricardo Silveira Cabral and Robert Stojnic and Roberta Raileanu and Rohan Maheswari and Rohit Girdhar and Rohit Patel and Romain Sauvestre and Ronnie Polidoro and Roshan Sumbaly and Ross Taylor and Ruan Silva and Rui Hou and Rui Wang and Saghar Hosseini and Sahana Chennabasappa and Sanjay Singh and Sean Bell and Seohyun Sonia Kim and Sergey Edunov and Shaoliang Nie and Sharan Narang and Sharath Raparthy and Sheng Shen and Shengye Wan and Shruti Bhosale and Shun Zhang and Simon Vandenhende and Soumya Batra and Spencer Whitman and Sten Sootla and Stephane Collot and Suchin Gururangan and Sydney Borodinsky and Tamar Herman and Tara Fowler and Tarek Sheasha and Thomas Georgiou and Thomas Scialom and Tobias Speckbacher and Todor Mihaylov and Tong Xiao and Ujjwal Karn and Vedanuj Goswami and Vibhor Gupta and Vignesh Ramanathan and Viktor Kerkez and Vincent Gonguet and Virginie Do and Vish Vogeti and Vítor Albiero and Vladan Petrovic and Weiwei Chu and Wenhan Xiong and Wenyin Fu and Whitney Meers and Xavier Martinet and Xiaodong Wang and Xiaofang Wang and Xiaoqing Ellen Tan and Xide Xia and Xinfeng Xie and Xuchao Jia and Xuewei Wang and Yaelle Goldschlag and Yashesh Gaur and Yasmine Babaei and Yi Wen and Yiwen Song and Yuchen Zhang and Yue Li and Yuning Mao and Zacharie Delpierre Coudert and Zheng Yan and Zhengxing Chen and Zoe Papakipos and Aaditya Singh and Aayushi Srivastava and Abha Jain and Adam Kelsey and Adam Shajnfeld and Adithya Gangidi and Adolfo Victoria and Ahuva Goldstand and Ajay Menon and Ajay Sharma and Alex Boesenberg and Alexei Baevski and Allie Feinstein and Amanda Kallet and Amit Sangani and Amos Teo and Anam Yunus and Andrei Lupu and Andres Alvarado and Andrew Caples and Andrew Gu and Andrew Ho and Andrew Poulton and Andrew Ryan and Ankit Ramchandani and Annie Dong and Annie Franco and Anuj Goyal and Aparajita Saraf and Arkabandhu Chowdhury and Ashley Gabriel and Ashwin Bharambe and Assaf Eisenman and Azadeh Yazdan and Beau James and Ben Maurer and Benjamin Leonhardi and Bernie Huang and Beth Loyd and Beto De Paola and Bhargavi Paranjape and Bing Liu and Bo Wu and Boyu Ni and Braden Hancock and Bram Wasti and Brandon Spence and Brani Stojkovic and Brian Gamido and Britt Montalvo and Carl Parker and Carly Burton and Catalina Mejia and Ce Liu and Changhan Wang and Changkyu Kim and Chao Zhou and Chester Hu and Ching-Hsiang Chu and Chris Cai and Chris Tindal and Christoph Feichtenhofer and Cynthia Gao and Damon Civin and Dana Beaty and Daniel Kreymer and Daniel Li and David Adkins and David Xu and Davide Testuggine and Delia David and Devi Parikh and Diana Liskovich and Didem Foss and Dingkang Wang and Duc Le and Dustin Holland and Edward Dowling and Eissa Jamil and Elaine Montgomery and Eleonora Presani and Emily Hahn and Emily Wood and Eric-Tuan Le and Erik Brinkman and Esteban Arcaute and Evan Dunbar and Evan Smothers and Fei Sun and Felix Kreuk and Feng Tian and Filippos Kokkinos and Firat Ozgenel and Francesco Caggioni and Frank Kanayet and Frank Seide and Gabriela Medina Florez and Gabriella Schwarz and Gada Badeer and Georgia Swee and Gil Halpern and Grant Herman and Grigory Sizov and Guangyi and Zhang and Guna Lakshminarayanan and Hakan Inan and Hamid Shojanazeri and Han Zou and Hannah Wang and Hanwen Zha and Haroun Habeeb and Harrison Rudolph and Helen Suk and Henry Aspegren and Hunter Goldman and Hongyuan Zhan and Ibrahim Damlaj and Igor Molybog and Igor Tufanov and Ilias Leontiadis and Irina-Elena Veliche and Itai Gat and Jake Weissman and James Geboski and James Kohli and Janice Lam and Japhet Asher and Jean-Baptiste Gaya and Jeff Marcus and Jeff Tang and Jennifer Chan and Jenny Zhen and Jeremy Reizenstein and Jeremy Teboul and Jessica Zhong and Jian Jin and Jingyi Yang and Joe Cummings and Jon Carvill and Jon Shepard and Jonathan McPhie and Jonathan Torres and Josh Ginsburg and Junjie Wang and Kai Wu and Kam Hou U and Karan Saxena and Kartikay Khandelwal and Katayoun Zand and Kathy Matosich and Kaushik Veeraraghavan and Kelly Michelena and Keqian Li and Kiran Jagadeesh and Kun Huang and Kunal Chawla and Kyle Huang and Lailin Chen and Lakshya Garg and Lavender A and Leandro Silva and Lee Bell and Lei Zhang and Liangpeng Guo and Licheng Yu and Liron Moshkovich and Luca Wehrstedt and Madian Khabsa and Manav Avalani and Manish Bhatt and Martynas Mankus and Matan Hasson and Matthew Lennie and Matthias Reso and Maxim Groshev and Maxim Naumov and Maya Lathi and Meghan Keneally and Miao Liu and Michael L. Seltzer and Michal Valko and Michelle Restrepo and Mihir Patel and Mik Vyatskov and Mikayel Samvelyan and Mike Clark and Mike Macey and Mike Wang and Miquel Jubert Hermoso and Mo Metanat and Mohammad Rastegari and Munish Bansal and Nandhini Santhanam and Natascha Parks and Natasha White and Navyata Bawa and Nayan Singhal and Nick Egebo and Nicolas Usunier and Nikhil Mehta and Nikolay Pavlovich Laptev and Ning Dong and Norman Cheng and Oleg Chernoguz and Olivia Hart and Omkar Salpekar and Ozlem Kalinli and Parkin Kent and Parth Parekh and Paul Saab and Pavan Balaji and Pedro Rittner and Philip Bontrager and Pierre Roux and Piotr Dollar and Polina Zvyagina and Prashant Ratanchandani and Pritish Yuvraj and Qian Liang and Rachad Alao and Rachel Rodriguez and Rafi Ayub and Raghotham Murthy and Raghu Nayani and Rahul Mitra and Rangaprabhu Parthasarathy and Raymond Li and Rebekkah Hogan and Robin Battey and Rocky Wang and Russ Howes and Ruty Rinott and Sachin Mehta and Sachin Siby and Sai Jayesh Bondu and Samyak Datta and Sara Chugh and Sara Hunt and Sargun Dhillon and Sasha Sidorov and Satadru Pan and Saurabh Mahajan and Saurabh Verma and Seiji Yamamoto and Sharadh Ramaswamy and Shaun Lindsay and Shaun Lindsay and Sheng Feng and Shenghao Lin and Shengxin Cindy Zha and Shishir Patil and Shiva Shankar and Shuqiang Zhang and Shuqiang Zhang and Sinong Wang and Sneha Agarwal and Soji Sajuyigbe and Soumith Chintala and Stephanie Max and Stephen Chen and Steve Kehoe and Steve Satterfield and Sudarshan Govindaprasad and Sumit Gupta and Summer Deng and Sungmin Cho and Sunny Virk and Suraj Subramanian and Sy Choudhury and Sydney Goldman and Tal Remez and Tamar Glaser and Tamara Best and Thilo Koehler and Thomas Robinson and Tianhe Li and Tianjun Zhang and Tim Matthews and Timothy Chou and Tzook Shaked and Varun Vontimitta and Victoria Ajayi and Victoria Montanez and Vijai Mohan and Vinay Satish Kumar and Vishal Mangla and Vlad Ionescu and Vlad Poenaru and Vlad Tiberiu Mihailescu and Vladimir Ivanov and Wei Li and Wenchen Wang and Wenwen Jiang and Wes Bouaziz and Will Constable and Xiaocheng Tang and Xiaojian Wu and Xiaolan Wang and Xilun Wu and Xinbo Gao and Yaniv Kleinman and Yanjun Chen and Ye Hu and Ye Jia and Ye Qi and Yenda Li and Yilin Zhang and Ying Zhang and Yossi Adi and Youngjin Nam and Yu and Wang and Yu Zhao and Yuchen Hao and Yundi Qian and Yunlu Li and Yuzi He and Zach Rait and Zachary DeVito and Zef Rosnbrick and Zhaoduo Wen and Zhenyu Yang and Zhiwei Zhao and Zhiyu Ma},
      year={2024},
      eprint={2407.21783},
      archivePrefix={arXiv},
      primaryClass={cs.AI},
      url={https://arxiv.org/abs/2407.21783}, 
}

@inproceedings{vaswani2017attentionneed,
      title={Attention Is All You Need},
      author={Ashish Vaswani and Noam Shazeer and Niki Parmar and Jakob Uszkoreit and Llion Jones and Aidan N. Gomez and Lukasz Kaiser and Illia Polosukhin},
      booktitle={Advances in Neural Information Processing Systems},
      year={2017},
}

@inproceedings{sahoo2025diffusionduality,
      title={The Diffusion Duality},
      author={Subham Sekhar Sahoo and Justin Deschenaux and Aaron Gokaslan and Guanghan Wang and Justin Chiu and Volodymyr Kuleshov},
      booktitle={International Conference on Machine Learning},
      year={2025},
}

@inproceedings{schiff2025simpleguidancemechanismsdiscrete,
      title={Simple Guidance Mechanisms for Discrete Diffusion Models},
      author={Yair Schiff and Subham Sekhar Sahoo and Hao Phung and Guanghan Wang and Sam Boshar and Hugo Dalla-torre and Bernardo P. de Almeida and Alexander Rush and Thomas Pierrot and Volodymyr Kuleshov},
      booktitle={International Conference on Learning Representations},
      year={2025},
}

@inproceedings{austin2023structureddenoisingdiffusionmodels,
      title={Structured Denoising Diffusion Models in Discrete State-Spaces},
      author={Jacob Austin and Daniel D. Johnson and Jonathan Ho and Daniel Tarlow and Rianne van den Berg},
      booktitle={Advances in Neural Information Processing Systems},
      year={2021},
}

@inproceedings{vignac2023digress,
      title={{DiGress}: Discrete Denoising Diffusion for Graph Generation},
      author={Cl{\'e}ment Vignac and Igor Krawczuk and Antoine Siraudin and Bohan Wang and Volkan Cevher and Pascal Frossard},
      booktitle={International Conference on Learning Representations},
      year={2023},
      url={https://openreview.net/forum?id=UaAD-Nu86WX},
}

@inproceedings{qin2025defog,
      title={{DeFoG}: Discrete Flow Matching for Graph Generation},
      author={Yiming Qin and Manuel Madeira and Dorina Thanou and Pascal Frossard},
      booktitle={International Conference on Machine Learning},
      year={2025},
      url={https://proceedings.mlr.press/v267/qin25d.html},
}

@inproceedings{shi2025simplifiedgeneralizedmaskeddiffusion,
      title={Simplified and Generalized Masked Diffusion for Discrete Data},
      author={Jiaxin Shi and Kehang Han and Zhe Wang and Arnaud Doucet and Michalis K. Titsias},
      booktitle={Advances in Neural Information Processing Systems},
      year={2024},
}

@inproceedings{vonrütte2025generalizedinterpolatingdiscretediffusion,
      title={Generalized Interpolating Discrete Diffusion},
      author={Dimitri von Rütte and Janis Fluri and Yuhui Ding and Antonio Orvieto and Bernhard Schölkopf and Thomas Hofmann},
      booktitle={International Conference on Machine Learning},
      year={2025},
}

@misc{Gokaslan2019OpenWeb,  
	title={OpenWebText Corpus},
	author={Aaron Gokaslan and Vanya Cohen},
	howpublished={\url{http://Skylion007.github.io/OpenWebTextCorpus}}, 
	year={2019}
}

@article{wang2025hierarchicalreasoning,
      title={Hierarchical Reasoning Model},
      author={Guan Wang and Jin Li and Yuhao Sun and Xing Chen and Changling Liu and Yue Wu and Meng Lu and Sen Song and Yasin Abbasi Yadkori},
      journal={arXiv preprint arXiv:2506.21734},
      year={2025},
      url={https://arxiv.org/abs/2506.21734},
}

@inproceedings{hersche2026softmasked,
      title={Soft-Masked Diffusion Language Models},
      author={Michael Hersche and Samuel Moor-Smith and Thomas Hofmann and Abbas Rahimi},
      booktitle={International Conference on Learning Representations},
      year={2026},
      url={https://arxiv.org/abs/2510.17206},
}

@inproceedings{sahoo2024simpleeffectivemaskeddiffusion,
      title={Simple and Effective Masked Diffusion Language Models},
      author={Subham Sekhar Sahoo and Marianne Arriola and Yair Schiff and Aaron Gokaslan and Edgar Marroquin and Justin T Chiu and Alexander Rush and Volodymyr Kuleshov},
      booktitle={Advances in Neural Information Processing Systems},
      year={2024},
}

@inproceedings{lou2024discretediffusionmodelingestimating,
      title={Discrete Diffusion Modeling by Estimating the Ratios of the Data Distribution},
      author={Aaron Lou and Chenlin Meng and Stefano Ermon},
      booktitle={International Conference on Machine Learning},
      year={2024},
}

@inproceedings{kim2025finetuningmaskeddiffusionprovable,
      title={Fine-Tuning Masked Diffusion for Provable Self-Correction},
      author={Jaeyeon Kim and Seunggeun Kim and Taekyun Lee and David Z. Pan and Hyeji Kim and Sham Kakade and Sitan Chen},
      booktitle={International Conference on Machine Learning},
      year={2026},
}

@misc{alp2024sudoku,
    author       = {Ali Alp},
    title        = {Sudoku Puzzle Generator},
    year         = {2024},
    howpublished = {\url{https://github.com/alicommit-malp/sudoku}},
  }

@misc{cobbe2021trainingverifierssolvemath,
      title={Training Verifiers to Solve Math Word Problems}, 
      author={Karl Cobbe and Vineet Kosaraju and Mohammad Bavarian and Mark Chen and Heewoo Jun and Lukasz Kaiser and Matthias Plappert and Jerry Tworek and Jacob Hilton and Reiichiro Nakano and Christopher Hesse and John Schulman},
      year={2021},
      eprint={2110.14168},
      archivePrefix={arXiv},
      primaryClass={cs.LG},
      url={https://arxiv.org/abs/2110.14168}, 
}

@misc{Radford2019LanguageMA,
  title={Language Models are Unsupervised Multitask Learners},
  author={Alec Radford and Jeffrey Wu and Rewon Child and David Luan and Dario Amodei and Ilya Sutskever},
  year={2019},
}

@inproceedings{allal2025smollm2smolgoesbig,
      title={SmolLM2: When Smol Goes Big --- Data-Centric Training of a Fully Open Small Language Model},
      author={Loubna Ben Allal and Anton Lozhkov and Elie Bakouch and Gabriel Martín Blázquez and Guilherme Penedo and Lewis Tunstall and Andrés Marafioti and Agustín Piqueres Lajarín and Hynek Kydlíček and Vaibhav Srivastav and Joshua Lochner and Caleb Fahlgren and Xuan-Son Nguyen and Ben Burtenshaw and Clémentine Fourrier and Haojun Zhao and Hugo Larcher and Mathieu Morlon and Cyril Zakka and Colin Raffel and Leandro von Werra and Thomas Wolf},
      booktitle={Conference on Language Modeling},
      year={2025},
}

@inproceedings{peebles2023scalablediffusionmodelstransformers,
      title={Scalable Diffusion Models with Transformers},
      author={William Peebles and Saining Xie},
      booktitle={Proceedings of the IEEE/CVF International Conference on Computer Vision},
      year={2023},
}

@article{su2023roformerenhancedtransformerrotary,
      title={{RoFormer}: Enhanced Transformer with Rotary Position Embedding},
      author={Jianlin Su and Murtadha Ahmed and Yu Lu and Shengfeng Pan and Wen Bo and Yunfeng Liu},
      journal={Neurocomputing},
      volume={568},
      pages={127063},
      year={2024},
}

@inproceedings{li2022diffusionlmimprovescontrollabletext,
      title={Diffusion-LM Improves Controllable Text Generation},
      author={Xiang Lisa Li and John Thickstun and Ishaan Gulrajani and Percy Liang and Tatsunori B. Hashimoto},
      booktitle={Advances in Neural Information Processing Systems},
      year={2022},
}

@inproceedings{batzolis2026cobit,
      title={Towards Closing the Autoregressive Gap in Language Modeling via Entropy-Gated Continuous Bitstream Diffusion},
      author={Georgios Batzolis and Mark Girolami and Luca Ambrogioni},
      booktitle={Advances in Neural Information Processing Systems},
      year={2026},
}

@inproceedings{yang2026continuousdiffusionscales,
      title={Continuous Diffusion Scales Competitively with Discrete Diffusion for Language},
      author={Zhihan Yang and Wei Guo and Shuibai Zhang and Subham Sekhar Sahoo and Yongxin Chen and Arash Vahdat and Morteza Mardani and John Thickstun},
      booktitle={Advances in Neural Information Processing Systems},
      year={2026},
}

@inproceedings{pynadath2025candihybriddiscretecontinuousdiffusion,
      title={CANDI: Hybrid Discrete-Continuous Diffusion Models},
      author={Patrick Pynadath and Jiaxin Shi and Ruqi Zhang},
      booktitle={International Conference on Machine Learning},
      year={2026},
}

@misc{campbell2024generativeflowsdiscretestatespaces,
      title={Generative Flows on Discrete State-Spaces: Enabling Multimodal Flows with Applications to Protein Co-Design}, 
      author={Andrew Campbell and Jason Yim and Regina Barzilay and Tom Rainforth and Tommi Jaakkola},
      year={2024},
      eprint={2402.04997},
      archivePrefix={arXiv},
      primaryClass={stat.ML},
      url={https://arxiv.org/abs/2402.04997}, 
}

@inproceedings{campbell2022continuoustimeframeworkdiscrete,
      title={A Continuous Time Framework for Discrete Denoising Models},
      author={Andrew Campbell and Joe Benton and Valentin De Bortoli and Tom Rainforth and George Deligiannidis and Arnaud Doucet},
      booktitle={Advances in Neural Information Processing Systems},
      year={2022},
}

@inproceedings{deschenaux2025partitiongenerativemodelingmasked,
      title={Partition Generative Modeling: Masked Modeling Without Masks},
      author={Justin Deschenaux and Lan Tran and Caglar Gulcehre},
      booktitle={International Conference on Learning Representations},
      year={2026},
}

@misc{deschenaux2024promisesoutlookschallengesdiffusion,
      title={Promises, Outlooks and Challenges of Diffusion Language Modeling}, 
      author={Justin Deschenaux and Caglar Gulcehre},
      year={2024},
      eprint={2406.11473},
      archivePrefix={arXiv},
      primaryClass={cs.CL},
      url={https://arxiv.org/abs/2406.11473}, 
}

@misc{dieleman2022continuousdiffusioncategoricaldata,
      title={Continuous diffusion for categorical data}, 
      author={Sander Dieleman and Laurent Sartran and Arman Roshannai and Nikolay Savinov and Yaroslav Ganin and Pierre H. Richemond and Arnaud Doucet and Robin Strudel and Chris Dyer and Conor Durkan and Curtis Hawthorne and Rémi Leblond and Will Grathwohl and Jonas Adler},
      year={2022},
      eprint={2211.15089},
      archivePrefix={arXiv},
      primaryClass={cs.CL},
      url={https://arxiv.org/abs/2211.15089}, 
}

@inproceedings{nagarajan2025roll,
      title={Roll the dice \& look before you leap: Going beyond the creative limits of next-token prediction},
      author={Nagarajan, Vaishnavh and Wu, Chen Henry and Ding, Charles and Raghunathan, Aditi},
      booktitle={International Conference on Machine Learning},
      year={2025},
}

@misc{kaplan2020scalinglawsneurallanguage,
      title={Scaling Laws for Neural Language Models},
      author={Jared Kaplan and Sam McCandlish and Tom Henighan and Tom B. Brown and Benjamin Chess and Rewon Child and Scott Gray and Alec Radford and Jeffrey Wu and Dario Amodei},
      year={2020},
      eprint={2001.08361},
      archivePrefix={arXiv},
      primaryClass={cs.LG},
      url={https://arxiv.org/abs/2001.08361},
}

@inproceedings{papadopoulos2024arrowstimelargelanguage,
      title={Arrows of Time for Large Language Models},
      author={Vassilis Papadopoulos and Jérémie Wenger and Clément Hongler},
      booktitle={International Conference on Machine Learning},
      year={2024},
}

@inproceedings{kitouni2024factorizationcursetokenspredict,
      title={The Factorization Curse: Which Tokens You Predict Underlie the Reversal Curse and More},
      author={Ouail Kitouni and Niklas Nolte and Diane Bouchacourt and Adina Williams and Mike Rabbat and Mark Ibrahim},
      booktitle={Advances in Neural Information Processing Systems},
      year={2024},
}

@inproceedings{roos2026categoricalflowmaps,
      title={Categorical Flow Maps},
      author={Daan Roos and Oscar Davis and Floor Eijkelboom and Michael Bronstein and Max Welling and İsmail İlkan Ceylan and Luca Ambrogioni and Jan-Willem van de Meent},
      booktitle={International Conference on Machine Learning},
      year={2026},
}

@inproceedings{lee2026onesteplanguagemodelingcontinuous,
      title={Flow Map Language Models: One-step Language Modeling via Continuous Denoising},
      author={Chanhyuk Lee and Jaehoon Yoo and Manan Agarwal and Sheel Shah and Jerry Huang and Aditi Raghunathan and Seunghoon Hong and Nicholas M. Boffi and Jinwoo Kim},
      booktitle={Advances in Neural Information Processing Systems},
      year={2026},
}

@misc{potaptchik2026discreteflowmaps,
      title={Discrete Flow Maps}, 
      author={Peter Potaptchik and Jason Yim and Adhi Saravanan and Peter Holderrieth and Eric Vanden-Eijnden and Michael S. Albergo},
      year={2026},
      eprint={2604.09784},
      archivePrefix={arXiv},
      primaryClass={stat.ML},
      url={https://arxiv.org/abs/2604.09784}, 
}

@misc{shen2026codarcontinuousdiffusionlanguage,
      title={CoDAR: Continuous Diffusion Language Models are More Powerful Than You Think},
      author={Junzhe Shen and Jieru Zhao and Ziwei He and Zhouhan Lin},
      year={2026},
      eprint={2603.02547},
      archivePrefix={arXiv},
      primaryClass={cs.CL},
      url={https://arxiv.org/abs/2603.02547},
}

@inproceedings{gulrajani2023likelihoodbaseddiffusionlanguagemodels,
      title={Likelihood-Based Diffusion Language Models},
      author={Ishaan Gulrajani and Tatsunori B. Hashimoto},
      booktitle={Advances in Neural Information Processing Systems},
      year={2023},
}

@inproceedings{jo2025continuousdiffusionmodellanguage,
      title={Continuous Diffusion Model for Language Modeling},
      author={Jaehyeong Jo and Sung Ju Hwang},
      booktitle={Advances in Neural Information Processing Systems},
      year={2025},
}

@misc{strudel2022selfconditionedembeddingdiffusion,
      title={Self-conditioned Embedding Diffusion for Text Generation},
      author={Robin Strudel and Corentin Tallec and Florent Altch\'e and Yilun Du and Yaroslav Ganin and Arthur Mensch and Will Grathwohl and Nikolay Savinov and Sander Dieleman and Laurent Sifre and R\'emi Leblond},
      year={2022},
      eprint={2211.04236},
      archivePrefix={arXiv},
      primaryClass={cs.CL},
      url={https://arxiv.org/abs/2211.04236},
}

@inproceedings{chen2022analogbits,
      title={Analog Bits: Generating Discrete Data using Diffusion Models with Self-Conditioning},
      author={Ting Chen and Ruixiang Zhang and Geoffrey Hinton},
      booktitle={International Conference on Learning Representations},
      year={2023},
}

@inproceedings{jo2025loopholingdiscretediffusion,
      title={Loopholing Discrete Diffusion: Deterministic Bypass of the Sampling Wall},
      author={Mingyu Jo and Jaesik Yoon and Justin Deschenaux and Caglar Gulcehre and Sungjin Ahn},
      booktitle={International Conference on Learning Representations},
      year={2026},
}

@misc{velickovic2026perplexitycannotalwaystellright,
      title={Perplexity Cannot Always Tell Right from Wrong},
      author={Petar Veli\v{c}kovi\'c and Federico Barbero and Christos Perivolaropoulos and Simon Osindero and Razvan Pascanu},
      year={2026},
      eprint={2601.22950},
      archivePrefix={arXiv},
      primaryClass={cs.LG},
      url={https://arxiv.org/abs/2601.22950},
}

@inproceedings{feng2025theoreticalbenefitlimitationdiffusion,
      title={Theoretical Benefit and Limitation of Diffusion Language Model},
      author={Guhao Feng and Yihan Geng and Jian Guan and Wei Wu and Liwei Wang and Di He},
      booktitle={Advances in Neural Information Processing Systems},
      year={2025},
}

@misc{liu2023tinygsm,
      title={{TinyGSM}: Achieving $>$80\% on {GSM8k} with Small Language Models},
      author={Bingbin Liu and Sebastien Bubeck and Ronen Eldan and Janardhan Kulkarni and Yuanzhi Li and Anh Nguyen and Rachel Ward and Yi Zhang},
      year={2023},
      eprint={2312.09241},
      archivePrefix={arXiv},
      primaryClass={cs.CL},
      url={https://arxiv.org/abs/2312.09241},
}

@inproceedings{ou2024radd,
      title={Your Absorbing Discrete Diffusion Secretly Models the Conditional Distributions of Clean Data},
      author={Jingyang Ou and Shen Nie and Kaiwen Xue and Fengqi Zhu and Jiacheng Sun and Zhenguo Li and Chongxuan Li},
      booktitle={International Conference on Learning Representations},
      year={2025},
}

@article{deschenaux2026hypersphericalflows,
  title={Language Modeling with Hyperspherical Flows},
  author={Deschenaux, Justin and Gulcehre, Caglar},
  journal={arXiv preprint arXiv:2605.11125},
  year={2026}
}

@inproceedings{hu2026elfembeddedlanguageflows,
      title={{ELF}: Embedded Language Flows},
      author={Hu, Keya and Qiu, Linlu and Lu, Yiyang and Zhao, Hanhong and Li, Tianhong and Kim, Yoon and Andreas, Jacob and He, Kaiming},
      booktitle={Advances in Neural Information Processing Systems},
      year={2026},
}

@inproceedings{chen2026langflowcontinuousdiffusion,
      title={{LangFlow}: Continuous Diffusion Rivals Discrete in Language Modeling},
      author={Chen, Yuxin and Liang, Chumeng and Sui, Hangke and Guo, Ruihan and Cheng, Chaoran and You, Jiaxuan and Liu, Ge},
      booktitle={Advances in Neural Information Processing Systems},
      year={2026},
}

@inproceedings{chemseddine2026sphericalflowscategorical,
      title={Spherical Flows for Sampling Categorical Data},
      author={Chemseddine, Jannis and Kornhardt, Gregor and Steidl, Gabriele},
      booktitle={Advances in Neural Information Processing Systems},
      year={2026},
}

@article{hoogeboom2026discretemmd,
  title={Beyond Single Tokens: Distilling Discrete Diffusion Models via Discrete {MMD}},
  author={Hoogeboom, Emiel and Ruhe, David and Heek, Jonathan and Mensink, Thomas and Salimans, Tim},
  journal={arXiv preprint arXiv:2603.20155},
  year={2026}
}

@inproceedings{li2026idlm,
  title={{IDLM}: Inverse-distilled Diffusion Language Models},
  author={Li, David and Gushchin, Nikita and Abulkhanov, Dmitry and Moulines, Eric and Oseledets, Ivan and Panov, Maxim and Korotin, Alexander},
  booktitle={International Conference on Machine Learning},
  year={2026}
}

@misc{gourevitch2026uniformdiffusionmodels,
      title={Uniform Diffusion Models Revisited: Leave-One-Out Denoiser and Absorbing State Reformulation},
      author={Samson Gourevitch and Yazid Janati and Dario Shariatian and Umut Simsekli and Eric Moulines and Eric P. Xing and Alain Durmus},
      year={2026},
      eprint={2605.22765},
      archivePrefix={arXiv},
      primaryClass={cs.LG},
      url={https://arxiv.org/abs/2605.22765},
}

@misc{carballocastro2026recursivescaling,
      title={Recursive Scaling in Masked Diffusion Models},
      author={Alba Carballo-Castro and Julianna Piskorz and Paulius Rauba and Mihaela van der Schaar and Pascal Frossard},
      year={2026},
      eprint={2606.18022},
      archivePrefix={arXiv},
      primaryClass={cs.LG},
      url={https://arxiv.org/abs/2606.18022},
}

@inproceedings{lee2026loopeddiffusion,
      title={Looped Diffusion Language Models},
      author={Sanghyun Lee and Chunsan Hong and Seungryong Kim and Jonghyun Lee and Jongho Park and Dongmin Park},
      booktitle={Advances in Neural Information Processing Systems},
      year={2026},
}

@inproceedings{miele2026fixedpointmasked,
      title={Fixed-Point Masked Generative Modeling},
      author={Andrea Miele and Yiming Qin and Alba Carballo-Castro and Justin Deschenaux and Pascal Frossard},
      booktitle={Advances in Neural Information Processing Systems},
      year={2026},
}

@misc{diffusiongemma2026,
      title={{DiffusionGemma} Technical Report},
      author={{DiffusionGemma Team} and Adrien Ali Ta{\"i}ga and James Assiene and Daniele Calandriello and Rahma Chaabouni and Jo{\~a}o Gante and Tamara von Glehn and Nate Keating and Chris Knutsen and Martin Kukla and Tianlin Liu and Ivan Lobov and Ofir Nabati and Jo{\~a}o Gabriel Oliveira and Nicolas Perez-Nieves and Nastasia Prutianova and Bobak Shahriari and Jean Tarbouriech and Pavel Tyletski and {\c{C}}a{\u{g}}lar {\"U}nl{\"u} and Cindy Wu and Glenn Cameron and Jerome Connor and Sertan Girgin and Maarten Grootendorst and Alon Levkovitch and Eliya Nachmani and Omar Sanseviero and Piotr Stanczyk and Quentin Berthet and Andrew Campbell and Cl{\'e}ment Crepy and Valentin De Bortoli and Arnaud Doucet and Romuald Elie and Alexandre Galashov and Klaus Greff and Alexis Jacq and David Ruhe and Yu-Han Wu and Sebastian Flennerhag and Brendan O'Donoghue and George Scrivener and Shantanu Thakoor},
      year={2026},
      eprint={2608.00146},
      archivePrefix={arXiv},
      primaryClass={cs.CL},
      url={https://arxiv.org/abs/2608.00146},
}

@misc{rozonoyer2026relayrepresentations,
      title={Learned Relay Representations for Forward-Thinking Discrete Diffusion Models},
      author={Benjamin Rozonoyer and Jacopo Minniti and Dhruvesh Patel and Neil Band and Avishek Joey Bose and Tim G. J. Rudner and Andrew McCallum},
      year={2026},
      eprint={2605.22967},
      archivePrefix={arXiv},
      primaryClass={cs.LG},
      url={https://arxiv.org/abs/2605.22967},
}

@article{ba2016layer,
  title={Layer normalization},
  author={Ba, Jimmy Lei and Kiros, Jamie Ryan and Hinton, Geoffrey E},
  journal={arXiv preprint arXiv:1607.06450},
  year={2016}
}

@inproceedings{geiping2025scalingtesttimecomputelatent,
      title={Scaling up Test-Time Compute with Latent Reasoning: A Recurrent Depth Approach},
      author={Jonas Geiping and Sean McLeish and Neel Jain and John Kirchenbauer and Siddharth Singh and Brian R. Bartoldson and Bhavya Kailkhura and Abhinav Bhatele and Tom Goldstein},
      booktitle={Advances in Neural Information Processing Systems},
      year={2025},
}

@inproceedings{noci2022signalpropagationtransformerstheoretical,
      title={Signal Propagation in Transformers: Theoretical Perspectives and the Role of Rank Collapse},
      author={Lorenzo Noci and Sotiris Anagnostidis and Luca Biggio and Antonio Orvieto and Sidak Pal Singh and Aurelien Lucchi},
      booktitle={Advances in Neural Information Processing Systems},
      year={2022},
}

@inproceedings{lehnert2024abetterplanningtransformers,
      title={Beyond A*: Better Planning with Transformers via Search Dynamics Bootstrapping},
      author={Lucas Lehnert and Sainbayar Sukhbaatar and DiJia Su and Qinqing Zheng and Paul Mcvay and Michael Rabbat and Yuandong Tian},
      booktitle={Conference on Language Modeling},
      year={2024},
}

@InProceedings{SciPyProceedings_11,
  author =       {Aric A. Hagberg and Daniel A. Schult and Pieter J. Swart},
  title =        {Exploring Network Structure, Dynamics, and Function using NetworkX},
  booktitle =   {Proceedings of the 7th Python in Science Conference},
  pages =     {11 - 15},
  address = {Pasadena, CA USA},
  year =      {2008},
  editor =    {Ga\"el Varoquaux and Travis Vaught and Jarrod Millman},
}

@misc{madeira2026trajectoryaware,
      title={On Trajectory-Aware Training for Masked Diffusion Language Models},
      author={Manuel Madeira and Amitis Shidani and Alice Bizeul and Victor Turrisi and Louis B{\'e}thune and Bhavika Devnani and Dan Busbridge and Pierre Ablin and Jo{\~a}o Monteiro},
      year={2026},
      eprint={2609.37974},
      archivePrefix={arXiv},
      url={https://arxiv.org/abs/2609.37974},
}

@misc{gushchin2026alphadlm,
      title={Alpha Diffusion Language Models: Factorization Alone Is Not the Problem},
      author={Nikita Gushchin and Dmitry Baranchuk and Alexander Korotin},
      year={2026},
      eprint={2609.38066},
      archivePrefix={arXiv},
      url={https://arxiv.org/abs/2609.38066},
}

@misc{agarwal2026posterior,
      title={Posterior Refinement: Fast Language Generation via Any-Order Flow Maps},
      author={Manan Agarwal and Sheel Shah and Chanhyuk Lee and Jaehoon Yoo and Jerry Huang and Seunghoon Hong and Aditi Raghunathan and Jinwoo Kim and Nicholas M. Boffi},
      year={2026},
      eprint={2606.24773},
      archivePrefix={arXiv},
      url={https://arxiv.org/abs/2606.24773},
}

@misc{tang2026dbtm,
      title={Discrete Beckmann Transport Models for One-Step Language Modeling and Reasoning},
      author={Sophia Tang and Shiyi Wang},
      year={2026},
      eprint={2609.15903},
      archivePrefix={arXiv},
      url={https://arxiv.org/abs/2609.15903},
}

@misc{deschenaux2026simplex,
      title={Simplex Diffusion Models},
      author={Justin Deschenaux and Alexandre Galashov and Andrew Campbell and Li Kevin Wenliang and James Thornton and Arnaud Doucet and Valentin De Bortoli},
      year={2026},
      eprint={2609.35553},
      archivePrefix={arXiv},
      url={https://arxiv.org/abs/2609.35553},
}

@misc{sun2026freda,
      title={Forward-Free Diffusion Language Models with {BPTT}-Free Looped Refinement},
      author={Haotian Sun and Rushi Qiang and Yuqian Zheng and Bo Dai},
      year={2026},
      eprint={2606.08357},
      archivePrefix={arXiv},
      primaryClass={cs.CL},
      url={https://arxiv.org/abs/2606.08357},
}

@misc{helbling2026flowreasoningmodelsturning,
      title={Flow Reasoning Models: Turning Flows Into Efficient Recurrent Reasoners},
      author={Alec Helbling and Andrey Bryutkin and Mauro Martino and Duen Horng Chau and Nima Dehmamy and Hendrik Strobelt},
      year={2026},
      eprint={2606.29150},
      archivePrefix={arXiv},
      primaryClass={cs.AI},
      url={https://arxiv.org/abs/2606.29150},
}

@inproceedings{ng2004featureselectionl1l2,
      title={Feature Selection, {L1} vs. {L2} Regularization, and Rotational Invariance},
      author={Andrew Y. Ng},
      booktitle={International Conference on Machine Learning},
      year={2004},
}

@inproceedings{grinsztajn2022treebasedmodelsstilloutperform,
      title={Why Do Tree-Based Models Still Outperform Deep Learning on Typical Tabular Data?},
      author={L{\'e}o Grinsztajn and Edouard Oyallon and Ga{\"e}l Varoquaux},
      booktitle={Advances in Neural Information Processing Systems, Datasets and Benchmarks Track},
      year={2022},
}

@article{damario2022dataefficiencydeeplearning,
      title={The Data Efficiency of Deep Learning Is Degraded by Unnecessary Input Dimensions},
      author={Vanessa D'Amario and Sanjana Srivastava and Tomotake Sasaki and Xavier Boix},
      journal={Frontiers in Computational Neuroscience},
      volume={16},
      pages={760085},
      year={2022},
}

@inproceedings{shi2023largelanguagemodelseasily,
      title={Large Language Models Can Be Easily Distracted by Irrelevant Context},
      author={Freda Shi and Xinyun Chen and Kanishka Misra and Nathan Scales and David Dohan and Ed H. Chi and Nathanael Sch{\"a}rli and Denny Zhou},
      booktitle={International Conference on Machine Learning},
      year={2023},
}
\bibliographystyle{iclr2027_conference}

\clearpage
\setcounter{tocdepth}{2}
\tableofcontents
\clearpage

\appendix
\section{Method Details}
\label{app:method-details}

\subsection{Uniform Diffusion: Posterior and NELBO}
\label{app:nelbo}
We reproduce the exact UDM quantities of Duo~\citep{sahoo2025diffusionduality}, with $\alpha_{t|s}=\alpha_t/\alpha_s$. The denoising posterior iterated in \Eqn{eq:udm-sampling} is
{\footnotesize
\begin{equation}
    \label{eq:udm-posterior-full}
    \qst(\cdot \mid \ztl, \xl) = \cat\left(\cdot; \frac{|\mathcal{V}|\at \ztl \odot \xl + (\alpha_{t|s} - \at)\ztl + (\alpha_s - \alpha_t)\xl + (1 - \alpha_{t|s})(1- \as)\tfrac{1}{|\mathcal{V}|}\mathbf{1}}{|\mathcal{V}| \at\langle \ztl, \xl\rangle  + 1 - \at}\right).
\end{equation}
}%
This posterior induces the per-sequence NELBO
\begin{equation*}
    \mathcal{L}(\theta;\x) = \mathbb{E}_{t \sim \mathcal{U}[0,1],\, \z_t\sim q_t(\cdot\mid\x)} \sum_{\ell=1}^{L}\felbo(\denoise^{\supl}(\z_t, t), \xl, \ztl, \at).
\end{equation*}
Its average over $\x\sim q_{\text{data}}$ is $\mathcal{L}(\theta)$~\Eqn{eq:udm-loss}, and its per-token loss is
{\footnotesize
\begin{align}
    \felbo(\denoise^{\supl}(\z_t, t), \xl, \ztl, & \at) = -\frac{\at'}{|\mathcal{V}|\at}\Bigg[\frac{|\mathcal{V}|}{\langle\barx^{\supl},\ztl\rangle} - \frac{|\mathcal{V}|}{\langle\barx_\theta^{\supl},\ztl\rangle} \nonumber - \left(\zeta_t \mathds{1}_{\ztl = \xl} + \mathds{1}_{\ztl \neq \xl}\right)\sum_{j=1}^{|\mathcal{V}|} \log\frac{\langle\barx^{\supl}_\theta,\ztl\rangle}{\langle\barx^{\supl}_\theta,\mathbf{e}_j\rangle} \\
    & - |\mathcal{V}|\frac{\at}{1 - \at} \log \frac{\langle\barx^{\supl}_\theta,\ztl\rangle}{\langle\barx^{\supl}_\theta,\xl\rangle}\mathds{1}_{\ztl \neq \xl} - \left((|\mathcal{V}| - 1)\zeta_t \mathds{1}_{\ztl = \xl} - \frac{1}{\zeta_t}\mathds{1}_{\ztl \neq \xl}\right)\log \zeta_t
    \Bigg].
\end{align}
}%
Here $\barx^{\supl} = |\mathcal{V}|\alpha_t\x^{\supl} + (1 - \alpha_t)\mathbf{1}$, $\barx^{\supl}_\theta = |\mathcal{V}|\alpha_t \denoise^{\supl}(\z_t, t) + (1 - \alpha_t)\mathbf{1}$, $\alpha_t'$ is the time derivative of $\alpha_t$, and $\zeta_t = \frac{1 - \alpha_t}{|\mathcal{V}| \alpha_t + 1 - \alpha_t}$.

\subsection{Parameterizations of the Reverse Transitions in Uniform Diffusion}
\label{app:parameterization_reverse}
Sampling from a DDM requires a reverse transition from $\mathbf{z}_t$ to $\mathbf{z}_s$, with $s<t$. A \emph{parameterization} defines the transition in terms of the denoisers' predictions $\mathbf{x}_\theta(\mathbf{z}_t,t)$. 
\paragraph{The Marginalization Parameterization.}
The \emph{marginalization} parameterization~\citep{austin2023structureddenoisingdiffusionmodels,gourevitch2026uniformdiffusionmodels} is derived from the exact reverse transition $q_{s|t}(\mathbf{z}_s^{\supl}\mid\mathbf{z}_t)$, which conditions on the full noisy sequence and marginalizes over the clean token:
\begin{equation}
\label{eq:udm-reverse-marginal}
\begin{aligned}
    q_{s|t}(\mathbf{z}_s^{\supl}\mid\mathbf{z}_t)
    &=\sum_{i=1}^{|\mathcal V|}
    q(\mathbf{z}_s^{\supl},\mathbf{x}^{\supl}=\mathbf{e}_i\mid\mathbf{z}_t) \\
    &=\sum_{i=1}^{|\mathcal V|}
    q(\mathbf{z}_s^{\supl}\mid\mathbf{z}_t,\mathbf{x}^{\supl}=\mathbf{e}_i)\,
    q(\mathbf{x}^{\supl}=\mathbf{e}_i\mid\mathbf{z}_t) \\
    &=\sum_{i=1}^{|\mathcal V|}
    q_{s|t}(\mathbf{z}_s^{\supl}\mid\mathbf{z}_t^{\supl},\mathbf{e}_i)\,
    q(\mathbf{x}^{\supl}=\mathbf{e}_i\mid\mathbf{z}_t).
\end{aligned}
\end{equation}
The second equality factors the joint distribution, and the third holds because the forward process corrupts each position independently. The factor $q_{s|t}(\mathbf{z}_s^{\supl}\mid\mathbf{z}_t^{\supl},\mathbf{x}^{\supl})$ is the denoising posterior, also called the \emph{bridge}~\citep{gourevitch2026uniformdiffusionmodels}, and has a closed form~\Eqn{eq:udm-posterior-full}. In contrast, the clean-token posterior $q(\mathbf{x}^{\supl}\mid\mathbf{z}_t)$ conditions on the full noisy sequence and is intractable. The \emph{marginalization} parameterization replaces $q(\mathbf{x}^{\supl}=\mathbf{e}_i\mid\mathbf{z}_t)$ with the predicted probability $\langle\mathbf{e}_i,\mathbf{x}_\theta^{\supl}(\mathbf{z}_t,t)\rangle$. A direct implementation evaluates the sum in \Eqn{eq:udm-reverse-marginal} for each of the $|\mathcal{V}|$ values of $\mathbf{z}_s^{\supl}$. While affordable for small vocabularies, as in graph and molecule generation~\citep{vignac2023digress}, summing is not tractable for language vocabularies with tens of thousands of tokens. Instead, one can draw $\widetilde{\mathbf{x}}^{\supl}\sim\mathrm{Cat}(\cdot;\mathbf{x}_\theta^{\supl}(\mathbf{z}_t,t))$ and then $\mathbf{z}_s^{\supl}\sim q_{s|t}(\cdot\mid\mathbf{z}_t^{\supl},\widetilde{\mathbf{x}}^{\supl})$. This two-stage draw samples from exactly the same transition as the sum~\citep[App.~G]{campbell2024generativeflowsdiscretestatespaces}, suiting both small~\citep{qin2025defog} and large vocabularies.
\paragraph{The Bridge Plug-In Parameterization.}
Recent UDMs applied to language modeling instead use the \emph{bridge plug-in} parameterization~\citep{schiff2025simpleguidancemechanismsdiscrete,sahoo2025diffusionduality,gourevitch2026uniformdiffusionmodels}, whose expression is in \Eqn{eq:udm-sampling}. The bridge plug-in evaluates the bridge~\Eqn{eq:udm-posterior-full} directly at the simplex-valued prediction $\mathbf{x}_\theta^{\supl}(\mathbf{z}_t,t)$, in place of the one-hot clean token $\mathbf{x}^{\supl}$. The bridge plug-in parameterization coincides with the marginalization parameterization when the bridge is affine in $\mathbf{x}^{\supl}$. The bridge is affine for masked diffusion~\citep{sahoo2024simpleeffectivemaskeddiffusion,shi2025simplifiedgeneralizedmaskeddiffusion,ou2024radd}, but not for UDMs, where the normalizer depends on $\mathbf{x}^{\supl}$~\citep{gourevitch2026uniformdiffusionmodels}.
\paragraph{Relationship to the NELBO.}
The NELBO depends on the prediction $\mathbf{x}_\theta$ only through the model transition $p^\theta_{s|t}$, which differs between parameterizations:
\begin{equation}
\label{eq:udm-model-transitions}
\begin{aligned}
\text{(marginalization)}\quad &\tilde{p}^\theta_{s|t}(\mathbf{z}_s^{\supl}\mid\mathbf{z}_t)=\sum_{i=1}^{|\mathcal V|} q_{s|t}(\mathbf{z}_s^{\supl}\mid\mathbf{z}_t^{\supl}, \mathbf{e}_i)\,\langle \mathbf{e}_i,\mathbf{x}_\theta^{\supl}(\mathbf{z}_t,t)\rangle,\\
\text{(bridge plug-in)}\quad &\hat{p}^\theta_{s|t}(\mathbf{z}_s^{\supl}\mid\mathbf{z}_t)=q_{s|t}\big(\mathbf{z}_s^{\supl}\mid\mathbf{z}_t^{\supl},\mathbf{x}_\theta^{\supl}(\mathbf{z}_t,t)\big).
\end{aligned}
\end{equation}
Up to a constant independent of $\theta$, the expected NELBO with $N$ steps is~\citep[Sec.~3.1]{gourevitch2026uniformdiffusionmodels}
\begin{equation}
\label{eq:udm-nelbo-kl}
\sum_{n=1}^{N}\sum_{\ell=1}^{L}\mathbb{E}_{\mathbf{z}_{t}}\Big[\mathrm{KL}\big(q_{s|t}(\cdot\mid\mathbf{z}_t)\,\big\|\,p^\theta_{s|t}(\cdot\mid\mathbf{z}_t)\big)\Big],
\qquad s=t_{n-1},\ t=t_n,
\end{equation}
where both distributions are over $\mathbf{z}_s^{\supl}$, $q_{s|t}(\cdot\mid\mathbf{z}_t)$ is the exact reverse transition~\Eqn{eq:udm-reverse-marginal}, and the learned transition $p^\theta_{s|t}$ takes the form of $\tilde{p}^\theta_{s|t}$ (marginalization) or $\hat{p}^\theta_{s|t}$ (bridge plug-in). Each KL term in \Eqn{eq:udm-nelbo-kl} vanishes when the learned transition equals the exact one.
For the marginalization parameterization, the NELBO is minimized with $\mathbf{x}_\theta^{\supl}(\mathbf{z}_t,t)=\mathbb{E}[\mathbf{x}^{\supl}\mid\mathbf{z}_t]$, which also minimizes the cross-entropy. With the bridge plug-in parameterization, the NELBO is minimized by the \emph{leave-one-out} posterior mean $\mathbb{E}[\mathbf{x}^{\supl}\mid\mathbf{z}_t^{-\supl}]$, where $\mathbf{z}_t^{-\supl}$ denotes $\mathbf{z}_t$ without the position (row) $\ell$~\citep[Prop.~1]{gourevitch2026uniformdiffusionmodels}. For simplicity and consistency with \citet{schiff2025simpleguidancemechanismsdiscrete,sahoo2025diffusionduality}, the main text uses the bridge plug-in in \Eqn{eq:udm-sampling}.
Although the parameterization and objective determine which denoiser we converge to, it does not influence the sampler for \method.
\subsection{Training and Sampling Algorithms}
\label{app:sampling-algorithm}
\algo{alg:law-sampling} summarizes the sampler of \method (\sec{sec:method-sampling}), and \algo{alg:law-training} one training update (\sec{sec:method-training}).

\begin{algorithm}[h]
\caption{Sampling with \method}
\label{alg:law-sampling}
\begin{algorithmic}[1]
\REQUIRE Backbone $f_\theta$, embedding $E_\theta$, LM head $g_\theta$, state map $\statemap_\theta^{\mathrm{E.E.}}$, decoder $\eta$, sampling temperature $T$, time grid $t_N > \cdots > t_0$
\STATE $\mathbf{z}_{t_N} \sim \mathrm{Cat}(\cdot\,;\, \tfrac{1}{|\mathcal{V}|}\mathbf{1})$ \Comment{Draw from the uniform prior}
\STATE $\mathbf{c} \gets \mathbf{z}_{t_N}E_\theta$ \Comment{Embed to initialize the state}
\FOR{$n = N$ \TO $2$}
    \STATE $\mathbf{h} \gets f_\theta(\mathbf{c}, t_n)$ \Comment{Backbone forward pass}
    \STATE $\mathbf{c} \gets \statemap_\theta^{\mathrm{E.E.}}(\mathbf{h}; T)$ \Comment{Propagate the SC state~\Eqn{eq:law-state-map}}
\ENDFOR
\STATE $\mathbf{h} \gets f_\theta(\mathbf{c}, t_1)$ \Comment{Final forward pass}
\RETURN $\mathbf{z}_{t_0} = \eta\!\left(\mathrm{softmax}(g_\theta(\mathbf{h})/T)\right)$ \Comment{Decode tokens~\Eqn{eq:law-decoding}}
\end{algorithmic}
\end{algorithm}

During sampling, the state map and the decoder share the temperature $T$~\Eqn{eq:law-state-map}.

\begin{algorithm}[h]
\caption{Training update of \method}
\label{alg:law-training}
\begin{algorithmic}[1]
\REQUIRE Backbone $f_\theta$, embedding $E_\theta$, LM head $g_\theta$, state map $\statemap_\theta^{\mathrm{E.E.}}$, clean sequence $\mathbf{x}$, multi-pass probability $p$, number of passes $M$, time shift $\delta$, temperature range $\rho$
\STATE With probability $1-p$, set $M \gets 1$ \Comment{Single-pass update}
\STATE $t \sim \mathcal{U}[(M-1)\delta, 1]$, \; $\mathbf{z}_t \sim q_t(\cdot \mid \mathbf{x})$ \Comment{Corrupt the data~\Eqn{eq:udm-forward}}
\STATE Sample $T$ with range $\rho$ \Comment{Temperature regularization~\Eqn{eq:law-temperature}}
\STATE $\mathbf{c} \gets \mathbf{z}_t E_\theta$ \Comment{Embed the noisy tokens}
\FOR{$m = 1$ \TO $M-1$}
    \STATE $\mathbf{c} \gets \mathrm{sg}\big[\statemap_\theta^{\mathrm{E.E.}}\big(f_\theta(\mathbf{c}, t-(m-1)\delta);\, T\big)\big]$ \Comment{Pass without gradient~\Eqn{eq:law-unroll}}
\ENDFOR
\STATE $\hat{\mathbf{x}} \gets \mathrm{softmax}\big(g_\theta(f_\theta(\mathbf{c}, t-(M-1)\delta))\big)$ \Comment{Last pass, with gradient}
\STATE Take a gradient step on $\sum_{\ell=1}^{L}\mathcal{L}_{\mathrm{LAW}}^{\supl}$ \Comment{NELBO or CE~\Eqn{eq:law-loss}}
\end{algorithmic}
\end{algorithm}

\subsection{Comparison to Loopholing Discrete Diffusion Models (LDDMs)}
\label{app:loophole-comparison}
\method and LDDMs both propagate an SC state across denoising steps. They differ in how the backbone uses this state, which affects both sampling and training, and in the state map that computes it~\Eqn{eq:law-state-map}.

\paragraph{Sampling.} At each step, an LDDM samples discrete tokens from the posterior $q_{s|t}$ and feeds the backbone the sum of their embeddings and the SC state~\Eqn{eq:loopholing}. \method instead feeds the backbone the SC state alone~\Eqn{eq:law-transition}. It never samples tokens between steps and decodes only once, after the final step~\Eqn{eq:law-decoding}.

\paragraph{Training.} Both methods corrupt the data once, $\mathbf{z}_t\sim q_t$. With probability $p$, both run several forward passes, stop the gradient through the SC state between passes, and apply the loss to the last pass only. Otherwise, both run a single pass (\sec{sec:bg-loopholing}). LDDMs run two passes at the same time $t$: the first pass receives $\mathbf{z}_t E_\theta$ with a zero SC state and computes the SC state $\mathbf{c}_t$, and the second pass receives $\mathbf{z}_t E_\theta+\mathbf{c}_t$. In contrast, the later passes of \method receive the SC state alone and run at successive times $t^{(m)}$, and the number of passes $M$ can exceed two~\Eqn{eq:law-unroll}.

\paragraph{State Map.} LDDMs propagate the layer-normalized hidden representation ($\statemap_\theta^{\mathrm{Hid.}}$), while \method propagates a convex combination of token embeddings, weighted by the predicted token probabilities ($\statemap_\theta^{\mathrm{E.E.}}$)~\Eqn{eq:law-state-map}.

\begin{figure}[!ht]
    \centering
    \includegraphics[width=\linewidth]{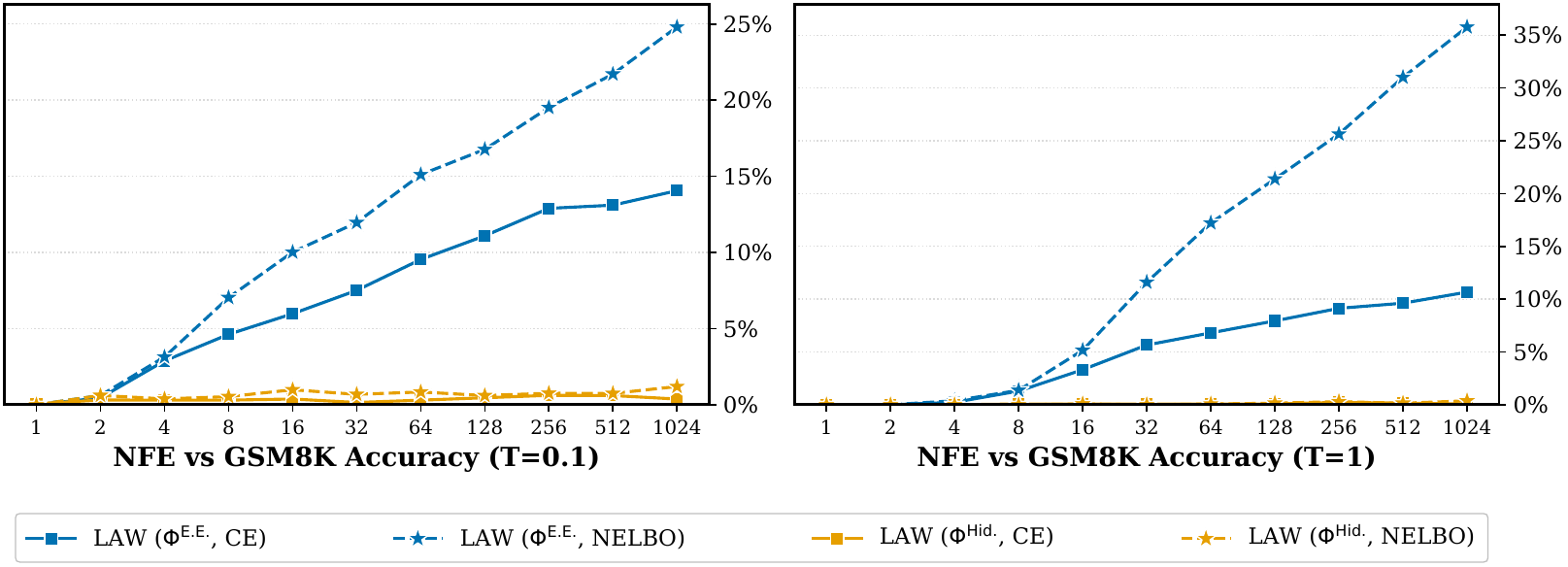}
    \caption{\textbf{SC state and loss.} \method with the Expected Embedding or the hidden representation as SC state, trained with the NELBO or the CE, without regularization.}
    \label{fig:app-gsm-state-loss}
\end{figure}

\paragraph{Why LDDMs Do Not Learn to Ignore Their Tokens.} LDDMs rely mainly on their tokens. When we set the token embeddings of the response to zero during sampling, they solve almost no GSM8K problems. Without their SC state, a configuration they see during training, their accuracy drops by half or more, but remains well above zero (\supp{app:tinygsm-lddm-tokens}). We see three reasons. First, LDDMs train on $\mathbf{z}_t\sim q_t(\cdot\mid\mathbf{x})$, which keeps each clean token with probability $\alpha_t$ (\sec{sec:bg-loopholing}), so the loss rewards relying on the tokens. Second, LDDMs feed the backbone the sum $\mathbf{z}_t E_\theta+\mathbf{c}_t$~\Eqn{eq:loopholing}. To generate from the SC state alone, they would have to learn to ignore the token embeddings in this sum, but neural networks struggle to ignore uninformative input features~\citep{ng2004featureselectionl1l2, grinsztajn2022treebasedmodelsstilloutperform, damario2022dataefficiencydeeplearning, shi2023largelanguagemodelseasily}. Third, \method with the state map of LDDMs ($\statemap_\theta^{\mathrm{Hid.}}$) solves almost no problems, unlike with $\statemap_\theta^{\mathrm{E.E.}}$ (\Fig{fig:app-gsm-state-loss}). A likely reason is that the Expected Embedding is a convex combination of token embeddings~\Eqn{eq:law-state-map}, so the backbone input stays in their convex hull, with no such guarantee for the hidden state.

\subsection{Gradient Flow Between Passes}
\label{app:backprop-loops}
In \Eqn{eq:law-unroll}, a stop-gradient is applied to the SC state fed to the next pass, $\mathbf{c}^{(m+1)} = \mathrm{sg}\big[\statemap_\theta^{\mathrm{E.E.}}(\mathbf{h}^{(m)};T)\big]$, where $\mathbf{h}^{(m)} = f_\theta(\mathbf{c}^{(m)}, t^{(m)})$ is the final-layer hidden representation of the $m$-th pass. In practice, we apply it to the predicted probabilities inside $\statemap_\theta^{\mathrm{E.E.}}$, so the token embeddings $E_\theta$ still receive a gradient through the SC state fed to the last pass. As a result, the loss~\Eqn{eq:law-loss} only updates the parameters through the last forward pass and through $E_\theta$ in this SC state, and the earlier passes run without gradient.

Alternatively, we can remove the stop-gradient and set $\mathbf{c}^{(m+1)} = \statemap_\theta^{\mathrm{E.E.}}(\mathbf{h}^{(m)};T)$. The gradient of the loss then also flows through the SC states into the earlier passes. For $M=2$, the gradient with respect to the output of the first pass is
\begin{equation}
    \frac{\partial \mathcal{L}_{\mathrm{LAW}}}{\partial \mathbf{h}^{(1)}}
    =
    \frac{\partial \mathcal{L}_{\mathrm{LAW}}}{\partial \mathbf{c}^{(2)}}\,
    \frac{\partial \statemap_\theta^{\mathrm{E.E.}}(\mathbf{h}^{(1)};T)}{\partial \mathbf{h}^{(1)}}.
\end{equation}
The stop-gradient sets this term to zero. Without it, the first pass learns to produce SC states that help the second pass. We use the stop-gradient by default and compare both variants on TinyGSM (\Fig{fig:app-gsm-backprop}).

\subsection{Mode Embedding}
\label{app:mode-emb}
The DiT conditions every block on the noise level through $\boldsymbol{\nu} = \operatorname{SiLU}(\operatorname{Emb}(\alpha_t))$, from which each block computes the scales, shifts, and gates of its adaLN-Zero modulation \citep{peebles2023scalablediffusionmodelstransformers}. At the first pass~\Eqn{eq:law-unroll} and the first sampling step, the denoiser receives token embeddings, whereas at every later pass it receives the SC state. The mode embedding tells the denoiser which of the two it is processing. We learn two vectors, $\mathbf{m}_{\mathrm{tok}}$ for token embeddings and $\mathbf{m}_{\mathrm{SC}}$ for SC states, and add the matching one to the noise-level conditioning, \eg $\boldsymbol{\nu} = \operatorname{SiLU}(\operatorname{Emb}(\alpha_t)) + \mathbf{m}_{\mathrm{SC}}$ at the later passes. We initialize $\mathbf{m}_{\mathrm{tok}}=\mathbf{m}_{\mathrm{SC}}=\mathbf{0}$, so a checkpoint without mode embedding, such as Duo, computes the same function after loading.

\section{Training Compute}
\label{app:compute}
We compare all methods at matched training FLOPs. \citet{jo2025loopholingdiscretediffusion} instead train LDDMs and their base models for the same number of updates, although each LDDM update is more expensive. This section states the matching rule and lists the training steps of every LDDM and \method run.

\subsection{FLOP Matching}
\label{app:flop-matching}
Let $F$ be the cost of one forward pass of the denoiser, and count a backward pass as $2F$. A standard update runs one forward and one backward pass, and costs $3F$. An update with $M$ passes and a stop-gradient between them~\Eqn{eq:law-unroll} runs $M-1$ passes without gradient and a last pass with gradient, and costs $(M+2)F$. Without the stop-gradient, the backward pass goes through all $M$ passes, and the update costs $3MF$ (\supp{app:backprop-loops}). LDDMs and \method run $M$ passes with probability $p$ and a single pass otherwise, so an update costs on average
\begin{equation}
    \label{eq:flop-cost}
    C = \big(3 + p(M-1)\big)F \quad \text{with the stop-gradient,} \qquad C = \big(3 + 3p(M-1)\big)F \quad \text{without it.}
\end{equation}
To match the FLOPs of $S$ baseline steps at the same batch size, a run trains for $3FS/C$ steps. For example, LDDMs and \method with $M=2$ and $p=0.9$ cost $C=3.9F$ per update and train for 77\% of the steps of the baselines.

\paragraph{Fine-Tuning.} When we fine-tune from Duo on TinyGSM (\supp{app:gsm8k-details}), we count the FLOPs of the 250k Duo steps ($3F$ each) and of the fine-tuning updates, both at batch size 512. With $M=2$ and $p=0.9$, fine-tuning adds 8\% to the compute of Duo after 15k steps and 26\% after 50k steps. With $M=4$ and $p=0.75$, 50k steps add 35\%.

\subsection{Training Steps}
\label{app:training-steps}
\tab{tab:app-compute-summary} gives the budget of the baselines and the default budget of LDDMs and \method for each task. \tab{tab:app-sudoku-steps}, \tab{tab:app-tinygsm-steps}, \tab{tab:app-maze-hard-steps}, and \tab{tab:app-owt-steps} list the configuration and training steps of every LDDM and \method run. In these tables, E.E.\ is the Expected Embedding and Hid.\ the hidden representation of LDDMs, a row with several values (difficulties or $\rho$) stands for one run per value with the same number of steps, and each run is listed once.

\begin{table}[h]
\centering
\small
\caption{\textbf{Training budget per task.} The baselines run standard updates ($3F$ each). By default, LDDMs and \method run $M=2$ passes with $p=0.9$ ($3.9F$ per update) and train for fewer steps to match the training FLOPs of the baselines~\Eqn{eq:flop-cost}.}
\label{tab:app-compute-summary}
\begin{tabular}{@{}lrrr@{}}
\toprule
Task & Batch & Baseline steps & LDDM / \method steps \\
\midrule
Sudoku      & 256 & 20k  & 15.4k \\
TinyGSM     & 512 & 250k & 192k  \\
Maze        & 256 & 100k & 76.9k \\
OpenWebText & 512 & 400k & 308k  \\
\bottomrule
\end{tabular}
\end{table}

\begin{table}[h]
\centering
\small
\setlength{\tabcolsep}{3pt}
\caption{\textbf{Training steps of the \method runs on Sudoku, grouped by experiment.} All runs are trained from scratch and match the training FLOPs of 20k baseline steps at batch size 256. Difficulty \emph{all} stands for easy, medium, and hard. Runs marked $\star$ appear in \tab{tab:sudoku}.}
\label{tab:app-sudoku-steps}
\begin{tabular}{@{}llllcccccrrr@{}}
\toprule
Method                             & Difficulty & SC state & Loss  & $M$ & $p$  & Stop-grad. & $\rho$  & Mode emb. & Batch & Cost    &    Steps \\
\midrule
\multicolumn{12}{@{}l}{\textit{Loss}} \\
\hspace*{0.8em}\method{}$^\star$   & all        & E.E.     & CE    & 2   & 0.9  & yes        & none    & no        & 256   & $3.9F$  & 15{,}385 \\
\hspace*{0.8em}\method{}$^\star$   & all        & E.E.     & NELBO & 2   & 0.9  & yes        & none    & no        & 256   & $3.9F$  & 15{,}385 \\
\midrule
\multicolumn{12}{@{}l}{\textit{Multi-pass probability (with the $p=0.9$ runs above)}} \\
\hspace*{0.8em}\method{}$^\star$   & all        & E.E.     & CE    & 2   & 0.75 & yes        & none    & no        & 256   & $3.75F$ & 16{,}000 \\
\hspace*{0.8em}\method             & hard       & E.E.     & NELBO & 2   & 0.75 & yes        & none    & no        & 256   & $3.75F$ & 16{,}000 \\
\hspace*{0.8em}\method             & hard       & E.E.     & CE    & 2   & 0.5  & yes        & none    & no        & 256   & $3.5F$  & 17{,}143 \\
\hspace*{0.8em}\method             & hard       & E.E.     & NELBO & 2   & 0.5  & yes        & none    & no        & 256   & $3.5F$  & 17{,}143 \\
\hspace*{0.8em}\method             & hard       & E.E.     & CE    & 2   & 0.25 & yes        & none    & no        & 256   & $3.25F$ & 18{,}462 \\
\hspace*{0.8em}\method             & hard       & E.E.     & NELBO & 2   & 0.25 & yes        & none    & no        & 256   & $3.25F$ & 18{,}462 \\
\hspace*{0.8em}\method             & hard       & E.E.     & CE    & 2   & 0.1  & yes        & none    & no        & 256   & $3.1F$  & 19{,}355 \\
\hspace*{0.8em}\method             & hard       & E.E.     & NELBO & 2   & 0.1  & yes        & none    & no        & 256   & $3.1F$  & 19{,}355 \\
\midrule
\multicolumn{12}{@{}l}{\textit{Number of passes}} \\
\hspace*{0.8em}\method             & hard       & E.E.     & CE    & 3   & 0.9  & yes        & none    & no        & 256   & $4.8F$  & 12{,}500 \\
\hspace*{0.8em}\method             & hard       & E.E.     & NELBO & 3   & 0.9  & yes        & none    & no        & 256   & $4.8F$  & 12{,}500 \\
\hspace*{0.8em}\method             & hard       & E.E.     & CE    & 3   & 0.75 & yes        & none    & no        & 256   & $4.5F$  & 13{,}333 \\
\hspace*{0.8em}\method             & hard       & E.E.     & NELBO & 3   & 0.75 & yes        & none    & no        & 256   & $4.5F$  & 13{,}333 \\
\midrule
\multicolumn{12}{@{}l}{\textit{Temperature regularization (with the corresponding runs above)}} \\
\hspace*{0.8em}\method{}$^\star$   & all        & E.E.     & CE    & 2   & 0.9  & yes        & 2       & no        & 256   & $3.9F$  & 15{,}385 \\
\hspace*{0.8em}\method             & hard       & E.E.     & CE    & 2   & 0.9  & yes        & 3--5    & no        & 256   & $3.9F$  & 15{,}385 \\
\hspace*{0.8em}\method             & hard       & E.E.     & NELBO & 2   & 0.9  & yes        & 2--5    & no        & 256   & $3.9F$  & 15{,}385 \\
\hspace*{0.8em}\method             & all        & E.E.     & CE    & 2   & 0.75 & yes        & 3       & no        & 256   & $3.75F$ & 16{,}000 \\
\hspace*{0.8em}\method             & hard       & E.E.     & CE    & 2   & 0.75 & yes        & 2, 4, 5 & no        & 256   & $3.75F$ & 16{,}000 \\
\hspace*{0.8em}\method             & hard       & E.E.     & NELBO & 2   & 0.75 & yes        & 2--5    & no        & 256   & $3.75F$ & 16{,}000 \\
\hspace*{0.8em}\method             & hard       & E.E.     & CE    & 3   & 0.9  & yes        & 2--5    & no        & 256   & $4.8F$  & 12{,}500 \\
\hspace*{0.8em}\method             & hard       & E.E.     & NELBO & 3   & 0.75 & yes        & 2--5    & no        & 256   & $4.5F$  & 13{,}333 \\
\bottomrule
\end{tabular}
\end{table}

\begin{table}[h]
\centering
\small
\setlength{\tabcolsep}{4pt}
\caption{\textbf{Training steps of the LDDM and \method runs on TinyGSM, grouped by the figures that show them.} Runs trained from scratch match the training FLOPs of 250k baseline steps at batch size 512. Each fine-tuning configuration starts from the Duo baseline and is trained at batch sizes 64, 128, 256 and 512. The cost per update is given at batch size 512. The run marked $\star$ is \method from scratch in \Fig{fig:gsm8k-nfe}, \Fig{fig:app-gsm-early-stop} and \tab{tab:app-tinygsm-few-step}, and $\dagger$ marks \method without regularization. Outside \Fig{fig:app-gsm-finetune}, the fine-tuned \method is the run with $M=2$ at batch size 512.}
\label{tab:app-tinygsm-steps}
\begin{tabular}{@{}lllcccccrrr@{}}
\toprule
Method & SC state & Loss & $M$ & $p$ & Stop-grad. & $\rho$ & Mode emb. & Batch & Cost & Steps \\
\midrule
\multicolumn{11}{@{}l}{\textit{LDDMs (\fig{fig:gsm8k-nfe})}} \\
\hspace*{0.8em}LDDM-U & Hid. & NELBO & 2 & 0.9  & yes & none & no  & 512 & $3.9F$  & 192k \\
\hspace*{0.8em}LDDM-M & Hid. & NELBO & 2 & 0.9  & yes & none & no  & 512 & $3.9F$  & 192k \\
\midrule
\multicolumn{11}{@{}l}{\textit{SC state and loss (\fig{fig:app-gsm-state-loss})}} \\
\hspace*{0.8em}\method{}$^\dagger$ & E.E. & NELBO & 2 & 0.9  & yes & none & no  & 512 & $3.9F$  & 192k \\
\hspace*{0.8em}\method & E.E. & CE    & 2 & 0.9  & yes & none & no  & 512 & $3.9F$  & 192k \\
\hspace*{0.8em}\method & Hid. & NELBO & 2 & 0.9  & yes & none & no  & 512 & $3.9F$  & 192k \\
\hspace*{0.8em}\method & Hid. & CE    & 2 & 0.9  & yes & none & no  & 512 & $3.9F$  & 192k \\
\midrule
\multicolumn{11}{@{}l}{\textit{Temperature regularization and mode embedding (\fig{fig:app-gsm-temp-mode}, with $\dagger$)}} \\
\hspace*{0.8em}\method & E.E. & NELBO & 2 & 0.9  & yes & none & yes & 512 & $3.9F$  & 192k \\
\hspace*{0.8em}\method{}$^\star$ & E.E. & NELBO & 2 & 0.9  & yes & 3    & no  & 512 & $3.9F$  & 192k \\
\hspace*{0.8em}\method & E.E. & NELBO & 2 & 0.9  & yes & 3    & yes & 512 & $3.9F$  & 192k \\
\hspace*{0.8em}\method & E.E. & NELBO & 2 & 0.9  & yes & 4    & no  & 512 & $3.9F$  & 192k \\
\hspace*{0.8em}\method & E.E. & NELBO & 2 & 0.9  & yes & 4    & yes & 512 & $3.9F$  & 192k \\
\midrule
\multicolumn{11}{@{}l}{\textit{Number of passes and multi-pass probability (\fig{fig:app-gsm-passes}, with $\star$)}} \\
\hspace*{0.8em}\method & E.E. & NELBO & 2 & 0.75 & yes & 3    & no  & 512 & $3.75F$ & 200k \\
\hspace*{0.8em}\method & E.E. & NELBO & 3 & 0.9  & yes & 3    & no  & 512 & $4.8F$  & 156k \\
\hspace*{0.8em}\method & E.E. & NELBO & 3 & 0.75 & yes & 3    & no  & 512 & $4.5F$  & 167k \\
\hspace*{0.8em}\method & E.E. & NELBO & 4 & 0.9  & yes & 3    & no  & 512 & $5.7F$  & 132k \\
\hspace*{0.8em}\method & E.E. & NELBO & 4 & 0.75 & yes & 3    & no  & 512 & $5.25F$ & 143k \\
\midrule
\multicolumn{11}{@{}l}{\textit{Gradient flow between passes (Figs.~\ref{fig:app-gsm-backprop} and~\ref{fig:app-gsm-backprop-steps}, with $\star$)}} \\
\hspace*{0.8em}\method & E.E. & NELBO & 2 & 0.9  & no  & 3    & no  & 512 & $5.7F$  & 132k \\
\hspace*{0.8em}\method & E.E. & NELBO & 2 & 1    & no  & 3    & no  & 512 & $6F$    & 125k \\
\midrule
\multicolumn{11}{@{}l}{\textit{Expected Embedding in LDDM-U (Figs.~\ref{fig:app-gsm-pure-sc} and~\ref{fig:gsm8k-ablations} left, with LDDM-U, $\star$ and $\dagger$)}} \\
\hspace*{0.8em}LDDM-U & E.E. & NELBO & 2 & 0.9  & yes & none & no  & 512 & $3.9F$  & 192k \\
\hspace*{0.8em}LDDM-U & E.E. & NELBO & 2 & 0.9  & yes & 3    & no  & 512 & $3.9F$  & 192k \\
\midrule
\multicolumn{11}{@{}l}{\textit{Fine-tuning from Duo (Figs.~\ref{fig:gsm8k-nfe}, \ref{fig:gsm8k-ablations} right, \ref{fig:app-gsm-finetune} and~\ref{fig:app-gsm-early-stop})}} \\
\hspace*{0.8em}\method & E.E. & NELBO & 2 & 0.9  & yes & 3    & no  & 64--512 & $3.9F$  & 50k \\
\hspace*{0.8em}\method & E.E. & NELBO & 4 & 0.75 & yes & 3    & no  & 64--512 & $5.25F$ & 50k \\
\bottomrule
\end{tabular}
\end{table}

\begin{table}[h]
\centering
\small
\setlength{\tabcolsep}{4pt}
\caption{\textbf{Training steps of the \method runs on Maze-Hard, grouped by experiment.} All runs are trained from scratch and match the training FLOPs of 100k baseline steps at batch size 256. The run marked $\star$ is \method in \Fig{fig:maze-nfe}, and $\dagger$ marks the same run without regularization.}
\label{tab:app-maze-hard-steps}
\begin{tabular}{@{}lllcccccrrr@{}}
\toprule
Method & SC state & Loss & $M$ & $p$ & Stop-grad. & $\rho$ & Mode emb. & Batch & Cost & Steps \\
\midrule
\multicolumn{11}{@{}l}{\textit{Loss}} \\
\hspace*{0.8em}\method & E.E. & CE    & 2 & 0.9  & yes & none & no & 256 & $3.9F$  & 76{,}923 \\
\hspace*{0.8em}\method & E.E. & NELBO & 2 & 0.9  & yes & none & no & 256 & $3.9F$  & 76{,}923 \\
\midrule
\multicolumn{11}{@{}l}{\textit{Multi-pass probability (with the $p=0.9$ runs above)}} \\
\hspace*{0.8em}\method{}$^\dagger$ & E.E. & NELBO & 2 & 0.75 & yes & none & no & 256 & $3.75F$ & 80{,}000 \\
\hspace*{0.8em}\method & E.E. & CE    & 2 & 0.75 & yes & none & no & 256 & $3.75F$ & 80{,}000 \\
\hspace*{0.8em}\method & E.E. & CE    & 2 & 0.5  & yes & none & no & 256 & $3.5F$  & 85{,}714 \\
\hspace*{0.8em}\method & E.E. & NELBO & 2 & 0.5  & yes & none & no & 256 & $3.5F$  & 85{,}714 \\
\hspace*{0.8em}\method & E.E. & CE    & 2 & 0.25 & yes & none & no & 256 & $3.25F$ & 92{,}308 \\
\hspace*{0.8em}\method & E.E. & NELBO & 2 & 0.25 & yes & none & no & 256 & $3.25F$ & 92{,}308 \\
\hspace*{0.8em}\method & E.E. & CE    & 2 & 0.1  & yes & none & no & 256 & $3.1F$  & 96{,}774 \\
\hspace*{0.8em}\method & E.E. & NELBO & 2 & 0.1  & yes & none & no & 256 & $3.1F$  & 96{,}774 \\
\midrule
\multicolumn{11}{@{}l}{\textit{Number of passes}} \\
\hspace*{0.8em}\method & E.E. & CE    & 3 & 0.9  & yes & none & no & 256 & $4.8F$  & 62{,}500 \\
\hspace*{0.8em}\method & E.E. & NELBO & 3 & 0.9  & yes & none & no & 256 & $4.8F$  & 62{,}500 \\
\hspace*{0.8em}\method & E.E. & CE    & 3 & 0.75 & yes & none & no & 256 & $4.5F$  & 66{,}667 \\
\hspace*{0.8em}\method & E.E. & NELBO & 3 & 0.75 & yes & none & no & 256 & $4.5F$  & 66{,}667 \\
\midrule
\multicolumn{11}{@{}l}{\textit{Temperature regularization (with the corresponding runs above)}} \\
\hspace*{0.8em}\method & E.E. & CE    & 2 & 0.9  & yes & 2--5 & no & 256 & $3.9F$  & 76{,}923 \\
\hspace*{0.8em}\method & E.E. & NELBO & 2 & 0.9  & yes & 2--5 & no & 256 & $3.9F$  & 76{,}923 \\
\hspace*{0.8em}\method & E.E. & CE    & 2 & 0.75 & yes & 2--5 & no & 256 & $3.75F$ & 80{,}000 \\
\hspace*{0.8em}\method{}$^\star$ & E.E. & NELBO & 2 & 0.75 & yes & 2 & no & 256 & $3.75F$ & 80{,}000 \\
\hspace*{0.8em}\method & E.E. & NELBO & 2 & 0.75 & yes & 3--5 & no & 256 & $3.75F$ & 80{,}000 \\
\hspace*{0.8em}\method & E.E. & CE    & 3 & 0.9  & yes & 2--5 & no & 256 & $4.8F$  & 62{,}500 \\
\hspace*{0.8em}\method & E.E. & NELBO & 3 & 0.9  & yes & 2--5 & no & 256 & $4.8F$  & 62{,}500 \\
\bottomrule
\end{tabular}
\end{table}

\begin{table}[h]
\centering
\small
\setlength{\tabcolsep}{4pt}
\caption{\textbf{Training steps of the LDDM and \method runs on OpenWebText.} All runs are trained from scratch and match the training FLOPs of 400k baseline steps at batch size 512. The run marked $\star$ is \method in \Fig{fig:owt_entropy_x_gen_ppl}.}
\label{tab:app-owt-steps}
\begin{tabular}{@{}lllcccccrrr@{}}
\toprule
Method & SC state & Loss & $M$ & $p$ & Stop-grad. & $\rho$ & Mode emb. & Batch & Cost & Steps \\
\midrule
\multicolumn{11}{@{}l}{\textit{LDDMs (\Fig{fig:owt_entropy_x_gen_ppl} and \Fig{fig:owt-frontier-lddm})}} \\
\hspace*{0.8em}LDDM-M & Hid. & NELBO & 2 & 0.9 & yes & none & no  & 512 & $3.9F$ & 308k \\
\hspace*{0.8em}LDDM-U & Hid. & NELBO & 2 & 0.9 & yes & none & no  & 512 & $3.9F$ & 308k \\
\hspace*{0.8em}LDDM-U & E.E. & NELBO & 2 & 0.9 & yes & none & no  & 512 & $3.9F$ & 308k \\
\hspace*{0.8em}LDDM-U & E.E. & NELBO & 2 & 0.9 & yes & 3    & no  & 512 & $3.9F$ & 308k \\
\midrule
\multicolumn{11}{@{}l}{\textit{\method (\Fig{fig:owt-loss} and \Fig{fig:owt-temp-mode})}} \\
\hspace*{0.8em}\method & E.E. & NELBO & 2 & 0.9 & yes & none & no  & 512 & $3.9F$ & 308k \\
\hspace*{0.8em}\method & E.E. & CE    & 2 & 0.9 & yes & none & no  & 512 & $3.9F$ & 308k \\
\hspace*{0.8em}\method & E.E. & CE    & 2 & 0.9 & yes & 3    & no  & 512 & $3.9F$ & 308k \\
\hspace*{0.8em}\method & E.E. & CE    & 2 & 0.9 & yes & 4    & no  & 512 & $3.9F$ & 308k \\
\hspace*{0.8em}\method{}$^\star$ & E.E. & CE & 2 & 0.9 & yes & 3 & yes & 512 & $3.9F$ & 308k \\
\bottomrule
\end{tabular}
\end{table}

\section{Experimental Setup}
\label{app:exp-details}

\subsection{Shared Setup}
\label{app:shared-setup}
\paragraph{Training Hyperparameters.}
\label{app:training-hparams}
All models are trained with AdamW with a learning rate of $3\times10^{-4}$, $(\beta_1,\beta_2)=(0.9,0.999)$, zero weight decay, 2.5k warmup steps followed by a constant learning rate, and gradient clipping at $1.0$. We maintain an exponential moving average of the weights with decay $0.9999$ and use it for evaluation. All DiT backbones use dropout $0.1$.

\paragraph{Baselines.}
\label{app:baselines}
We compare to the following baselines, using the subset given in each experiment. Beyond the shared hyperparameters above, we use the default hyperparameters of the original implementations unless stated otherwise.
\begin{itemize}
\item \textbf{ARM.} Causal Transformer trained with next-token prediction on the target, conditioned on the full prompt when there is one (\eg the unsolved grid on Maze).
\item \textbf{MDLM} \citep{sahoo2024simpleeffectivemaskeddiffusion}. Masked Diffusion with the SUBS parameterization, the log-linear schedule, and no time conditioning.
\item \textbf{Duo} \citep{sahoo2025diffusionduality}. Uniform Diffusion with the log-linear schedule and time conditioning.
\item \textbf{LDDM} \citep{jo2025loopholingdiscretediffusion}. We describe LDDM-M and LDDM-U in \sec{sec:bg-loopholing} and compare them with \method in \supp{app:loophole-comparison}.
\item \textbf{FLM} \citep{lee2026onesteplanguagemodelingcontinuous}. Gaussian flow on one-hot token representations, with time conditioning on $t\in[0,1]$.
\item \textbf{CANDI} \citep{pynadath2025candihybriddiscretecontinuousdiffusion}. Hybrid discrete--continuous diffusion with mixing coefficient $0.5$, $\sigma_{\min}=0.01$, $\sigma_{\max}=2$, and percentile scheduling with bounds $0.01$ and $0.25$.
\item \textbf{$\mathbb{S}$-FLM} and \textbf{$\mathbb{S}$-FLM ($\mathbb{S}$-arch.)} \citep{deschenaux2026hypersphericalflows}. Spherical flow trained with the CE loss and the log-linear schedule truncated at $\alpha_{\max}=\alpha^\star(0.1)$, which depends on the vocabulary size $|\mathcal{V}|$ and the embedding dimension $d$ \citep[Eq.~(17)]{deschenaux2026hypersphericalflows}. $\mathbb{S}$-FLM uses the standard DiT, whereas $\mathbb{S}$-FLM ($\mathbb{S}$-arch.) uses the Sphere-DiT backbone with nGPT initialization.
\end{itemize}

\paragraph{Architecture.}
All diffusion and flow models, including $\mathbb{S}$-FLM, use the standard DiT \citep{peebles2023scalablediffusionmodelstransformers} with RoPE \citep{su2023roformerenhancedtransformerrotary}, except FLM and CANDI, which use the DiT variants of their original implementations at the same size, and $\mathbb{S}$-FLM ($\mathbb{S}$-arch.), which uses the Sphere-DiT. The ARM uses the same Transformer with causal attention and without time conditioning. \method uses the same DiT as Duo, so a Duo checkpoint can be loaded into \method without any change.

\paragraph{Sampling.}
Unless stated otherwise, the methods use the following samplers:
\begin{itemize}
\item \textbf{MDLM}, \textbf{Duo}: ancestral sampling with the posterior of each process, sampling the tokens at the last step.
\item \textbf{LDDM}: ancestral sampling where the SC state of each step is passed to the next one, sampling the tokens at the last step.
\item \textbf{FLM}: Euler integration of the flow. When there is a prompt, it is projected back to its clean value at the last step.
\item \textbf{CANDI}: the CANDI sampler of the original implementation.
\item \textbf{$\mathbb{S}$-FLM}, \textbf{$\mathbb{S}$-FLM ($\mathbb{S}$-arch.)}: the sampling setup of the original implementation. At each step, the velocity only points toward the embedding of the most likely token (top-1 velocity), and the tokens are decoded with an argmax at the last step. $\mathbb{S}$-FLM therefore does not depend on the sampling temperature, and appears as the same curve at both temperatures in \Fig{fig:gsm8k-nfe}.
\item \textbf{\method}: \algo{alg:law-sampling} with the raw decoder $\eta_{\mathrm{raw}}$~\Eqn{eq:law-decoding}.
\end{itemize}

\subsection{Sudoku}
\label{app:sudoku-details}
\paragraph{Dataset.}
Following prior work \citep{deschenaux2026hypersphericalflows, kim2025finetuningmaskeddiffusionprovable}, we generate 48k training and 2k validation puzzles with unique solutions \citep{alp2024sudoku}. The puzzles have three difficulty levels, determined by the number of visible digits: easy (40/81), medium (35/81), and hard (30/81).

\paragraph{Tokenization.}
Each Sudoku grid is tokenized as a 1D sequence: the partial grid, a beginning-of-sequence token, then the solution. Missing digits are encoded with an empty-cell token, and rows are separated by a row-separator token. The vocabulary contains 12 tokens, and every sequence has length 180. The partial grid serves as the prompt and is never corrupted during training.

\paragraph{Architecture.}
LDDMs and \method use an 8-layer DiT with embedding dimension $d=512$ and 8 attention heads.

\paragraph{Training.}
LDDMs and \method are trained from scratch on a single GPU at batch size 256, with the shared hyperparameters of \supp{app:training-hparams}. LDDMs train for the FLOP-matched 15{,}385 steps, and \method for the steps of \tab{tab:app-sudoku-steps}. We train \method with the NELBO and the CE, with $M\in\{2,3\}$ passes, multi-pass probabilities $p\in\{0.1,0.25,0.5,0.75,0.9\}$, and temperature regularization with $\rho\in\{2,3,4,5\}$ (\supp{app:sudoku-ablations}). The results of the other baselines in \tab{tab:sudoku} are taken from prior work \citep{deschenaux2026hypersphericalflows}, which trains them for 20k steps at batch size 256.

\paragraph{Sampling.}
We sample with 180 NFE. The baselines use the samplers of \supp{app:baselines}, and \method uses \algo{alg:law-sampling} with the $\eta_{\mathrm{raw}}$ decoder~\Eqn{eq:law-decoding}.

\paragraph{Metric.}
We report the \emph{exact-match} accuracy, i.e., the fraction of the 2k validation puzzles for which the generated grid equals the unique solution.

\subsection{TinyGSM}
\label{app:gsm8k-details}
\paragraph{Dataset.}
We train on TinyGSM \citep{liu2023tinygsm}, which pairs 11.8M synthetic grade-school math questions with Python programs that solve them, and evaluate on the GSM8K test set \citep{cobbe2021trainingverifierssolvemath}.

\paragraph{Tokenization.}
We tokenize with the \texttt{HuggingFaceTB/SmolLM-135M} tokenizer and discard examples longer than 512 tokens. Shorter examples are padded to 512 tokens. Each sequence is a question--program pair: the question serves as the prompt, and the loss is applied only to the program.

\paragraph{Architecture.}
All diffusion and flow models use a 12-layer DiT with embedding dimension $d=768$ and 12 attention heads.

\paragraph{Training.}
Every model is trained from scratch on 8 GPUs at global batch size 512, with the shared hyperparameters of \supp{app:training-hparams}. The ARM, MDLM, Duo, FLM, CANDI, and $\mathbb{S}$-FLM ($\mathbb{S}$-arch.) train for 250k steps, and LDDMs and \method for the FLOP-matched steps of \tab{tab:app-tinygsm-steps}. CANDI reaches 0.2\% accuracy with 1024 NFE, close to FLM, and we leave it out of the figures for readability. LDDM-U with the Expected Embedding (\sec{sec:experiments-gsm8k}) adds the SC state to the token embeddings and runs both passes at the same time $t$, like the standard LDDM-U. Unlike it, it resets the SC state at the prompt positions to the embeddings of the clean prompt.

\paragraph{Fine-Tuning.}
We fine-tune the Duo baseline (250k steps) with \method for 50k steps, starting from its online weights, while the Exponential Moving Average (EMA) of the weights continues from that of Duo. The optimizer and the learning rate schedule, including the 2.5k warmup steps, start anew. We fine-tune on 4 GPUs at global batch sizes 64, 128, 256 and 512, and evaluate every 5k steps. \Fig{fig:gsm8k-nfe} reports the run with $M=2$, $p=0.9$ and batch size 512 after 50k steps, and \Fig{fig:app-gsm-early-stop} compares it with its checkpoint after 15k steps. \supp{app:flop-matching} gives the training compute of fine-tuning.

\paragraph{Sampling.}
For each GSM8K test problem, we generate and execute one program. We sweep the NFE over $\{1, 2, 4, \dots, 1024\}$ at the sampling temperatures $T\in\{1, 0.1\}$, using the last checkpoint of each run trained from scratch. The baselines use the samplers of \supp{app:baselines}, and \method uses \algo{alg:law-sampling} with the $\eta_{\mathrm{raw}}$ decoder~\Eqn{eq:law-decoding}. LDDM-U with the Expected Embedding uses the LDDM sampler, except that the sampling temperature also applies to the state map, as in \method. The accuracy over fine-tuning steps is measured with 256 NFE at $T=1$.

\paragraph{Metric.}
A generated program is correct if executing it returns the reference answer. We report the accuracy over the GSM8K test set.

\subsection{Maze}
\label{app:maze-details}
\paragraph{Dataset.}
We regenerate Maze-Hard \citep{wang2025hierarchicalreasoning} on $20\times20$ grids with an existing maze generator \citep{lehnert2024abetterplanningtransformers}. For each maze, we draw a wall probability from $\mathcal{U}[0.3,0.5]$, and each cell is independently a \texttt{wall} with this probability. The \texttt{start} and the \texttt{goal} are placed uniformly at random on two distinct \texttt{empty} cells. We keep only solvable mazes whose shortest path contains at least 71 cells, including the \texttt{start} and the \texttt{goal}. The original Maze-Hard \citep{wang2025hierarchicalreasoning} uses $30\times30$ grids and a threshold of 110 cells. To shorten the sequences from 1800 to 800 tokens, we use $20\times20$ grids. On our grids, a threshold of 71 cells keeps approximately the same fraction of generated mazes. We generate 1{,}000 training and 1{,}000 evaluation mazes with seed 42. The shortest paths of the evaluation mazes contain between 70 and 89 moves (71 to 90 cells), with a median of 72. The target is one shortest path, found by breadth-first search with a fixed neighbor order. We verify all 2{,}000 targets independently of the generator by recomputing the shortest path length with the NetworkX library \citep{SciPyProceedings_11}, and every target passes this check.

\paragraph{Tokenization.}
Each maze is tokenized as a 1D sequence: the unsolved grid followed by the solved grid, each flattened row by row, from top to bottom and left to right. Each cell is one of five tokens: \texttt{wall}, \texttt{empty}, \texttt{start}, \texttt{goal}, or \texttt{path}. With an additional padding token, the vocabulary contains 6 tokens, plus a mask token for MDLM, LDDM-M, CANDI, and the ARM. Every sequence has length $2\times20^2 = 800$. The unsolved grid serves as the prompt: it is never corrupted during training, the loss is computed only on the solved grid, and the sampler resets the prompt to its clean value at every step.

\paragraph{Architecture.}
All models use an 8-layer DiT with embedding dimension $d=256$ and 8 attention heads. For $\mathbb{S}$-FLM ($\mathbb{S}$-arch.), $|\mathcal{V}|=6$ and $d=256$ give $\alpha_{\max}=0.121$.

\paragraph{Training.}
Every model is trained from scratch on a single GPU at batch size 256, with the shared hyperparameters of \supp{app:training-hparams}. The baselines train for 100k steps, and LDDMs and \method for the FLOP-matched steps of \tab{tab:app-maze-hard-steps}. We train \method with the NELBO and the CE, with $M\in\{2,3\}$ passes, multi-pass probabilities $p\in\{0.1,0.25,0.5,0.75,0.9\}$, and temperature regularization with $\rho\in\{2,3,4,5\}$ (\supp{app:maze-ablations}).

\paragraph{Sampling.}
We evaluate on the 1{,}000 evaluation mazes. Unless stated otherwise, all diffusion and flow models sample with 800 NFE at sampling temperature $T=1$. The ARM decodes the 400 cells of the solution grid one at a time with a Key-Value (KV) cache, i.e., with 400 NFE. For the ARM, we report both greedy and sampled decoding. The baselines use the samplers of \supp{app:baselines}, and \method uses \algo{alg:law-sampling} with the $\eta_{\mathrm{raw}}$ decoder~\Eqn{eq:law-decoding}. We sweep the NFE over $\{1, 2, 4, \dots, 512, 800\}$ with all other settings fixed.

\paragraph{Metrics.}
\label{app:maze-metrics}
Prior work reports \emph{exact-match} accuracy on Maze-Hard \citep{wang2025hierarchicalreasoning}. Since a maze can have several shortest paths, \emph{exact match} can underestimate the accuracy: a correct shortest path that differs from the target is counted as wrong (\supp{app:maze-exact-vs-valid}). We therefore report the \emph{valid-solution} accuracy, i.e., the fraction of evaluation mazes for which the generated grid (i) reproduces every \texttt{wall}, \texttt{empty}, \texttt{start}, and \texttt{goal} cell of the maze, and (ii) marks a shortest path from the \texttt{start} to the \texttt{goal}. To check (ii), let $n$ be the number of cells on the target path (i.e., the length of a shortest path), including the \texttt{start} and the \texttt{goal}. We check that the cells labeled \texttt{path} lie on \texttt{empty} cells of the maze and, together with the \texttt{start} and the \texttt{goal}, form a connected path of exactly $n$ cells.

\subsection{OpenWebText}
\label{app:owt-details}
\paragraph{Dataset and Tokenization.}
We train on OpenWebText \citep{Gokaslan2019OpenWeb} with the GPT-2 tokenizer \citep{Radford2019LanguageMA} and a context length of 1024.

\paragraph{Architecture.}
All diffusion and flow models use a 12-layer DiT with embedding dimension $d=768$ and 12 attention heads. The reported \method model also uses the mode embedding (\supp{app:mode-emb}).

\paragraph{Training.}
We train on 16 GPUs at global batch size 512, with the shared hyperparameters of \supp{app:training-hparams}. The baselines train for 400k steps, and every LDDM and \method run, ablations included, for the FLOP-matched 308k steps (\tab{tab:app-owt-steps}). We compare the NELBO and the CE~\Eqn{eq:law-loss}, and ablate temperature regularization with $\rho\in\{3,4\}$~\Eqn{eq:law-temperature} and the mode embedding (\supp{app:owt-ablations}). As on TinyGSM, we also train LDDM-U with the Expected Embedding as its SC state, with and without temperature regularization ($\rho=3$).

\paragraph{Sampling.}
For the Gen.\ PPL / entropy frontiers of \Fig{fig:owt_entropy_x_gen_ppl} and \Fig{fig:owt-frontier-lddm}, we sample with 64 and 1024 NFE and sweep the sampling temperature over $T \in \{0.70, 0.75, \dots, 1.20\}$ for all models. \supp{app:owt-samples} shows samples drawn with 64 and 1024 NFE. \method uses \algo{alg:law-sampling} with the $\eta_{\mathrm{raw}}$ decoder~\Eqn{eq:law-decoding}. Since the ARM generates one token per NFE, we only sample it with 1024 NFE.

\paragraph{Generative Perplexity.}
\label{app:genppl-entropy}
We measure sample quality with the Generative Perplexity (Gen.\ PPL), following prior work \citep{lou2024discretediffusionmodelingestimating, sahoo2024simpleeffectivemaskeddiffusion, deschenaux2025partitiongenerativemodelingmasked, deschenaux2026hypersphericalflows}. The Gen.\ PPL measures how well a reference model, here GPT-2 Large~\citep{Radford2019LanguageMA}, predicts each token of a generated sample from the previous tokens. We follow the evaluation protocol of $\mathbb{S}$-FLM \citep{deschenaux2026hypersphericalflows}. For each sampling temperature and NFE, we draw $K=512$ samples of length $L=1024$. Following the code of MDLM~\citep{sahoo2024simpleeffectivemaskeddiffusion}, we decode each sample, re-tokenize it and evaluate it with GPT-2 Large, ignoring the end-of-text tokens after the first one and the last token of the sequence. With $m^{(k,\ell)}=1$ if token $\ell$ of sample $\mathbf{x}^{(k)}$ is scored and $0$ otherwise, we report the perplexity over all scored tokens of the $K$ samples:
\begin{equation}
    \mathrm{PPL}_{\mathrm{gen}} = \exp\left(-\frac{\sum_{k=1}^{K}\sum_{\ell=2}^{L-1} m^{(k,\ell)} \log p^{\text{GPT-2}}(\mathbf{x}^{(k),\supl} \mid \mathbf{x}^{(k),<\supl})}{\sum_{k=1}^{K}\sum_{\ell=2}^{L-1} m^{(k,\ell)}}\right).
\end{equation}

\paragraph{Unigram Entropy.} Repetitive text can reach a low Gen.\ PPL~\citep{dieleman2022continuousdiffusioncategoricaldata}. We detect such degenerate samples with the mean unigram entropy
\begin{equation}
    H = -\frac{1}{K}\sum_{k=1}^{K}\sum_{v\in\mathcal{V}} \frac{c(v, \mathbf{x}^{(k)})}{L}\log\frac{c(v, \mathbf{x}^{(k)})}{L},
\end{equation}
where $c(v, \mathbf{x}^{(k)})$ counts token $v$ in sample $\mathbf{x}^{(k)}$, end-of-text tokens included. Repetitive samples have a low entropy.

\paragraph{Plotted Range.} Sweeping $T$ gives one curve per method in \Fig{fig:owt_entropy_x_gen_ppl} and \Fig{fig:owt-frontier-lddm}. We keep points with $4.5 \leq H \leq 6.0$ and $\mathrm{PPL}_{\mathrm{gen}} \leq 500$, dropping degenerate settings.

\paragraph{Hidden-State Correlation.}
\label{app:hidden-corr}
As in prior work \citep{geiping2025scalingtesttimecomputelatent}, we measure representation collapse~\citep{noci2022signalpropagationtransformerstheoretical} with the correlation of the hidden states across positions. We take the final-layer hidden representation $\mathbf{h}\in\mathbb{R}^{L\times d}$~\Eqn{eq:denoiser-chain}, i.e., the input to the LM head $g_\theta$, at the final sampling step. We center each position over the embedding dimension, $\tilde{\mathbf{h}}^{\supl} = \mathbf{h}^{\supl} - \frac{1}{d}\sum_{j=1}^{d} h^{\supl}_j\,\mathbf{1}$, and compute the mean cosine similarity over all pairs of distinct positions,
\begin{equation}
    r = \frac{1}{L(L-1)}\sum_{\ell\neq\ell'} \frac{\langle \tilde{\mathbf{h}}^{\supl}, \tilde{\mathbf{h}}^{\supl'} \rangle}{\lVert \tilde{\mathbf{h}}^{\supl} \rVert\, \lVert \tilde{\mathbf{h}}^{\supl'} \rVert}.
\end{equation}
We report $r$ averaged over the samples. A value close to 1 means that all positions share the same hidden state, i.e., the representations have collapsed.

\paragraph{Evaluation During Training.}
For the ablations of \supp{app:owt-ablations}, we evaluate every 5k training steps with the EMA weights. We compute the Gen.\ PPL and the unigram entropy from 512 samples drawn with 256 NFE at $T=1$. We also compute the validation loss (the \method training loss~\Eqn{eq:law-loss} on the validation split) and the hidden-state correlation.

\FloatBarrier
\section{Additional Results}
\label{app:additional-results}

\subsection{Sudoku: Ablations}
\label{app:sudoku-ablations}
All Sudoku ablations use hard puzzles, since \method already reaches about 99\% on easy and medium puzzles (\tab{tab:sudoku}). We report the \emph{exact-match} accuracy with 180 NFE at $T=1$, at the last checkpoint of each run. \tab{tab:app-sudoku-steps} lists the configuration and training steps of every run.

\paragraph{Loss.} The CE is more accurate than the NELBO in every configuration (\tab{tab:app-sudoku-pass-probability}, \tab{tab:app-sudoku-passes}, and \tab{tab:app-sudoku-temperature}). The best CE run solves 96.5\% of the hard puzzles, against 88.9\% for the best NELBO run.

\paragraph{Multi-Pass Probability.} $p=0.75$ is the best value for both losses, with 94.5\% for the CE and 88.9\% for the NELBO (\tab{tab:app-sudoku-pass-probability}). With two passes on only 10\% of the updates ($p=0.1$), the CE solves only 84.6\% of the puzzles.

\paragraph{Number of Passes.} A third pass only helps with the CE and $p=0.9$, from 92.2\% to 94.5\% (\tab{tab:app-sudoku-passes}). With the NELBO, $M=3$ is clearly less accurate than $M=2$ (75.4\% against 85.3\% with $p=0.9$).

\paragraph{Temperature Regularization.} With the best $\rho$, temperature regularization improves every CE configuration, up to 96.5\% with $\rho=2$, against 92.2\% for the same run without regularization (\tab{tab:app-sudoku-temperature}). With the NELBO, its effect depends on the configuration. For $M=3$ with $p=0.75$, it improves the accuracy from 79.6\% to 86.5\%, but no value of $\rho$ improves $M=2$ with $p=0.75$.

\begin{table}[!htb]
\centering
\footnotesize
\setlength{\tabcolsep}{4pt}
\caption{\textbf{Multi-pass probability on hard Sudoku.} \method with $M=2$ passes run with probability $p$, without regularization. Best per column in \textbf{bold}.}
\label{tab:app-sudoku-pass-probability}
\newcommand{\tabrow}{\hspace*{0.8em}}
\begin{tabular}{@{}lcc@{}}
\toprule
Training         &            CE &         NELBO \\
\midrule
\tabrow $p=0.9$  &          92.2 &          85.3 \\
\tabrow $p=0.75$ & \textbf{94.5} & \textbf{88.9} \\
\tabrow $p=0.5$  &          93.4 &          84.3 \\
\tabrow $p=0.25$ &          90.1 &          84.9 \\
\tabrow $p=0.1$  &          84.6 &          79.3 \\
\bottomrule
\end{tabular}
\end{table}

\begin{table}[!htb]
\centering
\footnotesize
\setlength{\tabcolsep}{4pt}
\caption{\textbf{Number of passes on hard Sudoku.} \method with $M$ passes run with probability $p$, without regularization. Best per column in \textbf{bold}.}
\label{tab:app-sudoku-passes}
\newcommand{\tabrow}{\hspace*{0.8em}}
\begin{tabular}{@{}lcc@{}}
\toprule
Training                &            CE &         NELBO \\
\midrule
\tabrow $M=2$, $p=0.9$  &          92.2 &          85.3 \\
\tabrow $M=3$, $p=0.9$  & \textbf{94.5} &          75.4 \\
\tabrow $M=2$, $p=0.75$ & \textbf{94.5} & \textbf{88.9} \\
\tabrow $M=3$, $p=0.75$ &          94.2 &          79.6 \\
\bottomrule
\end{tabular}
\end{table}

\begin{table}[!htb]
\centering
\footnotesize
\setlength{\tabcolsep}{4pt}
\caption{\textbf{Temperature regularization on hard Sudoku.} Each row is one training configuration, without regularization (none) and with range $\rho$~\Eqn{eq:law-temperature}. Best per row in \textbf{bold}.}
\label{tab:app-sudoku-temperature}
\newcommand{\tabrow}{\hspace*{0.8em}}
\begin{tabular}{@{}lccccc@{}}
\toprule
Training                       &          none &      $\rho=2$ &      $\rho=3$ &      $\rho=4$ & $\rho=5$ \\
\midrule
\tabrow CE, $M=2$, $p=0.9$     &          92.2 & \textbf{96.5} &          94.3 &          95.2 &     94.5 \\
\tabrow NELBO, $M=2$, $p=0.9$  &          85.3 &          84.5 &          85.2 & \textbf{87.1} &     86.7 \\
\tabrow CE, $M=2$, $p=0.75$    &          94.5 &          95.4 & \textbf{96.2} &          94.6 &     94.5 \\
\tabrow NELBO, $M=2$, $p=0.75$ & \textbf{88.9} &          88.5 &          87.8 &          86.0 &     85.9 \\
\tabrow CE, $M=3$, $p=0.9$     &          94.5 &          94.5 &          94.9 & \textbf{95.1} &     93.3 \\
\tabrow NELBO, $M=3$, $p=0.75$ &          79.6 & \textbf{86.5} &          83.1 &          83.5 &     81.1 \\
\bottomrule
\end{tabular}
\end{table}

\FloatBarrier
\subsection{TinyGSM: Ablations}
\label{app:tinygsm-ablations}
This section shows every TinyGSM ablation of \sec{sec:experiments-gsm8k}, which discusses the main findings. We only add the details that the main text omits. Unless stated otherwise, the figures show the GSM8K accuracy against NFE at $T=0.1$ (left) and $T=1$ (right). All runs are trained from scratch with the settings of \supp{app:gsm8k-details}, and we evaluate the last checkpoint of each run. \tab{tab:app-tinygsm-steps} lists the configuration and training steps of every run.

\begin{figure}[!ht]
    \centering
    \includegraphics[width=\linewidth]{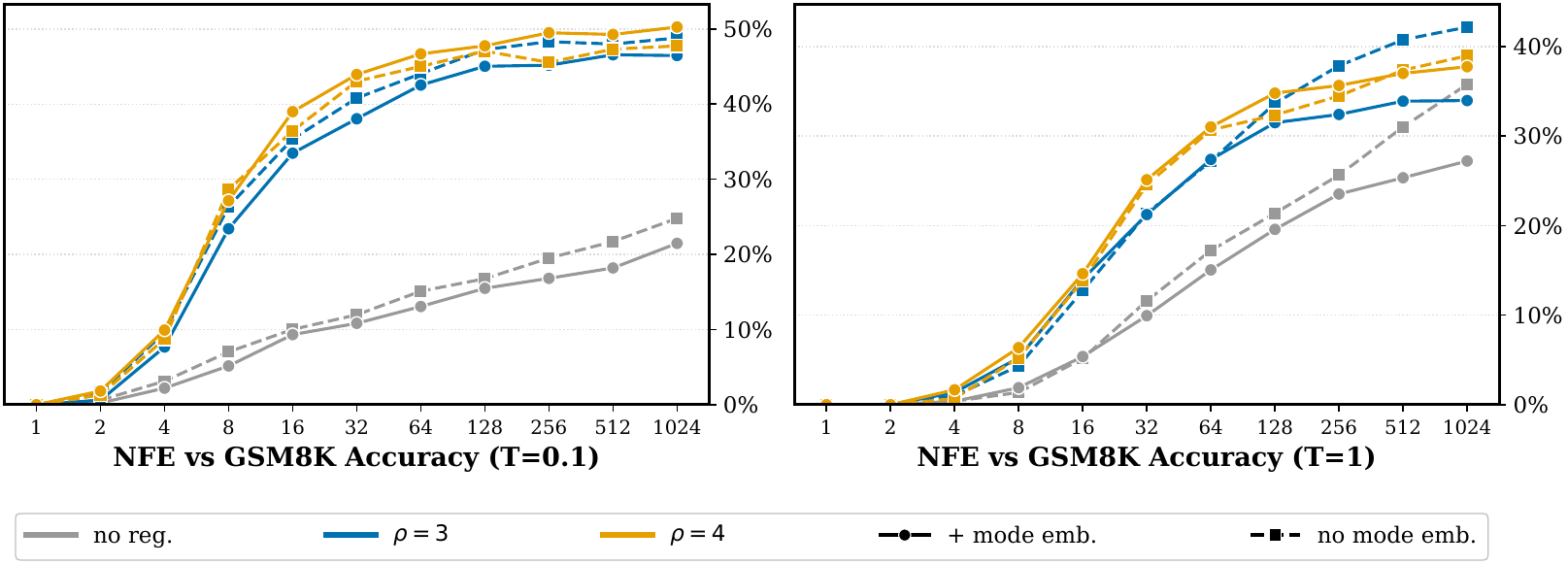}
    \caption{\textbf{Temperature regularization and mode embedding.} \method with temperature regularization ($\rho\in\{3,4\}$), the mode embedding (\supp{app:mode-emb}), both, or neither.}
    \label{fig:app-gsm-temp-mode}
\end{figure}

\paragraph{Temperature Regularization and Mode Embedding.} $\rho=4$ gives results similar to $\rho=3$ (\Fig{fig:app-gsm-temp-mode}). The mode embedding brings no consistent gain. With 1024 NFE, it lowers the accuracy at $T=1$ in every configuration. At $T=0.1$, the accuracies with and without it differ by at most 3.3\%.

\begin{figure}[!ht]
    \centering
    \includegraphics[width=\linewidth]{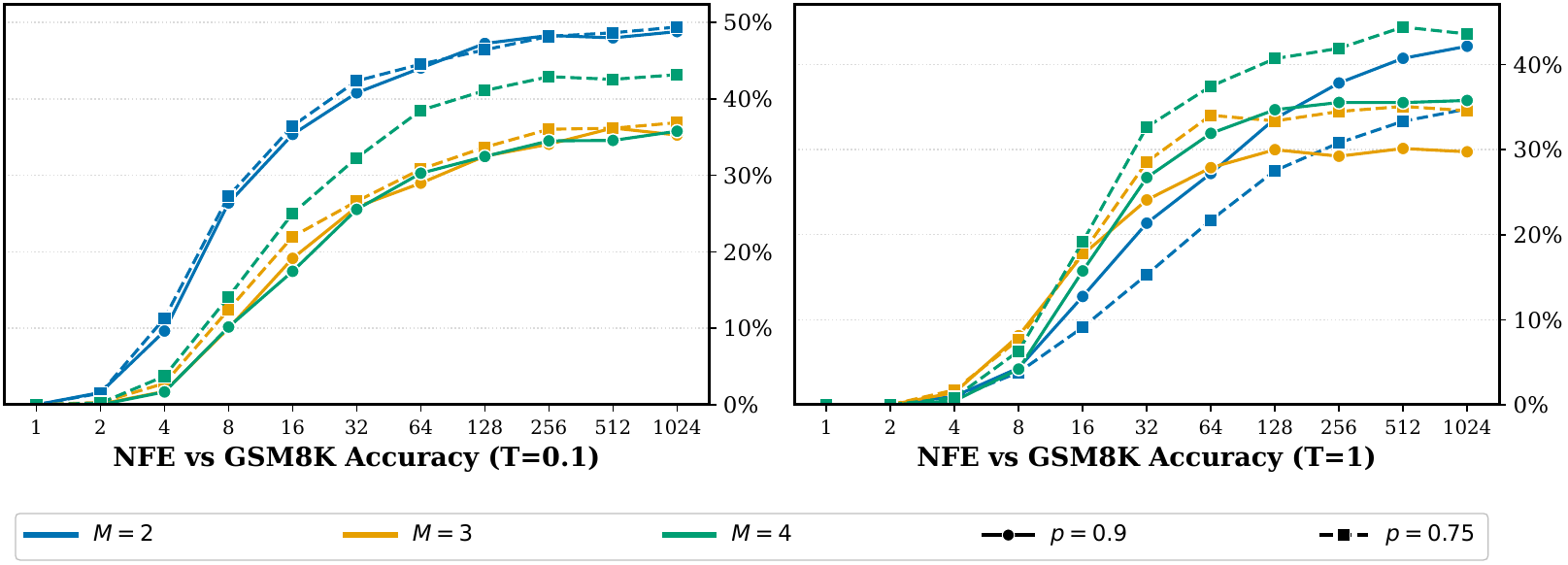}
    \caption{\textbf{Number of passes and multi-pass probability.} \method with $M\in\{2,3,4\}$ passes, run with probability $p\in\{0.75,0.9\}$, and with $\rho=3$. Runs with more passes or a higher $p$ train for fewer steps, to match the training FLOPs.}
    \label{fig:app-gsm-passes}
\end{figure}

\paragraph{Number of Passes and Multi-Pass Probability.} At $T=0.1$, $M=2$ is the most accurate, with 48.8\% ($p=0.9$) and 49.4\% ($p=0.75$) with 1024 NFE, against at most 43.2\% with more passes (\Fig{fig:app-gsm-passes}). At $T=1$, $M=4$ with $p=0.75$ is also the most accurate with 1024 NFE (43.6\% against 42.2\% for $M=2$ with $p=0.9$).

\begin{figure}[!ht]
    \centering
    \includegraphics[width=\linewidth]{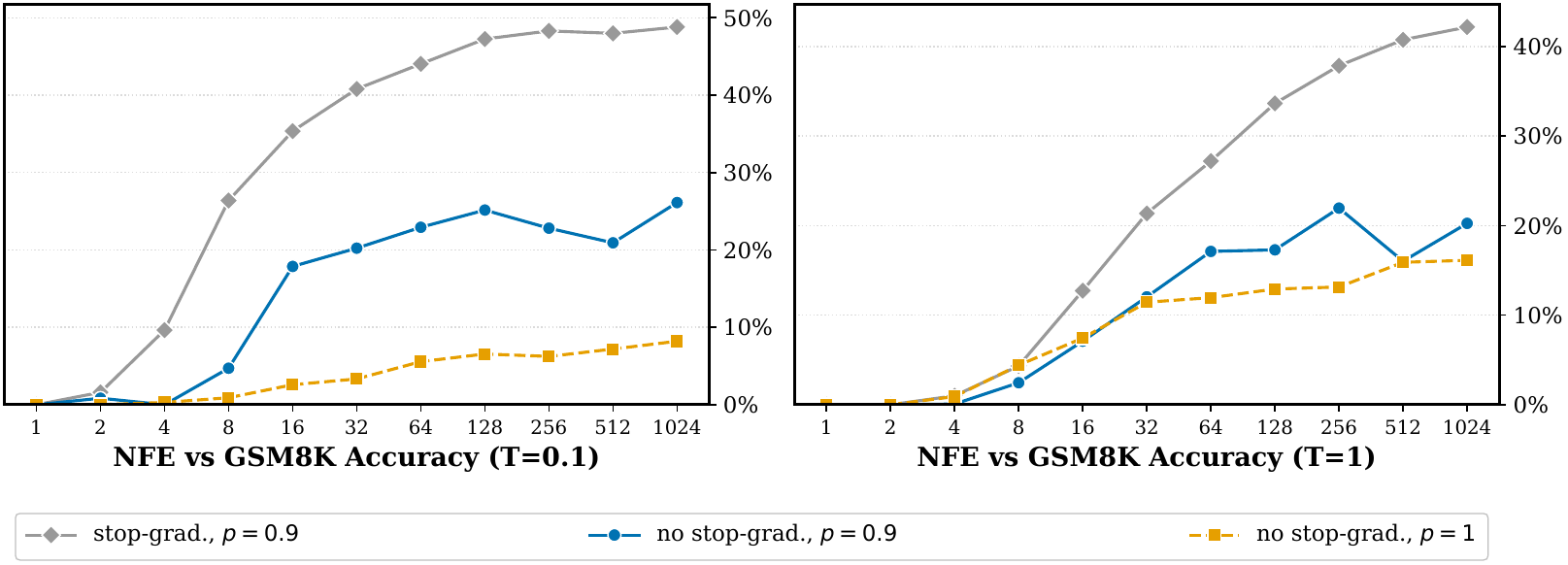}
    \caption{\textbf{Gradient flow between passes.} \method with $M=2$ and $\rho=3$, with the stop-gradient between passes ($p=0.9$) and without it ($p\in\{0.9,1\}$). Without the stop-gradient, each update is more expensive, so these runs train for 132k ($p=0.9$) and 125k ($p=1$) steps, against 192k with the stop-gradient.}
    \label{fig:app-gsm-backprop}
\end{figure}

\begin{figure}[!ht]
    \centering
    \includegraphics[width=0.6\linewidth]{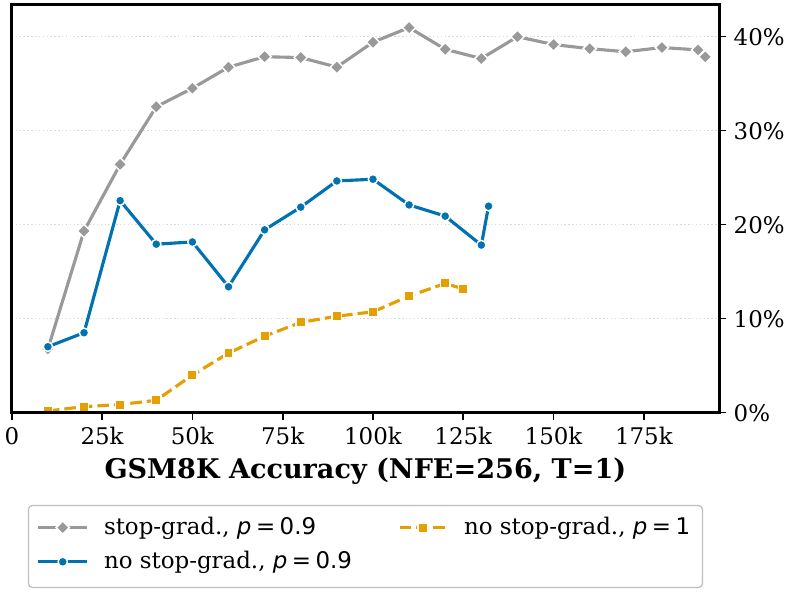}
    \caption{\textbf{Gradient flow between passes, over training.} GSM8K accuracy with 256 NFE at $T=1$ over training steps, for the runs of \Fig{fig:app-gsm-backprop}. After 120k steps, the accuracy is 38.7\% with the stop-gradient, against 20.9\% ($p=0.9$) and 13.7\% ($p=1$) without it. With the stop-gradient, the accuracy plateaus from about 70k steps. Without it, the run with $p=0.9$ does not improve after 100k steps, and the run with $p=1$ improves slowly.}
    \label{fig:app-gsm-backprop-steps}
\end{figure}

\paragraph{Gradient Flow Between Passes.} Without the stop-gradient (\supp{app:backprop-loops}), \method reaches 20.2\% ($p=0.9$) and 16.1\% ($p=1$) with 1024 NFE at $T=1$, against 42.2\% with it (\Fig{fig:app-gsm-backprop} and \Fig{fig:app-gsm-backprop-steps}).

\begin{figure}[!ht]
    \centering
    \includegraphics[width=\linewidth]{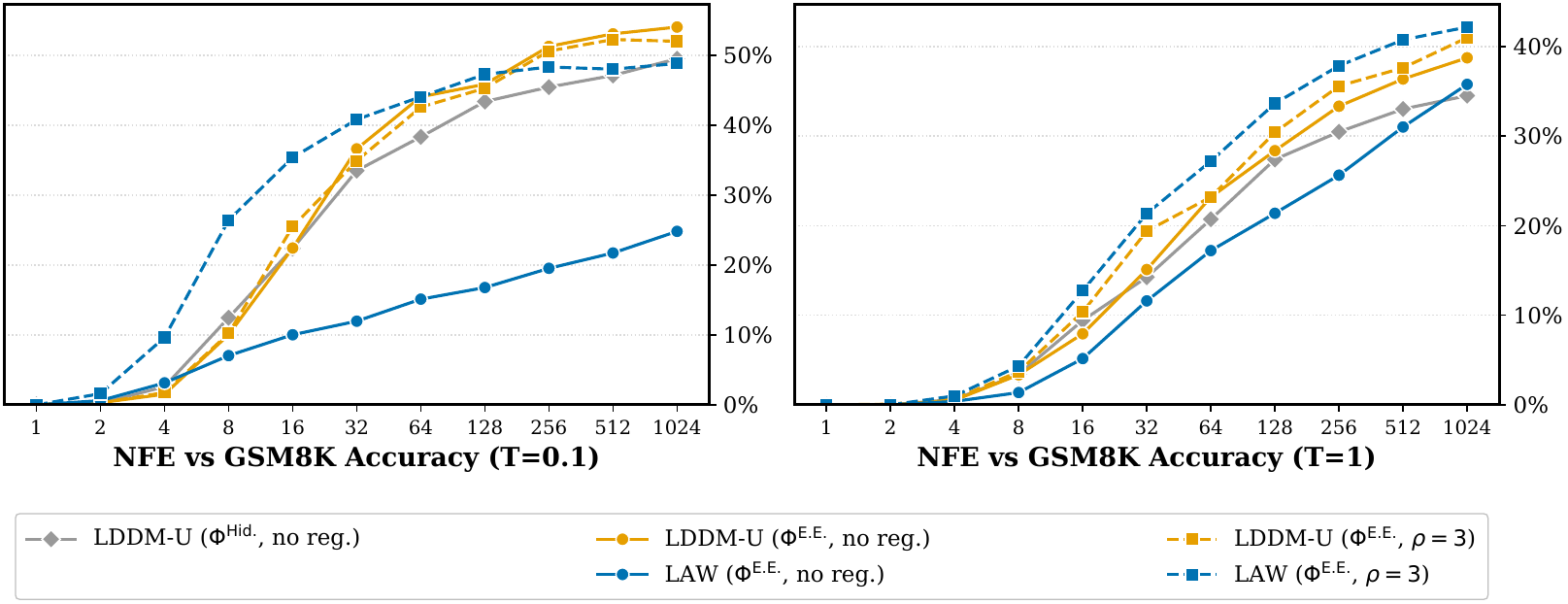}
    \caption{\textbf{\method against LDDM-U with the Expected Embedding.} The legend gives the SC state and the temperature regularization ($\rho=3$ or none).}
    \label{fig:app-gsm-pure-sc}
\end{figure}

\paragraph{Expected Embedding in LDDM-U.} \Fig{fig:app-gsm-pure-sc} shows both temperatures for the comparison of the left panel of \Fig{fig:gsm8k-ablations}, which only shows $T=0.1$.

\FloatBarrier
\subsection{TinyGSM: Fine-Tuning from Duo}
\label{app:tinygsm-finetune}
\begin{figure}[!ht]
    \centering
    \includegraphics[width=0.6\linewidth]{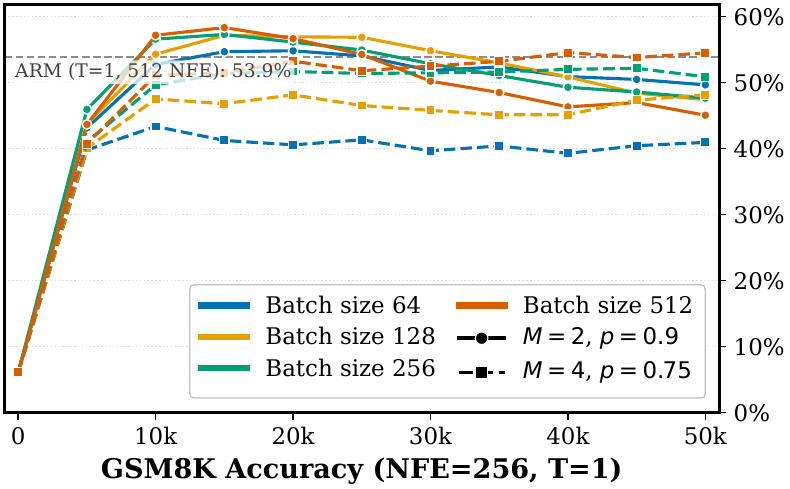}
    \caption{\textbf{Fine-tuning from Duo.} GSM8K accuracy over fine-tuning steps with 256 NFE at $T=1$, for $M=2$, $p=0.9$ (solid) and $M=4$, $p=0.75$ (dashed), both with $\rho=3$, at batch sizes 64 to 512. Step 0 is the Duo checkpoint, sampled with the sampler of \method. The dashed gray line is the ARM sampled at $T=1$ (53.9\%, 512 NFE).}
    \label{fig:app-gsm-finetune}
\end{figure}

\begin{figure}[!ht]
    \centering
    \includegraphics[width=\linewidth]{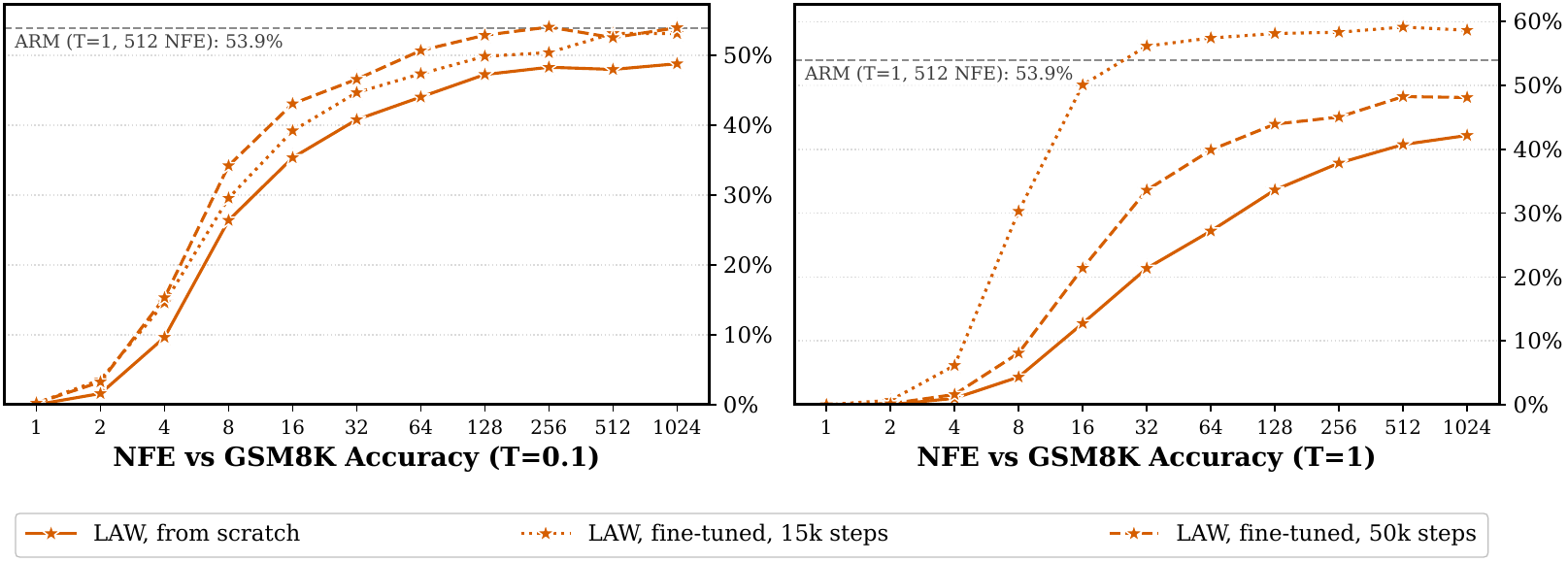}
    \caption{\textbf{Early stopping.} \method fine-tuned from Duo with $M=2$, $p=0.9$, $\rho=3$ and batch size 512, after 15k and 50k fine-tuning steps, against \method trained from scratch (\Fig{fig:gsm8k-nfe}). The dashed gray line is the ARM, as in \Fig{fig:app-gsm-finetune}.}
    \label{fig:app-gsm-early-stop}
\end{figure}

With $M=2$, the accuracy peaks after 15k to 20k fine-tuning steps and then decreases, whereas with $M=4$ it does not decrease (\Fig{fig:app-gsm-finetune}, all batch sizes). \Fig{fig:app-gsm-early-stop} compares the $M=2$ checkpoints after 15k and 50k steps against NFE (\sec{sec:experiments-gsm8k}).

\FloatBarrier
\subsection{TinyGSM: Hard-Token SC States and LDDM-U}
\label{app:tinygsm-hard-tokens}
To separate the effect of the soft SC state from that of avoiding sampling between steps, we replace the Expected Embedding $\mathbf{p}E_\theta$, with $\mathbf{p}=\operatorname{softmax}(g_\theta(\mathbf{h})/T)$~\Eqn{eq:law-state-map}, by the row of $E_\theta$ of a single token at each position. This token is either the argmax token $\arg\max_v p_v$ or a token drawn from $\mathbf{p}$. As with $\statemap_\theta^{\mathrm{E.E.}}$, no gradient flows through $\mathbf{p}$, but $E_\theta$ receives a gradient through the embedding lookup. We fine-tune the Duo baseline with each SC state, and with LDDM-U, for 50k steps at batch size 64. \method uses $M=2$ and $p=0.9$, as the Expected Embedding run at batch size 64 of \Fig{fig:app-gsm-finetune}. The temperature does not change the argmax token, so we train it without temperature regularization, and the sampled token with $\rho=3$.

\begin{figure}[!ht]
    \centering
    \includegraphics[width=0.6\linewidth]{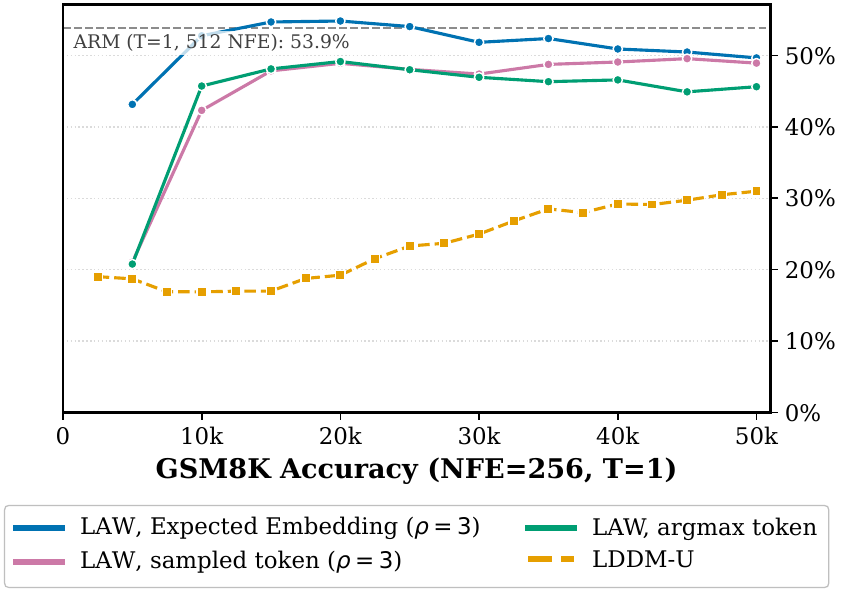}
    \caption{\textbf{Hard-token SC states and LDDM-U, fine-tuned from Duo.} GSM8K accuracy over fine-tuning steps with 256 NFE at $T=1$. \method uses the Expected Embedding, a sampled token or the argmax token as SC state. All runs use batch size 64, and the Expected Embedding run is that of \Fig{fig:app-gsm-finetune}. The dashed gray line is the ARM, as in \Fig{fig:app-gsm-finetune}.}
    \label{fig:app-gsm-hard-tokens}
\end{figure}

With the Expected Embedding, \method is the most accurate at every checkpoint (\Fig{fig:app-gsm-hard-tokens}). The gap to the hard-token SC states is largest early in fine-tuning and shrinks towards the end. The sampled and argmax tokens are similarly accurate during the first 30k steps, after which the argmax token is less accurate. The argmax token avoids sampling between steps but remains less accurate than the Expected Embedding, so the accuracy of \method does not come only from avoiding the sampling noise. LDDM-U adapts slowly. Its accuracy first decreases, then increases until the end of fine-tuning, but stays below that of \method with every SC state.

\FloatBarrier
\subsection{TinyGSM: LDDMs Rely on Their Discrete Tokens}
\label{app:tinygsm-lddm-tokens}
\begin{figure}[!ht]
    \centering
    \includegraphics[width=\linewidth]{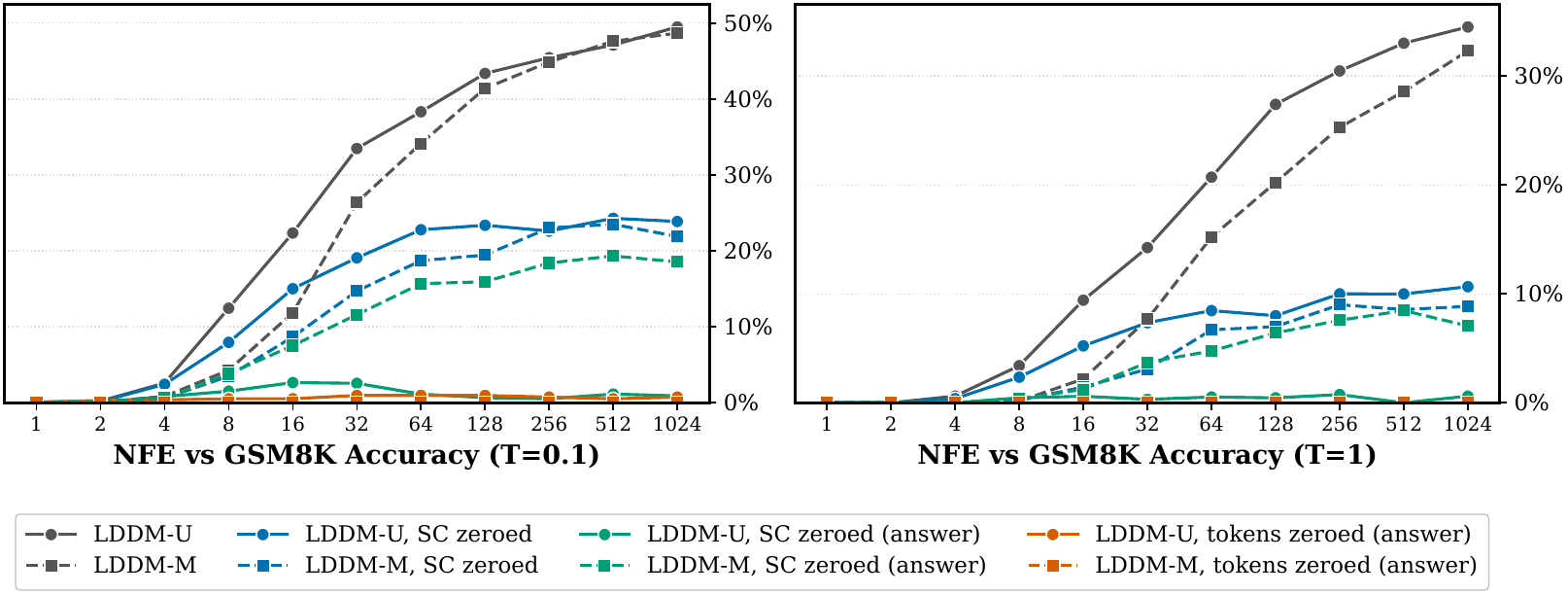}
    \caption{\textbf{LDDMs with zeroed inputs on GSM8K.} Accuracy against NFE at $T=0.1$ (left) and $T=1$ (right) of LDDM-U and LDDM-M trained from scratch (\Fig{fig:gsm8k-nfe}), when the SC state or the token embeddings are set to zero during sampling. Curves marked (answer) zero only the response positions.}
    \label{fig:app-lddm-zero}
\end{figure}

We evaluate how much LDDM-U and LDDM-M, trained from scratch on TinyGSM (\Fig{fig:gsm8k-nfe}), rely on their discrete tokens and on their SC state. During sampling, we set to zero (i) the SC state at all positions, (ii) the SC state at the response positions, or (iii) the token embeddings at the response positions, and sample at $T\in\{0.1, 1\}$ with 1 to 1024 NFE (\Fig{fig:app-lddm-zero}). With 1024 NFE at $T=1$, LDDM-M and LDDM-U solve 32.3\% and 34.5\% of the problems. Without the token embeddings (iii), both solve 0.0\%. Without the SC state at all positions (i), they solve 8.9\% and 10.7\%. Since LDDMs train with a zero SC state on 10\% of the updates, (i) is a configuration seen during training, whereas (ii) and (iii) are not. Zeroing the SC state only at the response positions (ii) gives 7.1\% for LDDM-M, but 0.6\% for LDDM-U.

Zero token embeddings never occur during training. As a control, we instead feed fresh uniform tokens at the response positions at every step to LDDM-U fine-tuned from Duo for 50k steps (\supp{app:tinygsm-hard-tokens}). LDDM-U sees uniform tokens during training, at high noise levels. With 256 NFE at $T=1$, it then solves 0.2\% of the problems, against 1.8\% with zeroed token embeddings and 31\% without ablation. Hence, the SC state of LDDMs carries useful information, but LDDMs rely mainly on their discrete tokens and cannot generate from the SC state alone, which is what \method does.

\FloatBarrier
\subsection{TinyGSM: Comparison with Few-Step Methods}
\label{app:tinygsm-few-step}
\tab{tab:app-tinygsm-few-step} compares \method with recent few-step methods on TinyGSM, from 1 to 64 NFE.

\begin{table}[!ht]
\centering
\small
\setlength{\tabcolsep}{4pt}
\caption{\textbf{GSM8K accuracy (\%) with 1 to 64 NFE on TinyGSM.} MDLM, Duo and IDLM are taken from \citet{gushchin2026alphadlm}, who re-evaluate the checkpoints of \citet{deschenaux2026hypersphericalflows} and \citet{li2026idlm} (IDLM at 32 and 64 NFE is taken from \citet{li2026idlm}). FMLM+ is taken from \citet{agarwal2026posterior}, DBTM from \citet{tang2026dbtm}, AlphaDLM from \citet{gushchin2026alphadlm}, and SDM from \citet{deschenaux2026simplex}, whose sampler uses the stochasticity $\kappa=1$. SDM (distilled) starts from SDM (Expected Embedding), with a configuration selected on the GSM8K test set. FMLM+ (init.\ from MDLM) starts from an MDLM checkpoint, IDLM is distilled from MDLM or Duo, and the fine-tuned \method starts from the Duo baseline. Greedy denotes argmax decoding, and for FMLM+ and DBTM, an NFE is one refinement round. MDLM (confidence) at 1 and 2 NFE is computed from the counts reported by \citet{gushchin2026alphadlm}. -- marks values that are not reported. Best per column in \textbf{bold}.}
\label{tab:app-tinygsm-few-step}
\begin{tabular}{@{}llrrrrrrr@{}}
\toprule
Method & Decoding & 1 & 2 & 4 & 8 & 16 & 32 & 64 \\
\midrule
\multicolumn{9}{@{}l}{\textit{Discrete Diffusion}} \\
\hspace*{0.8em}MDLM (ancestral) & $T=0.1$ & -- & -- & 1.7 & 5.3 & 13.0 & 22.1 & 28.2 \\
\hspace*{0.8em}MDLM (confidence) & greedy & 0.8 & 0.7 & 4.2 & 20.1 & 37.7 & 43.9 & 40.6 \\
\hspace*{0.8em}Duo (ancestral) & $T=0.1$ & -- & -- & 5.2 & 14.0 & 23.0 & 30.8 & 31.2 \\
\midrule
\multicolumn{9}{@{}l}{\textit{Discrete Diffusion, distilled}} \\
\hspace*{0.8em}IDLM (MDLM) & $T=1$ & -- & -- & 0.4 & 1.8 & 6.5 & 12.8 & 14.9 \\
\hspace*{0.8em}IDLM (Duo) & $T=1$ & -- & -- & 2.3 & 7.1 & 11.7 & 15.4 & 19.0 \\
\midrule
\multicolumn{9}{@{}l}{\textit{Posterior Refinement}} \\
\hspace*{0.8em}FMLM+ (from scratch) & greedy & 0.0 & 0.1 & 2.9 & 8.7 & 13.4 & 19.0 & 19.1 \\
\hspace*{0.8em}FMLM+ (init.\ from MDLM) & greedy & 0.3 & 0.7 & 5.1 & 15.1 & 26.1 & 31.8 & 33.6 \\
\midrule
\multicolumn{9}{@{}l}{\textit{Discrete Beckmann Transport}} \\
\hspace*{0.8em}DBTM & greedy & 0.8 & 1.1 & 5.7 & 10.1 & 14.3 & 16.8 & 16.2 \\
\midrule
\multicolumn{9}{@{}l}{\textit{Alpha Diffusion Language Models}} \\
\hspace*{0.8em}AlphaDLM ($k=2$) & greedy & 1.0 & 1.3 & 10.6 & 28.6 & 43.1 & -- & -- \\
\hspace*{0.8em}AlphaDLM ($k=4$) & greedy & 1.8 & 2.6 & 16.7 & 35.3 & 46.5 & -- & -- \\
\hspace*{0.8em}AlphaDLM ($k=8$) & greedy & 2.7 & 7.7 & 28.9 & \textbf{44.0} & 49.2 & 49.9 & 50.0 \\
\hspace*{0.8em}AlphaDLM ($k=16$) & greedy & \textbf{6.3} & \textbf{18.0} & \textbf{34.6} & 39.0 & 40.6 & 40.9 & 40.9 \\
\midrule
\multicolumn{9}{@{}l}{\textit{Simplex Diffusion ($\kappa=1$)}} \\
\hspace*{0.8em}SDM (Expected Embedding) & $T=1$ & 0.0 & 0.0 & 1.0 & 6.3 & 17.1 & 25.4 & 32.7 \\
\hspace*{0.8em}SDM (Expected Embedding) & $T=0.1$ & 0.0 & 0.2 & 5.4 & 17.8 & 28.2 & 36.5 & 39.5 \\
\hspace*{0.8em}SDM (argmax + SC) & $T=1$ & 0.0 & 0.0 & 0.6 & 4.5 & 7.7 & 16.8 & 22.4 \\
\hspace*{0.8em}SDM (argmax + SC) & $T=0.1$ & 0.0 & 1.0 & 8.0 & 22.7 & 33.9 & 40.7 & 42.0 \\
\hspace*{0.8em}SDM (distilled) & $T=1$ & -- & -- & -- & 32.1 & -- & 37.6 & 37.8 \\
\midrule
\multicolumn{9}{@{}l}{\textit{\method (\textbf{Ours})}} \\
\hspace*{0.8em}From scratch & $T=1$ & 0.0 & 0.0 & 1.0 & 4.3 & 12.7 & 21.4 & 27.2 \\
\hspace*{0.8em}From scratch & $T=0.1$ & 0.0 & 1.6 & 9.6 & 26.4 & 35.4 & 40.8 & 44.0 \\
\hspace*{0.8em}Fine-tuned, 15k steps & $T=1$ & 0.0 & 0.7 & 6.1 & 30.3 & \textbf{50.1} & \textbf{56.2} & \textbf{57.4} \\
\hspace*{0.8em}Fine-tuned, 15k steps & $T=0.1$ & 0.2 & 3.6 & 14.6 & 29.6 & 39.2 & 44.7 & 47.4 \\
\hspace*{0.8em}Fine-tuned, 50k steps & $T=1$ & 0.0 & 0.2 & 1.6 & 8.1 & 21.4 & 33.6 & 39.9 \\
\hspace*{0.8em}Fine-tuned, 50k steps & $T=0.1$ & 0.2 & 3.3 & 15.3 & 34.2 & 43.1 & 46.6 & 50.7 \\
\bottomrule
\end{tabular}
\end{table}

\FloatBarrier
\subsection{Maze: Exact Match vs.\ Valid Solution}
\label{app:maze-exact-vs-valid}
\begin{figure}[!ht]
    \centering
    \includegraphics[width=\linewidth]{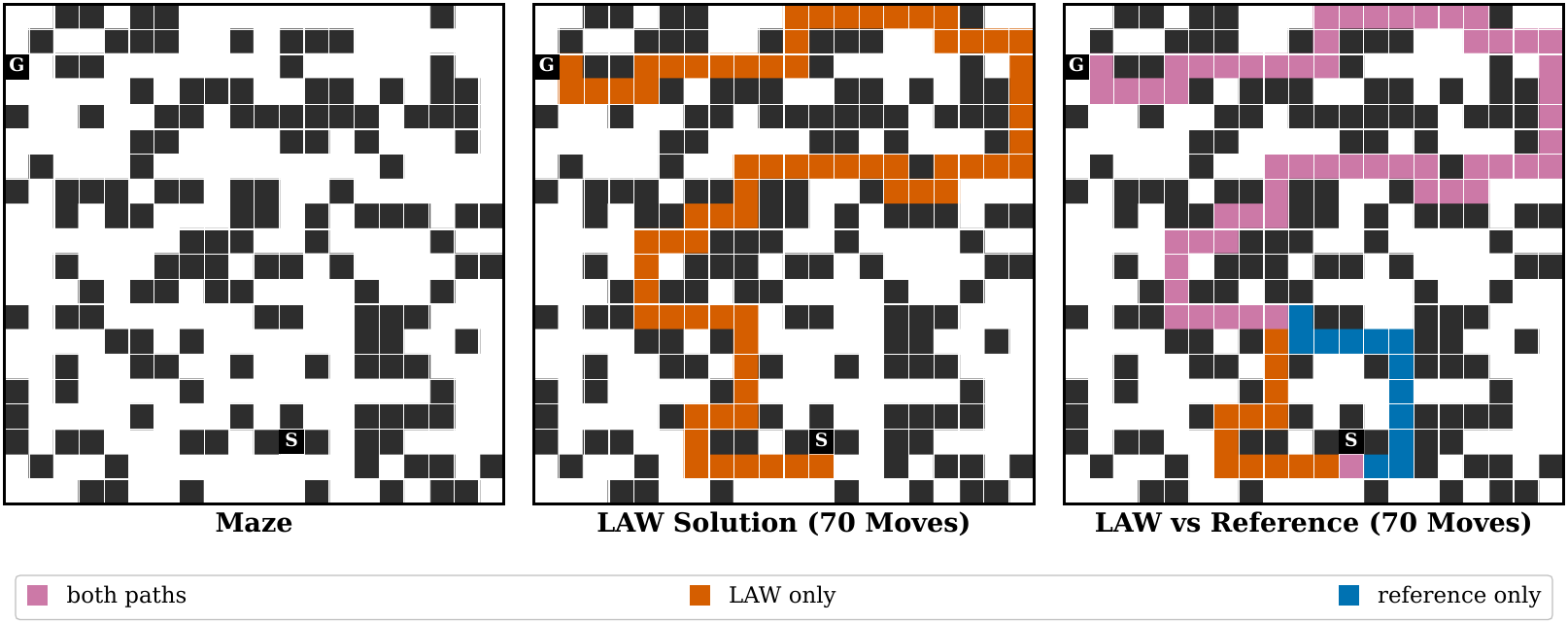}
    \caption{\textbf{A valid solution that fails \emph{exact match}.} On this Maze-Hard evaluation maze, \method generates a shortest path that differs from the target. \textbf{Left:} the maze. \textbf{Middle:} the path generated by \method. \textbf{Right:} cells on both paths (pink), only on the \method path (vermilion), and only on the target path (blue). Both paths take 70 moves, so the generated path solves the maze, but \emph{exact match} counts it as wrong.}
    \label{fig:maze-hard-alt-paths}
\end{figure}

\begin{figure}[!ht]
    \centering
    \includegraphics[width=\linewidth]{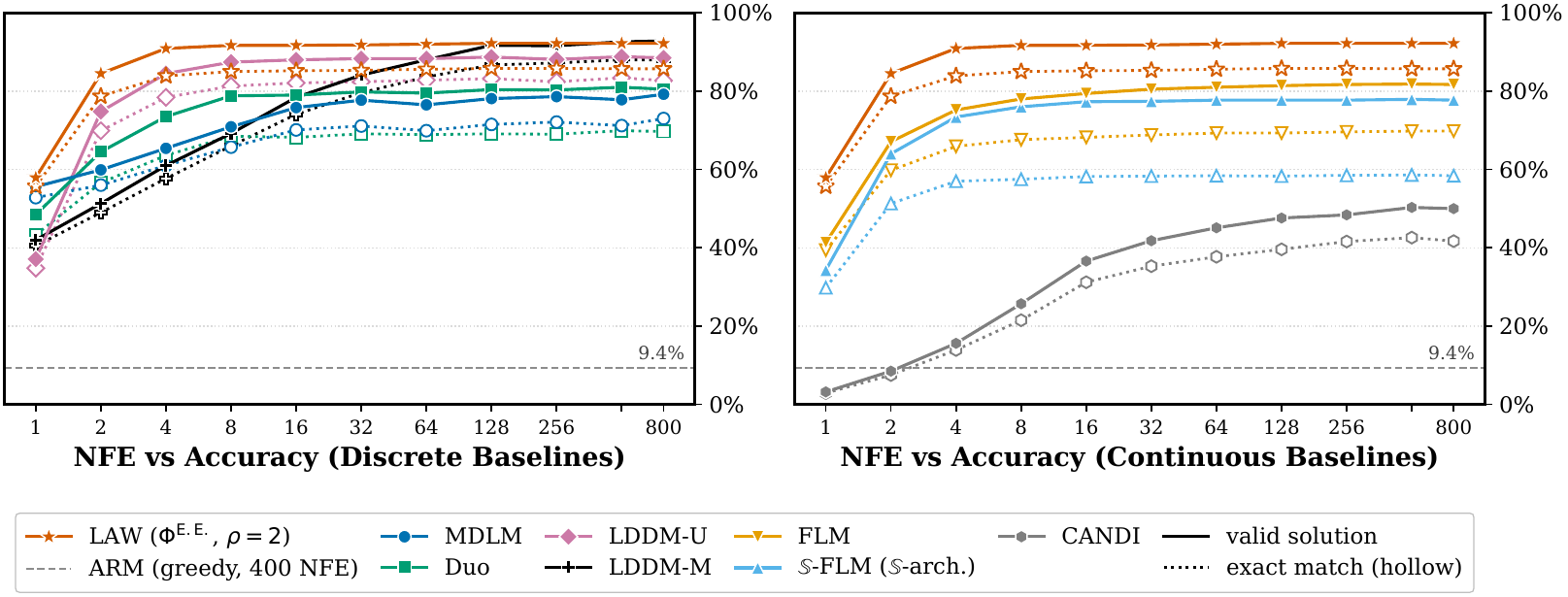}
    \caption{\textbf{Maze-Hard \emph{exact-match} against \emph{valid-solution} accuracy.} \emph{Valid-solution} (solid) and \emph{exact-match} (dotted, hollow markers) accuracy against NFE of the baselines and of \method with $\rho=2$ (\Fig{fig:maze-nfe}). \textbf{Left:} DDM baselines. \textbf{Right:} continuous baselines.}
    \label{fig:maze-nfe-exact-vs-valid}
\end{figure}

\emph{Exact match} counts many correct paths as wrong (\Fig{fig:maze-nfe-exact-vs-valid}). For example, $\mathbb{S}$-FLM ($\mathbb{S}$-arch.) generates a valid solution for 77.7\% of the mazes with 128 NFE, but only 58.3\% of them match the target path. Since every exact match is also a valid solution, the \emph{exact-match} accuracy never exceeds the \emph{valid-solution} accuracy (\supp{app:maze-metrics}). \Fig{fig:maze-hard-alt-paths} shows a valid path of \method that differs from the target.

\FloatBarrier
\subsection{Maze: Ablations}
\label{app:maze-ablations}
We report the \emph{valid-solution} accuracy at the last checkpoint of each run, with $T=1$. \tab{tab:app-maze-hard-steps} lists the configuration and training steps of every run.

\begin{figure}[!ht]
    \centering
    \includegraphics[width=\linewidth]{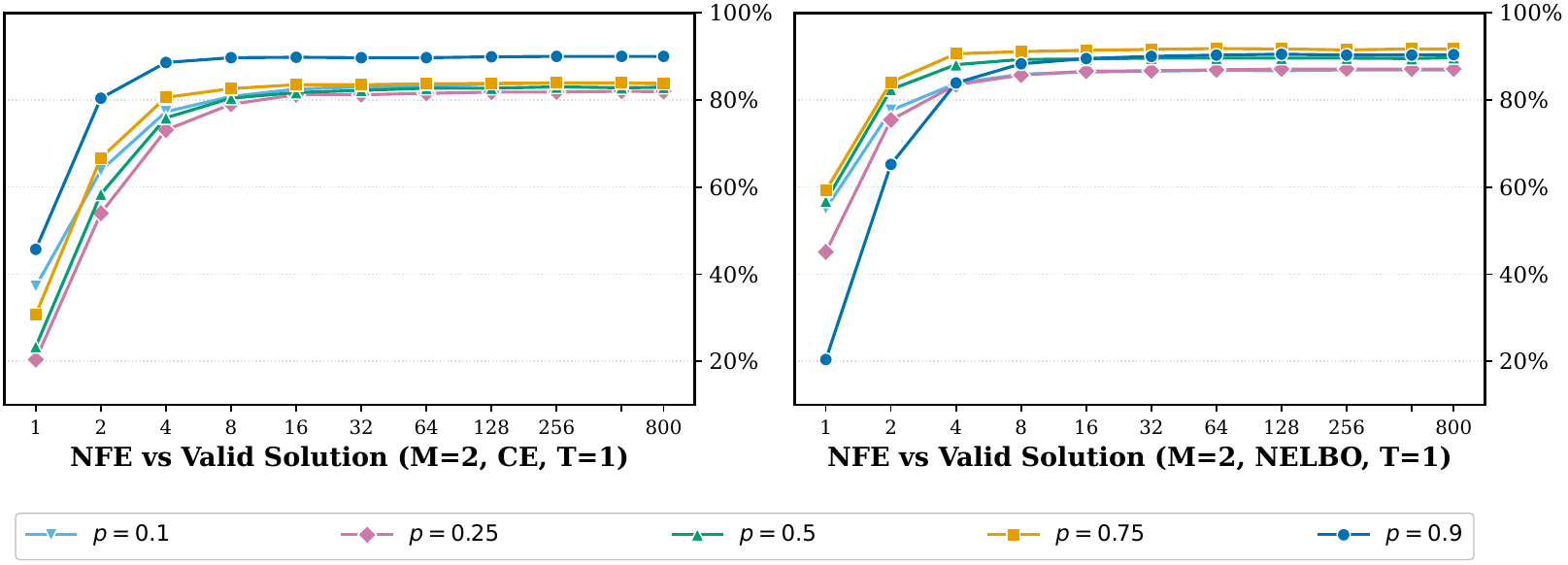}
    \caption{\textbf{Multi-pass probability on Maze-Hard.} \emph{Valid-solution} accuracy against NFE at $T=1$ of \method with $M=2$ passes run with probability $p$, trained with the CE (left) or the NELBO (right).}
    \label{fig:app-maze-p}
\end{figure}

\paragraph{Multi-Pass Probability and Loss.} With the NELBO, $p=0.75$ gives the highest accuracy, 91.7\% with 800 NFE (\Fig{fig:app-maze-p}). The NELBO is more accurate than the CE at every $p$ except $p=0.9$, where both solve about 90\% of the mazes.

\begin{figure}[!ht]
    \centering
    \includegraphics[width=\linewidth]{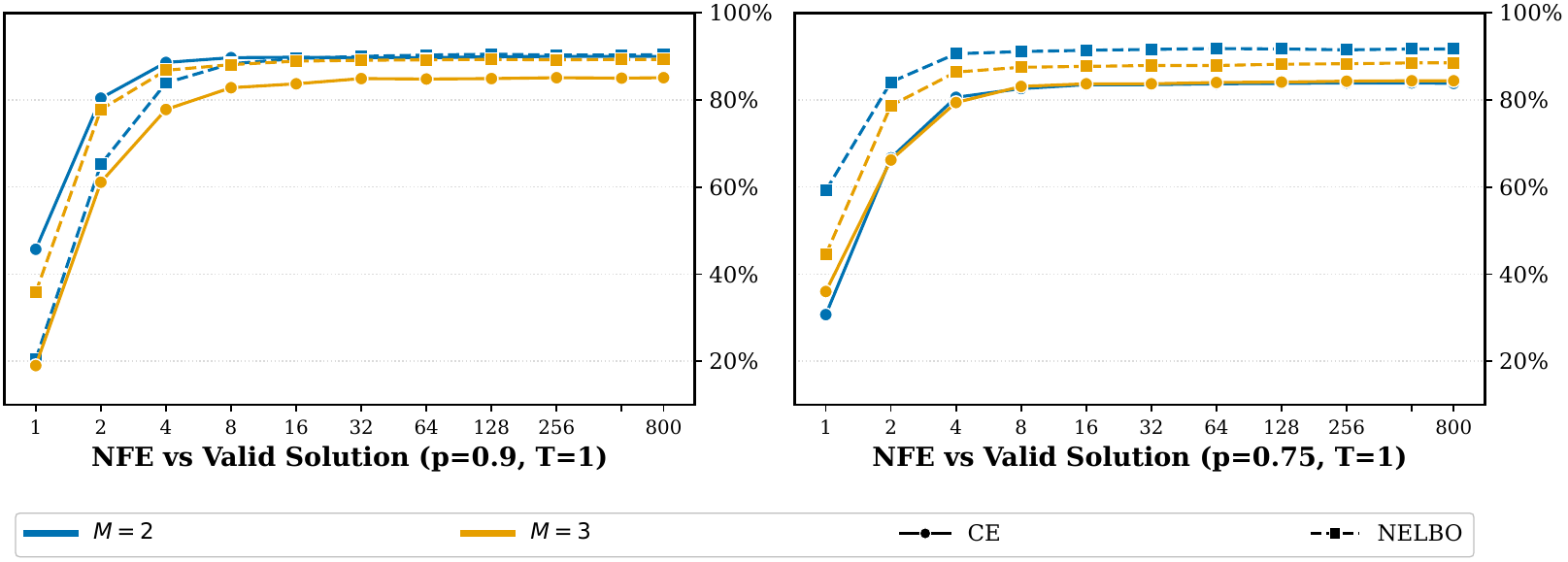}
    \caption{\textbf{Number of passes on Maze-Hard.} \emph{Valid-solution} accuracy against NFE at $T=1$ of \method with $M=2$ or $M=3$ passes, run with probability $p=0.9$ (left) or $p=0.75$ (right), and trained with the CE (solid) or the NELBO (dashed).}
    \label{fig:app-maze-passes}
\end{figure}

\paragraph{Number of Passes.} A third pass does not improve the final accuracy (\Fig{fig:app-maze-passes}). With $p=0.9$ and the CE, $M=3$ is less accurate than $M=2$ (85.1\% against 90.0\% with 800 NFE). With the NELBO, it helps with 2 NFE (77.8\% against 65.2\%), but not with many steps. With $p=0.75$, $M=3$ is below $M=2$ with the NELBO and close to it with the CE.

\begin{figure}[!ht]
    \centering
    \includegraphics[width=\linewidth]{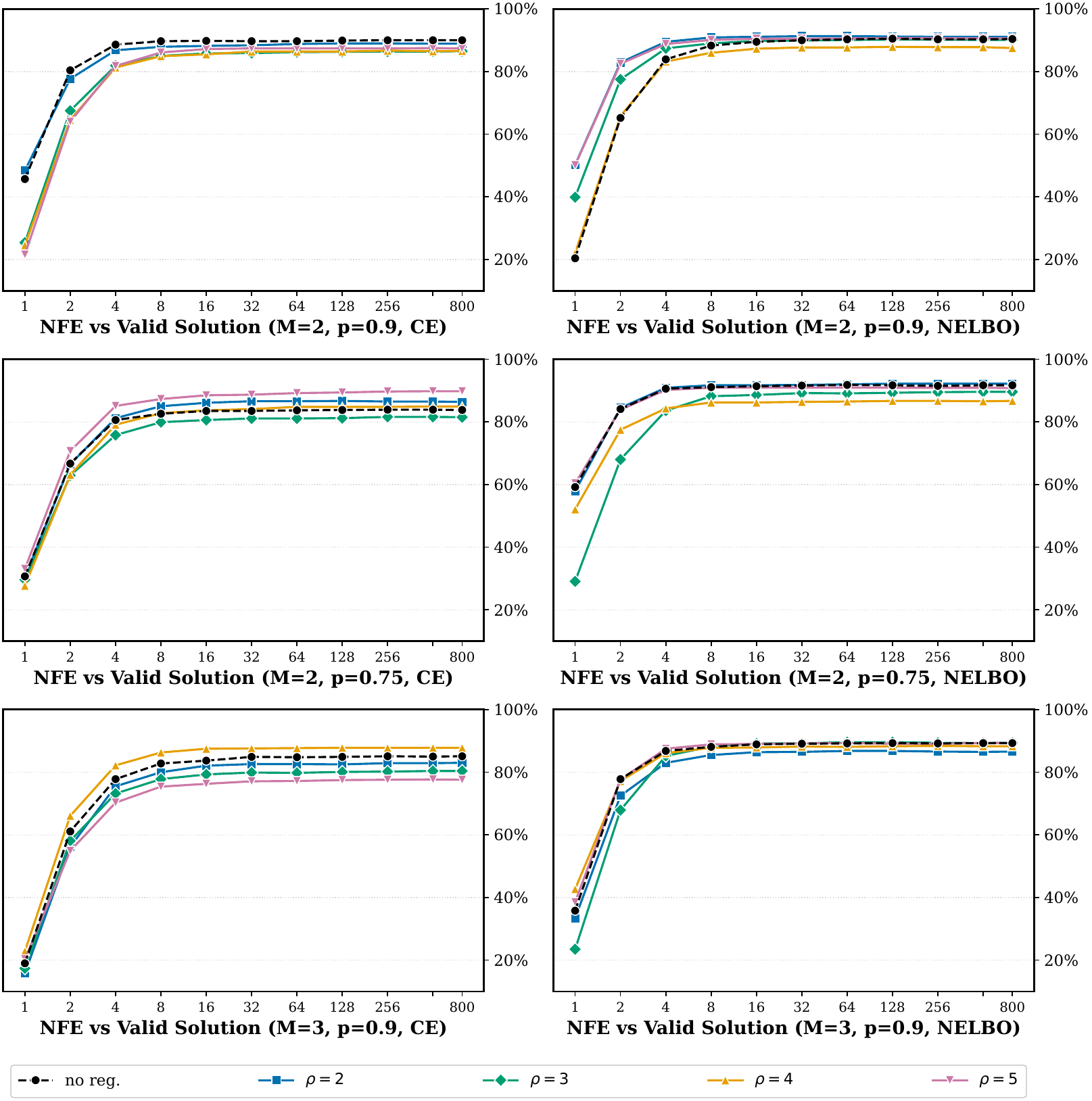}
    \caption{\textbf{Temperature regularization on Maze-Hard.} \emph{Valid-solution} accuracy against NFE at $T=1$. Each panel is one training configuration (x-axis label), without regularization (dashed black) and with $\rho\in\{2,3,4,5\}$.}
    \label{fig:app-maze-temp}
\end{figure}

\paragraph{Temperature Regularization.} Temperature regularization does not consistently improve the final accuracy (\Fig{fig:app-maze-temp}). We train \method with $\rho\in\{2,3,4,5\}$ and both losses for $M=2$ with $p\in\{0.9,0.75\}$ and for $M=3$ with $p=0.9$. The best configuration (NELBO, $M=2$, $p=0.75$, $\rho=2$) solves 92.2\% of the mazes with 800 NFE, against 91.7\% without regularization.

\FloatBarrier
\subsection{OpenWebText: Ablations}
\label{app:owt-ablations}
We first choose the loss, then add temperature regularization and the mode embedding, one change at a time (\tab{tab:app-owt-steps}). We track the metrics of \supp{app:owt-details} over training.

\begin{figure}[!ht]
    \centering
    \includegraphics[width=0.49\linewidth]{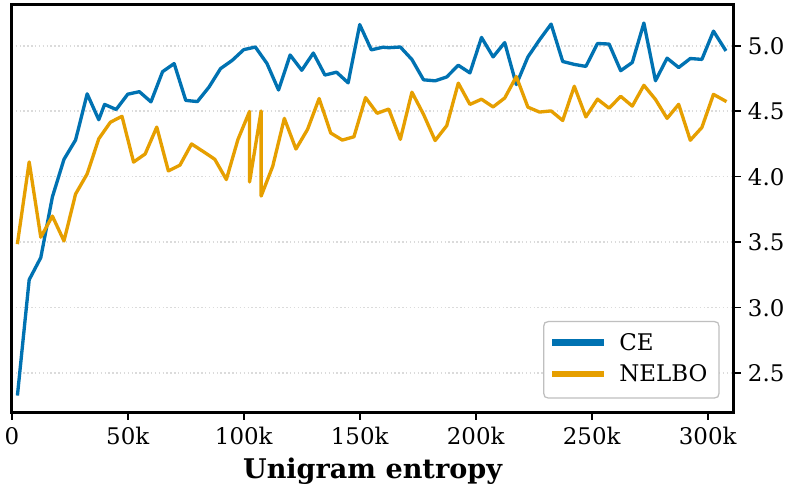}
    \hfill
    \includegraphics[width=0.49\linewidth]{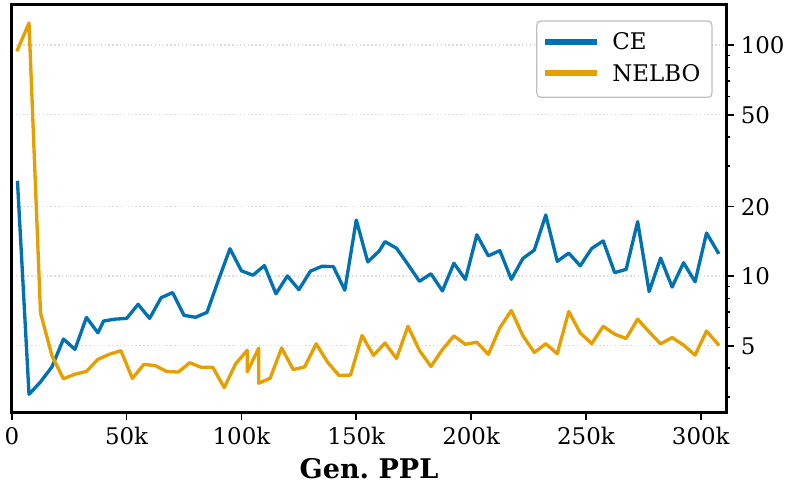}
    \caption{\textbf{Loss.} Unigram entropy (left) and Gen.\ PPL (right) over training steps of \method trained from scratch on OpenWebText with the NELBO or the CE~\Eqn{eq:law-loss}, without temperature regularization, and without the mode embedding.}
    \label{fig:owt-loss}
\end{figure}

\paragraph{Loss.} The CE~\Eqn{eq:law-loss} gives a higher entropy than the NELBO, i.e., more diverse samples, at a higher Gen.\ PPL (\Fig{fig:owt-loss}). We therefore use the CE in the following experiments.

\begin{figure}[!ht]
    \centering
    \includegraphics[width=\linewidth]{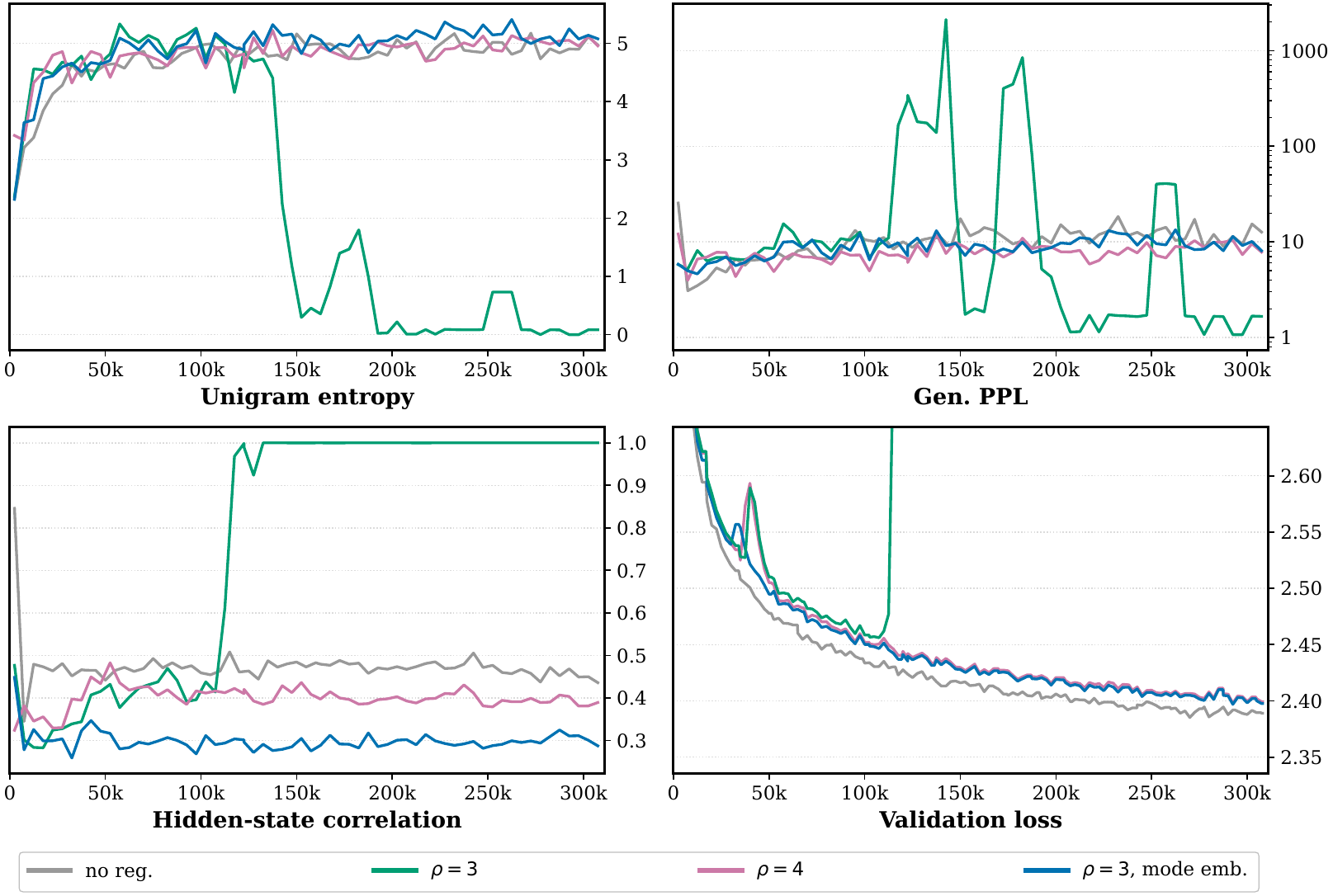}
    \caption{\textbf{Temperature regularization and mode embedding.} Unigram entropy (top left), Gen.\ PPL (top right), hidden-state correlation (bottom left), and validation loss (bottom right) over training steps of \method trained with the CE. No reg.\ is the CE run of \Fig{fig:owt-loss}. The other runs add temperature regularization with $\rho\in\{3,4\}$~\Eqn{eq:law-temperature}, then the mode embedding (\supp{app:mode-emb}) on top of $\rho=3$.}
    \label{fig:owt-temp-mode}
\end{figure}

\paragraph{Temperature Regularization and Mode Embedding.} The mode embedding prevents the collapse of the run with $\rho=3$ (\Fig{fig:owt-temp-mode}). Without it, the representations collapse midway through training, as the hidden-state correlation goes to 1 and the entropy drops close to zero. With it, the run reaches a lower hidden-state correlation and a slightly higher entropy than without regularization. The run with $\rho=4$ stays stable even without the mode embedding, and reaches a lower hidden-state correlation than the run without regularization, at a similar entropy.

\FloatBarrier
\subsection{OpenWebText: Comparison to LDDMs}
\label{app:owt-frontier-lddm}
\begin{figure}[!ht]
    \centering
    \includegraphics[width=\linewidth]{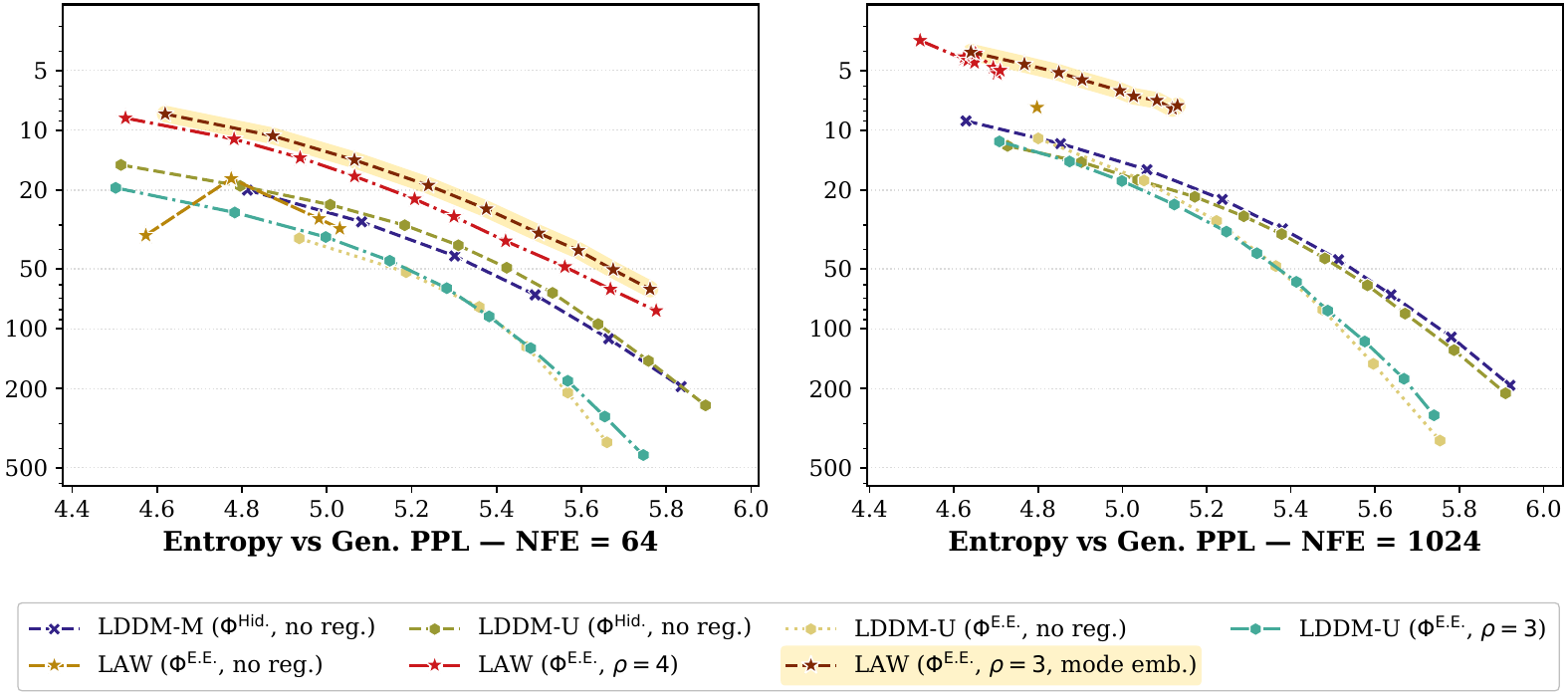}
    \caption{\textbf{\method against LDDMs on OpenWebText.} Gen.\ PPL against unigram entropy with 64 (left) and 1024 NFE (right), over the sampling temperatures of \Fig{fig:owt_entropy_x_gen_ppl}. The legend gives the SC state and the regularization (\supp{app:loophole-comparison} compares how LDDMs and \method use the SC state). The highlighted run is the \method model of \Fig{fig:owt_entropy_x_gen_ppl}.}
    \label{fig:owt-frontier-lddm}
\end{figure}

With temperature regularization, every \method run reaches a lower Gen.\ PPL than all LDDMs at both NFE (\Fig{fig:owt-frontier-lddm}), and with 64 NFE, the best model uses $\rho=3$ and the mode embedding. The figure compares the \method runs of \supp{app:owt-ablations} that do not collapse with LDDM-M and with LDDM-U using either SC state (\supp{app:owt-details}). Without regularization, \method only reaches a low entropy and a higher Gen.\ PPL than LDDM-U. Unlike on TinyGSM, the Expected Embedding increases the Gen.\ PPL of LDDM-U, with or without temperature regularization, most clearly at high entropy. The gain of \method over LDDMs therefore does not come from the Expected Embedding alone.

\FloatBarrier
\subsection{OpenWebText: Gradient Moment and 4-gram Overlap}
\label{app:owt-gm}
We evaluate 5000 samples per model, drawn at $T=1$, with two metrics (\tab{tab:app-owt-gm}). The Gradient Moment (GM; \citealp{hoogeboom2026discretemmd}) compares the samples with a reference set of 5000 OpenWebText validation sequences (lower is better). The 4-gram overlap measures how similar the samples of a model are to each other. For each sample, we compute the share of its distinct 4-grams (GPT-2 tokens) that occur in at least one of 5 other random samples of the same model, and average over the samples.

With 1024 NFE, only the ARM has a lower GM than \method. With 64 NFE, \method has a lower GM than both $\mathbb{S}$-FLM models, but a higher GM than the other baselines, and LDDM-U has the lowest GM. With 1024 NFE, the 4-gram overlap of \method is 0.70\%, against 0.68\% for the data. Hence, the samples of \method are not near-copies of each other, although their unigram entropy is lower than that of the baselines (\Fig{fig:owt_entropy_x_gen_ppl}). With 64 NFE, the 4-gram overlap of \method (0.20\%) is below that of the data.

\begin{table}[h]
\centering
\footnotesize
\setlength{\tabcolsep}{4pt}
\caption{\textbf{Gradient Moment (GM) and 4-gram overlap on OpenWebText}, from 5000 samples per model at $T=1$. The 4-gram overlap is the share (\%) of the distinct 4-grams of a sample that occur in at least one of 5 other samples. Its standard error is at most 0.02. For the data, the GM uses 5000 other validation sequences, and the 4-gram overlap uses the reference set. The ARM runs only with 1024 NFE. The OpenWebText data and the ARM are references and are not ranked. Lowest GM per column in \textbf{bold}, second lowest \underline{underlined}.}
\label{tab:app-owt-gm}
\begin{tabular}{@{}lcccc@{}}
\toprule
 & \multicolumn{2}{c}{64 NFE} & \multicolumn{2}{c}{1024 NFE} \\
\cmidrule(lr){2-3}\cmidrule(lr){4-5}
Model & GM ($\downarrow$) & 4-gram overlap (\%) & GM ($\downarrow$) & 4-gram overlap (\%) \\
\midrule
OpenWebText data & $0.001 \pm 0.004$ & 0.68 & $0.001 \pm 0.004$ & 0.68 \\
ARM & -- & -- & $0.150 \pm 0.005$ & 0.69 \\
\midrule
MDLM & $1.018 \pm 0.007$ & 0.68 & $0.883 \pm 0.007$ & 0.73 \\
Duo & $0.869 \pm 0.007$ & 0.90 & $0.849 \pm 0.008$ & 0.79 \\
FLM & $1.152 \pm 0.007$ & 0.27 & $1.200 \pm 0.014$ & 2.47 \\
$\mathbb{S}$-FLM & $1.800 \pm 0.013$ & 0.89 & $1.774 \pm 0.014$ & 0.98 \\
$\mathbb{S}$-FLM ($\mathbb{S}$-arch.) & $1.665 \pm 0.016$ & 1.90 & $1.632 \pm 0.017$ & 2.03 \\
LDDM-M & $\underline{0.841} \pm 0.006$ & 0.54 & $\underline{0.629} \pm 0.006$ & 0.79 \\
LDDM-U & $\mathbf{0.721} \pm 0.006$ & 0.94 & $0.658 \pm 0.007$ & 0.84 \\
\midrule
\method & $1.210 \pm 0.011$ & 0.20 & $\mathbf{0.565} \pm 0.008$ & 0.70 \\
\bottomrule
\end{tabular}
\end{table}

\clearpage
\subsection{OpenWebText: Samples}
\label{app:owt-samples}
We show unconditional samples of $L=1024$ tokens from the \method model of \sec{sec:experiments-owt}, drawn at $T=1$.

\begin{figure}[h]
    \lawsample{\eot{}bird to MozillaBluebird\textless{}/strong\textgreater{} \textless{}strong\textgreater{} \textless{}i\textgreater{}[Internet homophobe\textgreater{}: Filimonious blatant anti-aggression at 09:20 PM PDT\textless{}/\textless{}/\textgreater{}2\textgreater{} Mozozilla BlueHTML IE is incompatible with HTML5. Sincerely, Mozilla Bluebird\textless{}/strong\textgreater{} \textless{}strong\textgreater{}Microsoft: IE was never spanned HTML5\textless{}/strong\textgreater{} \textless{}strong\textgreater{}All my ducks\textless{}/blockquote\textgreater{}\textless{}a href=\qdbl{}\url{https://twitter.com/TMZWB/status/758308800725903404}\qdbl{}\textgreater{}\textless{}/script\textgreater{} \textless{}p\textgreater{}Firefox knuckleball\textless{}a href=\qdbl{}\url{https://twitter.com/jonathanjoc}\qdbl{}\textgreater{}\url{https://twitter.com/jonathan_jockels/}\textgreater{}\textless{}img link=\qdbl{}/tweetShowtime embed\qdbl{}\textgreater{}\textless{}/img\textgreater{} \textless{}i\textgreater{}[email protected\textless{}/i\textgreater{}(Message ID: Send embed message to your Firefox Web browser)\textless{}/i\textgreater{}\textless{}strong\textgreater{}All my ducks\textless{}/blockquote\textgreater{}\textless{}\textless{}a href=\qdbl{}\url{https://blockquote.com/TMZWB/status/76912880071803404}\qdbl{}\textgreater{}\textless{}/script\textgreater{} \textless{}i\textgreater{}[Internet homophobe\textgreater{}: Filimonious anti-aggression, 09:20 PM PDT\textless{}/t\textgreater{}\textgreater{} Mozozilla HTMLbird browser is incompatible with HTML5. Sincerely, Mozilla Bluebird and MozillaBluebird\textless{}/strong\textgreater{} \textless{}a href=\qdbl{}\url{https://blockquote.com/TMZWB/status/82317880070103404}\qdbl{}\textgreater{}\textless{}/script\textgreater{} \textless{}i\textgreater{}[email protected\textless{}/i\textgreater{}(Photo credit: iStockphoto/Getty Images)\textless{}/i\textgreater{}\textless{}p\textgreater{}Firefox knuckleball\textless{}a href=\qdbl{}\url{https://twitter.com/jonathanjoc}\qdbl{}\textgreater{}\url{https://twitter.com/jonathan_jockels/}\textgreater{}\textless{}img link=\qdbl{}/tweetShowtime embed\qdbl{}\textgreater{}\textless{}/a\textgreater{} \textless{}strong\textgreater{}Microsoft: IE was never spanned\textless{}/strong\textgreater{} \textless{}/strong\textgreater{}\eot{}Video Loading Video Unavailable Click to play Tap to play The video will start in 8 Cancel Play now\nl{}\nl{}Liverpool are poised to reject a £7.75million bid from Lyon for world-record signing Alexandre Lacazette but but Brendan Rodgers is keen to begin talks with the France international.\nl{}\nl{}The 35-year-old France international is close to start his ninth Premier League season at Stamford Bridge after agreeing previously agreed terms over £7.5m wages and release clause.\nl{}\nl{}Rodgers is keen to keep Lac a Frenchman, has scored 35 goals in 155 Premier League games since joining from Ukrainian giants Dynamo Moscow and Shakhtar Donetsk in January 2012.\nl{}\nl{}Lac Lazette 20 Chelsea Lyon season after completing a £9.25m loan deal from Spanish academy club Boca Raton and £11.15m from German side Bayer Leverkusen.\nl{}\nl{}The 25-year-old was capped 17 times by France at Euro 2012, has made 25 first-team appearances at the Etihad Stadium and captained Dynamo Moscow after scoring a brace in a 2-1 win over Dortmund in the opening round of the Europa League quarter-final.\nl{}\nl{}Video Loading Video Unavailable Click to play Tap to play The video will start in 8 Cancel Play now\nl{}\nl{}Rodgers is keen to reject a chance bid for Bosnia striker Raferski Kljanic who Arsenal has rejected Marse bids from AC Milan, Roma and Lazio with the club-record £75m release clause which expires when January transfer window reopens.\nl{}\nl{}The 26-year-old France international netted 15 goals and 15 assists in nine Premier League starts since joining from Ukrainian giants Shakhtar Donetsk in January 2012.\nl{}\nl{}The 35-year-old France international was also to offer a of £4.3million from Bundesliga heavyweights Borussia Dortmund and Bayern Munich this summer after agreeing previously agreed terms on a £3m wages and release clause.\nl{}\nl{}Lacazette is keen to become a regular in the Reds' starting-team at winning Brendan Cup from out in time, League. Rodgers\nl{}Rod Rodgers is keen to begin talks with Lacazette for his first Premier League season at Stamford Bridge.\nl{}\nl{}Liverpool boss Jurgen Klopp also admitted Arsenal's Real Sociedad Alexis Sanchez will be omitted from the starting XI with a torn anterior cruciate ligament/left thigh injury.\nl{}\nl{}The 37-year-old Chilean is expected to miss crucial Champions League games against Paris Saint-Germain, Real Madrid and Barcelona at the Etihad on Sunday.\eot{}Image copyright PA Image caption The riot police have cordoned off areas of the stadium following Chelsea's 2-2 draw with Bournemouth\nl{}\nl{}Riot police have cordoned off areas at the stadium following Chelsea's 2-2 draw with Bournemouth.\nl{}\nl{}Video footage shows police cars speeding through the pitch at knee-high speed.\nl{}\nl{}Chelsea fans responded off ``Goite!'' chants as police cleared effective areas of the pitch.\nl{}\nl{}The hosts had who outemouth 5-0 in goal opening two minutes.\nl{}\nl{}Manager Jose Mourinho claims a have `` the it's by\eot{}}
    \caption{\textbf{Sample with 64 NFE.} Gen.\ PPL 18.31, unigram entropy 5.60.}
    \label{fig:owt-sample-nfe-64}
\end{figure}

\begin{figure}[h]
    \lawsample{\eot{} spill in be devastating to reef's economy marine and are communities and damaging Great Barrier Reef Heritage status and natural heritage.''\nl{}\nl{}Federal environment minister Patrick McLuhan said further losses from Maccorp would be ``catastrophic'' for the reef ecosystem.\nl{}\nl{}``Maccor's are, highly on are oil sands is and are dumping and dredging down our runways and waterways,'' McLuhan said.\nl{}\nl{}``These oil spills will wreak havoc on this reef environment, wreak havoc on the tourism and fishing industries and devastate the health and livelihoods of the reef.''\nl{}\nl{}Queensland marine environment minister Greg Hunt said it was ``surprising'' the fossil fuel industry ``has become involved'' in the unfolding spill.\nl{}\nl{}``[Industries such as ExxonMobil, Shell, Halliburton, Symantec and Bonhoeven Energy are become involved in their day-to-day operations,'' Hunt said.\nl{}\nl{}``ExxonMobil, Halliburton, Symantec, Bonhoeven Energy and Shell marine field activities could greatly impacted.''\nl{}\nl{}Hunt also the Australian Federal Environment and Consumer Protection Agency was monitoring the situation.\nl{}\nl{}`Oil spill impacts Great Barrier Reef ecosystem'\nl{}\nl{}Great Barrier Reef Environment scientist Andy Bourke said there was ``fundhuge concern'' about the spill.\nl{}\nl{}Reflecting on the recent car crash off Queensland's eastern coast, Bourke said Reef Marine Incident Management Team alerted the authorities a vast quantity crude oil had washed up corals.\nl{}\nl{}``After extensive dredging and clearing corals we discovered large quantities of oily sediment that could've been contaminated with crude oil,'' Bourke said.\nl{}\nl{}Environment Minister Greg Egan told RN One radio the government was taking urgent action to prevent oily material from being released into the waters. Deepwater activists have warned similar spill spill be spill well after the massive flood which inundated the Great Barrier Reef in the late 1990s and early 2000s.\nl{}\nl{}Maccorp used a mixture of flammable dispersants, lexit 9590 and Corexit 8590 – toxic chemicals. The U.S. Environmental Protection Agency says some of the chemicals emitted by Corexit dispersants exceeded EPA's maximum allowable amount 100 micrograms.\nl{}\nl{}Opposition senator Clive Palmer has labelled the government's decision to suspend a moratorium on offshore exploration of the reef as ``disastrous''.\nl{}\nl{}``It has dramatically changed the way we think about the future of the Great Barrier Reef,'' Senator Clive Palmer told ABC\nl{}\nl{}``It is an irresponsible, reckless decision to suspend a moratorium on offshore exploration of the reef.''-{}-{}-\nl{}-{}-{}-environment-{}-{}-impact-{}-{}-issues-{}-{}-environment-{}-{}-government, federal-{}-{}-state-{}-{}-issues-{}-{}-national-{}-{}-security, brisbane-4000, sydney-6748, qld, australia\nl{}\nl{}First posted\eot{}Long-time meteorologist Thomas Hockenberger talks about Comet Listen to the interview BBC radio presenter Alistair Clarkson (left) gave about Comet in the 1985 Life Science documentary `What caused the asteroid impact?' AP Long-time meteorologist Thomas Hockenberger was asked about Comet in the 1985 American documentary documentary `What caused an asteroid impact?' BBC radio presenter Alistair Clarkson (right) was asked about Comet in the 1985 PBS science documentary `What caused an asteroid impact?' AP Image 1 of / 37 Caption Close NASA celebrates 25th anniversary of biggest asteroid impact event recorded Image 1 of / 37 Caption Close Long-time meteorologist Thomas Hockenberger talks about Comet 1 / 37 Back to Gallery\nl{}\nl{}MADISON, Wis. — Astrophysicist Thomas Hockenberger spoke about Comet as NASA celebrates the 25th anniversary of the biggest asteroid impact event in recorded history.\nl{}\nl{}Hockenberger presents part on The{\color{gray}?} The details, which of a cataclysmic asteroid impact believed to have wiped out thousands of vertebrates, from dinosaurs to centroids to arthropods.\nl{}\nl{}Astronomers believe an ejection meteorite fragment from the Cretaceous-Earth-Triassic boundaryogene collided with the asteroid belt and ejected carbon dioxide and dust. That ejected carbon dioxide particles from the atmosphere and smashed through forming a tangle of soft rocks that eroded the Earth’s geologic crust.\nl{}\nl{}Scientists also that late Cretaceous boundaryogene smashed the ejection meteorite fragments and ejected carbon dioxide and dust deposited in Earth's mantle, created a comet nucleus, impacting on asteroid in Earth.\nl{}\nl{}[SLIDESHOW: Earth’s Biggest Asteroid Impact Event Revealed]\nl{}\nl{}Listen to the fascinating interview about Comet in the podcast below.\nl{}\nl{}Watch legendary meteorologist Thomas Hockenberger talk to our daily newsletter To stay up to date with the latest news and videos, sign up for the Us Weekly iPhone app or subscribe to the Us Weekly Android app.\eot{}(CNN) Japan has extended issued tsunami warning warnings for major quakes along its northeastern coast, prompting experts to warn that a 6.9-magnitude quake is still Japan at\eot{}}
    \caption{\textbf{Sample with 1024 NFE.} Gen.\ PPL 22.09, unigram entropy 5.69.}
    \label{fig:owt-sample-nfe-1024}
\end{figure}

\end{document}